\documentclass[11pt]{article}

\PassOptionsToPackage{table}{xcolor}
\usepackage[final]{acl}

\usepackage[english,bidi=default]{babel}
\babelfont{rm}[
  BoldFont=TeXGyreTermesX-Bold.otf,
  ItalicFont=TeXGyreTermesX-Italic.otf,
  BoldItalicFont=TeXGyreTermesX-BoldItalic.otf
]{TeXGyreTermesX-Regular.otf}
\usepackage{latexsym}

\usepackage{microtype}

\usepackage{inconsolata}

\usepackage{graphicx}

\usepackage{hyperref}
\usepackage{url}
\usepackage{float}

\usepackage{hyperref}       
\usepackage{url}            
\usepackage{booktabs}       
\usepackage{amsfonts}       
\usepackage{nicefrac}       
\usepackage{microtype}      
\usepackage{xcolor}         
\usepackage{enumitem}
\usepackage{amsmath} 
\usepackage{amssymb} 
\usepackage{booktabs} 
\usepackage{makecell} 
\usepackage{multirow} 
\usepackage{wrapfig}
\usepackage{colortbl}   
\usepackage{array}    
\usepackage[linesnumbered,ruled,vlined]{algorithm2e} 
\SetKwInOut{KwIn}{Input}
\SetKwInOut{KwOut}{Output}
\SetKwComment{Comment}{$\triangleright$\ }{} 
\usepackage[most]{tcolorbox}
\usepackage{listings}
\usepackage{xcolor}

\usepackage{pifont}
\usepackage{graphicx} 
\usepackage[most]{tcolorbox}
\usepackage{xcolor}
\definecolor{myback}{RGB}{255,252,246}
\definecolor{mytitle}{RGB}{252,252,240}

\tcbset{
    prompt/.style={
        colback=myback,
        colframe=black,
        boxsep=1pt,
        arc=5pt,
        boxrule=1pt,
        width=\textwidth,
        enhanced,
        drop shadow={black!30!white},
        boxed title style={
            colback=mytitle,
            colframe=black,
            boxrule=1pt,
            sharp corners,
            size=small,
            boxsep=2pt,
            left=3mm,
            right=3mm,
            fontupper=\color{black}\bfseries,
            center title
        },
        attach boxed title to top center={ 
            yshift=-\tcboxedtitleheight/2
        },
        top=4mm,
        bottom=2mm,
        coltitle=black
    }
}

\providecommand{\rev}[1]{#1}

\definecolor{tabgreenhead}{RGB}{129,199,132}
\definecolor{tabgreenrowa}{RGB}{232,245,233}
\definecolor{tabgreenrowb}{RGB}{200,230,201}

\definecolor{tabbluehead}{RGB}{100,181,246}
\definecolor{tabbluerowa}{RGB}{227,242,253}
\definecolor{tabbluerowb}{RGB}{187,222,251}

\definecolor{tablavhead}{RGB}{149,117,205}
\definecolor{tablavrowa}{RGB}{237,231,246}
\definecolor{tablavrowb}{RGB}{209,196,233}

\title{SocialMaze: A Benchmark for Evaluating and Enhancing Social Reasoning in Large Language Models in Complex Social Environments}

\author{
  \textbf{Zixiang Xu}\textsuperscript{1,2},
  \textbf{Yanbo Wang}\textsuperscript{1},
  \textbf{Yue Huang}\textsuperscript{3},
  \textbf{Haomin Zhuang}\textsuperscript{3},
  \textbf{Yujun Zhou}\textsuperscript{3} \\
  \textbf{Jiayi Ye}\textsuperscript{1},
  \textbf{Sixian Li}\textsuperscript{4},
  \textbf{Zirui Song}\textsuperscript{1},
  \textbf{Lang Gao}\textsuperscript{1},
  \textbf{Chenxi Wang}\textsuperscript{1},
  \textbf{Zhaorun Chen}\textsuperscript{5} \\
  \textbf{Wang Pan}\textsuperscript{6},
  \textbf{Yue Zhao}\textsuperscript{2},
  \textbf{Jieyu Zhao}\textsuperscript{2},
  \textbf{Xiangliang Zhang}\textsuperscript{3},
  \textbf{Xiuying Chen}\textsuperscript{1} \\[0pt]
  {\normalfont
  \begin{tabular}{@{}c@{}}
    \textsuperscript{1}\,Mohamed bin Zayed University of Artificial Intelligence \\
    \textsuperscript{2}\,University of Southern California \quad
    \textsuperscript{3}\,University of Notre Dame \\
    \textsuperscript{4}\,University of Michigan, Ann Arbor \quad
    \textsuperscript{5}\,University of Chicago \quad
    \textsuperscript{6}\,Microsoft
  \end{tabular}}
}

\hypersetup{
  pdftitle={SocialMaze: A Benchmark for Evaluating and Enhancing Social Reasoning in Large Language Models in Complex Social Environments},
  pdfauthor={Zixiang Xu, Yanbo Wang, Yue Huang, Haomin Zhuang, Yujun Zhou, Jiayi Ye, Sixian Li, Zirui Song, Lang Gao, Chenxi Wang, Zhaorun Chen, Wang Pan, Yue Zhao, Jieyu Zhao, Xiangliang Zhang, Xiuying Chen},
  pdfsubject={EMNLP 2026 Findings},
  pdfkeywords={SocialMaze, social reasoning, large language models, benchmark, dynamic interaction, information uncertainty}
}

\begin{document}
\maketitle
\begin{abstract}
Large language models (LLMs) are increasingly deployed in socially grounded applications, where success requires interpreting context, inferring others' mental states, and reasoning about unreliable information.
Yet existing benchmarks rarely evaluate these demands jointly in complex, evolving settings.
We introduce \textbf{\textit{SocialMaze}}, a benchmark that organizes six tasks across social deduction games, daily-life interactions, and digital community platforms along three descriptive design axes: deep reasoning, dynamic interaction, and information uncertainty.
These axes characterize intended sources of task difficulty rather than latent, factor-analytic dimensions of model capability.
Automated checks and human validation support data quality.
Evaluations of twelve proprietary and open-weight LLMs show substantial variation in the use of evolving interaction histories; stronger chain-of-thought reasoners perform better on tasks requiring deeper inference, while uncertainty consistently degrades performance.
Reasoning workflows help weaker short-chain-of-thought backbones but saturate on stronger reasoners.
Finally, targeted fine-tuning on curated reasoning traces substantially improves structured social-reasoning tasks, whereas transfer to language-aggregation tasks remains statistically inconclusive.
The project homepage is available at \url{https://xzx34.github.io/socialmaze/}.
\end{abstract}

\section{Introduction}

LLMs demonstrate significant capabilities across domains such as scientific discovery \citep{guo2023can,chen2025unveiling} and medicine \citep{zhou2024pre,chen2025evaluating}. They are also increasingly applied to socially grounded tasks, including online community moderation, media content analysis, and social deduction games \citep{li2023theory, wei2025exploring}. These applications require \textit{\textbf{social reasoning}}: interpreting social context, inferring others' mental states, and judging claims in light of incomplete or unreliable information.

\begin{figure*}[t] 
    \centering
    \vspace{-1em}
    \includegraphics[width=\linewidth]{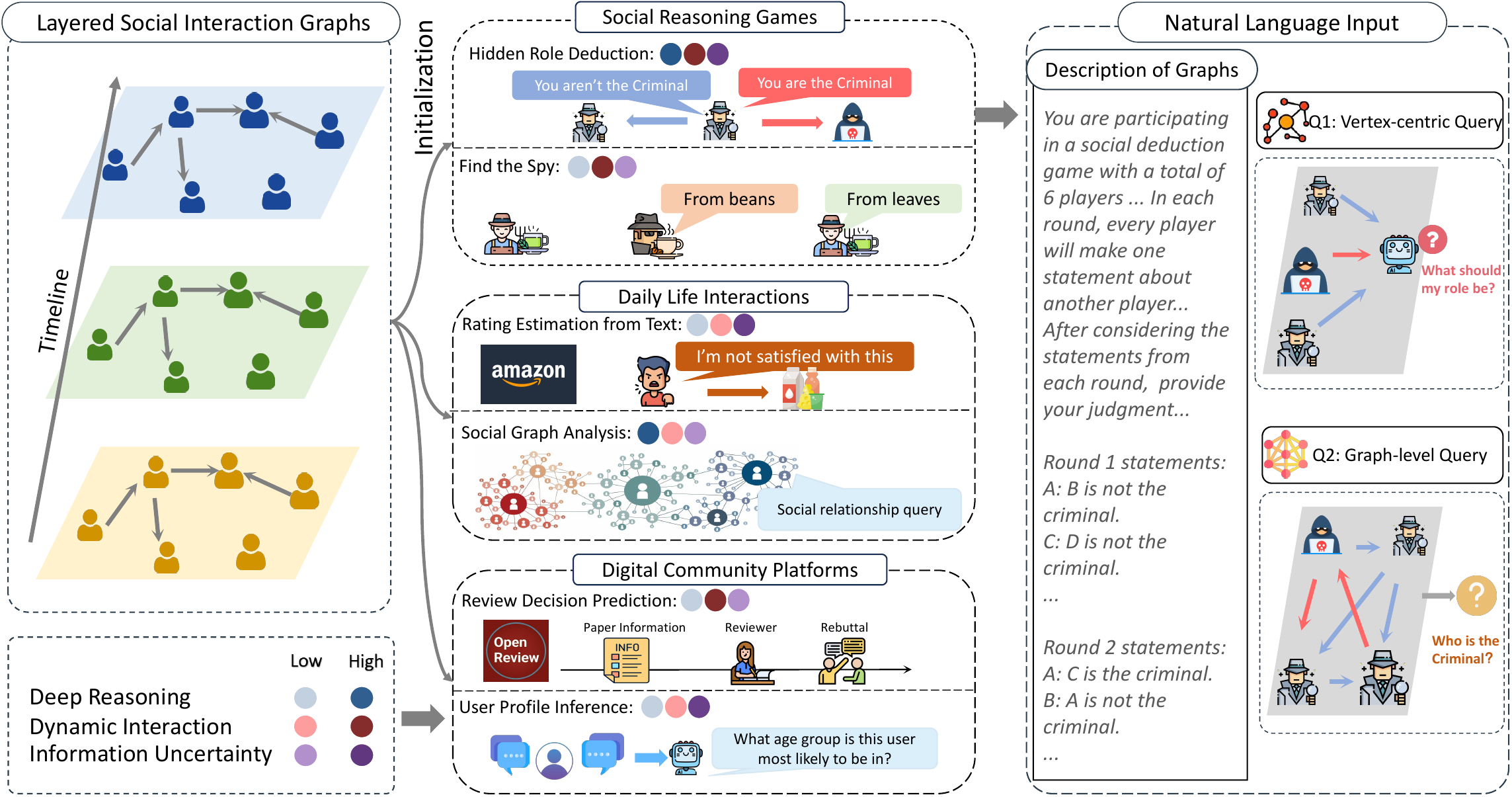} 
    \caption{Overview of the SocialMaze Benchmark.
All tasks are built upon (a) a time-aware modeling framework for complex social networks.
Based on this template, we instantiate (b) six diverse task types, covering social reasoning games, daily-life interactions, and digital community platforms.
(c) An illustrative Hidden Role Deduction instance with graph descriptions and vertex-centric and graph-level queries.}
    \label{fig:pipeline} 
    \vspace{-1em}
\end{figure*}

While existing benchmarks effectively evaluate the general capabilities of LLMs \citep{huang2025trustworthiness,kosinski2024evaluating}, benchmarks specifically designed to assess social reasoning abilities face significant limitations: (1) reliance on static scenarios lacking dynamic interaction \citep{sap2019socialiqa,sap2019atomic,onoe2021creak,demszky2020goemotions}, (2) overly sanitized information devoid of the noise, bias, or deception common in real social environments \citep{kosinski2024evaluating,lin-etal-2020-commongen, shapira2023clever, nematzadeh2018evaluating}, and (3) tasks too simple to capture the deeper cognitive aspects of social inference \citep{forbes2020social,hendrycks2021ethics,tiwari2025debatebench}. For example, SocialIQA \citep{sap2019socialiqa} primarily assesses commonsense reasoning within simplified, predefined social contexts, and Theory-of-Mind (ToM) benchmarks \citep{gandhi2023understanding} evaluate mental state inference based on static narratives lacking deceptive elements or informational uncertainty. Strategic and deduction games such as Diplomacy \citep{bakhtin2022mastering}, Avalon, and Werewolf \citep{wei2025exploring,light2023avalonbench, bailis2024werewolf, xu2023exploring} place LLMs in dynamic environments but evaluate task outcomes (completion, winning) rather than whether the model reasons coherently about the underlying social logic.

To address these limitations, we argue that the assessment of social reasoning should explicitly incorporate three aspects: \textbf{\textit{Deep Reasoning}} (inferring latent mental states, causal relations, counterfactuals, and strategic plans) \citep{premack1978does, wellman2014making, kunda1999social, frith2007social, gandhi2023understanding, roese1997counterfactual, huang2023clomo, zhang2024llm}, \textbf{\textit{Dynamic Interaction}} (tracking evolving context across multiple turns and adapting to anticipated responses) \citep{emirbayer1998agency,pentland2014social,shoham2008multiagent,park2023generative,chen2025shieldagent}, and \textbf{\textit{Information Uncertainty}} (handling misinformation, subjective bias, and intentional deception in noisy social information landscapes) \citep{wang2018fusing,baldwin2013noisy,preiser2018social,lazer2018science}. \rev{We treat these as descriptive design axes that organize task construction; the labels reflect intended difficulty drivers, not a factor-analytic decomposition of model capability.}

Based on these three principles, we present \textbf{\textit{SocialMaze}}, a benchmark designed to reflect the challenges posed by all three dimensions, as shown in \autoref{fig:pipeline}.
It consists of six diverse tasks spanning three categories: Social Reasoning Games, Daily Life Interactions, and Digital Community Platforms.
Each task intentionally varies the demands along the three dimensions. 
Our experiments with SocialMaze reveal several key insights into the capabilities of LLMs: Models with stronger chain-of-thought reasoning perform better on tasks that require deeper inference. We also observe that dynamic interaction affects performance in varied ways across tasks, and information uncertainty significantly hinders reasoning. Moreover, reasoning agents and workflows \citep{hu2024automated,zhang2024aflow,zhang2025multi} \rev{offer regime-dependent gains in these social reasoning challenges, helping weak Short-CoT backbones substantially but saturating on reasoning-capable backbones}, while targeted fine-tuning on curated reasoning examples leads to substantial performance improvements.

Our contributions are threefold:
1) \rev{We identify \textit{deep reasoning, interaction dynamics, and information uncertainty} as three descriptive design axes for constructing evaluation tasks (\autoref{subsec:limitation_quantification} for the construct-scope discussion).}
2) Based on these dimensions, we construct SocialMaze, a benchmark dataset comprising six tasks across three real-world-inspired scenarios (social games, daily life interactions, and digital platforms), covering a wide range of reasoning types and difficulty levels.
3) Our experiments reveal key limitations in LLM social reasoning, with targeted fine-tuning \rev{boosting structured-reasoning tasks but yielding inconclusive transfer to language-aggregation tasks (\S4.4)}.


\begin{table*}[t]
  \centering
  \small
  \setlength{\tabcolsep}{6pt}
  \renewcommand{\arraystretch}{1.15}
  \caption{Overview of SocialMaze task categories and key characteristics. Tasks vary along three key dimensions: the level of \textit{Deep Reasoning}, the degree of \textit{Dynamic Interaction}, and the extent of \textit{Information Uncertainty}, each categorized as High or Low.}
  \label{tab:task_overview}
  \vspace{-0.5em}
  \resizebox{\textwidth}{!}{%
  \begin{tabular}{@{}l l p{7.4cm} c c c r@{}}
    \toprule[1pt]
    \textbf{Scenario} & \textbf{Task} & \textbf{Description} & \textbf{\makecell{Deep\\Reasoning}} & \textbf{\makecell{Dynamic\\Interaction}} & \textbf{\makecell{Information\\Uncertainty}} & \textbf{Instances} \\
    \midrule
    \multirow{2}{*}{\makecell[l]{Social Reasoning\\Games}}
        & Hidden Role Deduction & Identify the Criminal and infer one's own role under Rumormongers and Lunatics across multiple rounds. & High & High & High & 20,000 \\
        & Find the Spy & Detect the odd word-receiver from $n{-}1$ Civilians' descriptions over $T$ rounds. & Low & High & Low & 6,000 \\
    \cmidrule(lr){2-7}
    \multirow{2}{*}{\makecell[l]{Daily Life\\Interactions}}
        & Rating Estimation from Text & Predict 1--5 star product rating from a mix of genuine and biased textual reviews. & Low & Low & High & 6,000 \\
        & Social Graph Analysis & Reason over transitive friendship graphs (group membership, friend counts, relationship queries). & High & Low & Low & 20,000 \\
    \cmidrule(lr){2-7}
    \multirow{2}{*}{\makecell[l]{Digital Community\\Platforms}}
        & Review Decision Prediction & Predict paper accept/reject across three review stages (info, reviews, rebuttal). & Low & High & Low & 12,000 \\
        & User Profile Inference & Recover age $\times$ gender (9-class) from LLM-generated review text. & Low & Low & High & 6,000 \\
    \bottomrule[1pt]
  \end{tabular}%
  }
  \vspace{-1em}
\end{table*}

\section{Problem Formulation}
\label{sec:problem_formulation}
We present a graph-based formalization of SocialMaze, where social entities and their evolving interactions are modeled as layered graphs.

\textbf{Modeling Social Entities and Interactions as Graph Structures.}
We use graph structures to formally represent the participants and interactions within a social scenario. Let $\mathcal{S} = \{s_1, s_2, ..., s_n\}$ be the set of social members involved in a given setting, where each $s_i$ represents a distinct individual, such as a game player, a forum user, or a reviewer in a peer-review process. These social members form the vertex set $\mathcal{V}$ of a graph $G = (\mathcal{V}, \mathcal{E})$, with $\mathcal{V}$ serving as an abstract representation of the social members $\mathcal{S}$.
Social interactions between members are represented as edges in the graph. Since interactions are time-dependent, we define a separate edge set $\mathcal{E}_t$ for each interaction round $t$. This leads to a sequence of time-specific graphs $G_t = (\mathcal{V}, \mathcal{E}_t)$, where an edge $(u, v) \in \mathcal{E}_t$ indicates that members $u$ and $v$ interacted during round $t$.
The nature of the edges (directed or undirected) reflects the type of interaction. For example, directed edges may represent one-way actions (e.g., sending a message), while undirected edges may represent mutual interactions (e.g., a conversation or vote). 

\textbf{Temporal Dynamics as Layered Graphs.}
Social interactions are inherently dynamic and typically unfold over multiple rounds. To capture this temporal dimension, we represent the entire interaction process as a layered graph $\mathcal{G} = (G_1, G_2, ..., G_T)$, where $T$ denotes the total number of interaction rounds. Each layer $G_t = (\mathcal{V}, \mathcal{E}_t)$ captures the state of social members and their relationships during round $t$.
Importantly, all layers share the same vertex set $\mathcal{V}$, reflecting a consistent group of participants throughout the interaction. However, the edge sets $\mathcal{E}_t$ vary across layers to reflect the evolving nature of relationships over time.
In SocialMaze, LLMs receive natural language descriptions that encapsulate the information from these layered graphs, rather than raw graph structures. This design choice is intentional, aiming to mimic how humans comprehend social scenarios through language-based narratives.
Recent graph-reasoning benchmarks show that LLM performance is sensitive to graph representation and can benefit from structured decomposition \citep{xu2025gta}, while LLM-guided knowledge-graph traversal combines exploration with replay memory to recover multi-hop evidence efficiently \citep{NEURIPS2025_4d880ba0}; SocialMaze instead uses natural-language interaction histories to test inference over socially grounded graph structure.

\textbf{Query Categorization. }
Based on the layered graph representation, we classify the queries posed within SocialMaze tasks into three distinct types, each targeting a different level of understanding of the graph structure:
\textbf{\textit{Vertex-centric Query ($\mathcal{Q}_v(v_i)$):}} This type of query probes the model's understanding of individual social members. Given a specific vertex $v_i \in \mathcal{V}$ (representing social member $s_i$), the task is to infer an attribute associated with $v_i$.
\textbf{\textit{Edge-centric Query ($\mathcal{Q}_e(v_i, v_j)$):}}  Edge-centric queries assess the model's comprehension of the relationships between social members. Given two vertices $v_i, v_j \in \mathcal{V}$, the task is to determine the nature of their relationship, as represented by the edges connecting them.
\textbf{\textit{Graph-level Query ($\mathcal{Q}_G(\mathcal{G})$):}} Graph-level queries require the model to synthesize information from the entire layered graph $\mathcal{G}$ to derive a holistic understanding of the social scenario.  These queries demand a comprehensive assessment of the overall interaction dynamics. 

\section{SocialMaze}

As formalized in \autoref{sec:problem_formulation}, and building on the layered social interaction graph framework, we introduce \textbf{SocialMaze}, a benchmark designed to operationalize the core challenges of social reasoning—\textit{Dynamic Interaction}, \textit{Information Uncertainty}, and \textit{Deep Reasoning}. 
The benchmark spans three representative social contexts: Social Reasoning Games, Daily Life Interactions, and Digital Community Platforms. 
These settings comprise six major task categories, each carefully designed to vary along the key dimensions, enabling systematic evaluation of LLMs under diverse social conditions. 
\autoref{tab:task_overview} summarizes the tasks by required reasoning depth, degree of interaction dynamics, and level of information uncertainty. 
A detailed comparison between SocialMaze and prior benchmarks is provided in \autoref{appendix:related_works}.

\subsection{Hidden Role Deduction}
\label{subsec:task1}
This task abstracts the core mechanics of various social reasoning games, such as \textit{Blood on the Clocktower} \citep{bloodontheclocktower}, \textit{The Resistance: Avalon} \citep{Avalon} and \textit{Werewolf} \citep{Mafia}, into a reasoning-only format. Unlike traditional interaction-based gameplay, where players engage in direct communication, all player statements are rule-generated. The model acts as a reasoner, analyzing all available information to logically infer each player's role.

\textbf{Task Rules.}
The game features four roles: Investigators, Criminal, Rumormongers, and Lunatics. Investigators always tell the truth. The Criminal can choose to lie or tell the truth. Rumormongers believe they are Investigators, but their statements are randomly true or false. Lunatics believe they are the Criminal. The role each player sees may not reflect their true identity—Rumormongers are shown the role of Investigator, and Lunatics are shown the role of Criminal.

The game consists of $n$ players, and the model participates from the perspective of Player 1 ($s_1$), meaning that it observes only what that player would see and say during the $T$ rounds.
In each round, every player selects another player and makes a public statement, such as “Player $v$ says Player $u$ is (not) the criminal.”
After observing all interactions, the model is tasked with answering two key questions: identifying the true Criminal ($\mathcal{Q}_G$), and inferring its own actual role in the game ($\mathcal{Q}_v(v_i)$). The introduction of Rumormongers and Lunatics significantly increases \textit{information uncertainty} and makes the reasoning process more challenging. The dataset includes four types of tasks: Original task, Rumormonger task, Lunatic task, and Full task. Details are provided in \autoref{appendix:roleplaying_rules}.

\textbf{Design Rationale.}
This task challenges large language models along three critical dimensions of social reasoning.
First, the game unfolds over multiple rounds, requiring the model to track the temporal evolution of information, interpret changing relationships among players, and maintain consistent judgments across rounds—posing a challenge of \textit{Dynamic Interaction}.
Second, due to the presence of roles such as the Criminal, Rumormonger, and Lunatic—who may lie or provide misleading information—the environment is \textit{highly uncertain}. The model must determine which statements are trustworthy and filter out deceptive cues, thereby grappling with information uncertainty.
Most importantly, the model must reason not only about others’ roles but also about its own true identity, which may differ from the one initially assigned. Addressing this requires strong \textit{Deep Reasoning} capabilities, including resolving conflicts, managing uncertainty that extends to self-perception, and dynamically updating internal beliefs to approach the ground truth.

\textbf{Data Generation and Quality Assurance.}
All player statements are automatically generated based on a set of predefined rules. Investigators begin by selecting a target they find suspicious, using a strategy function informed by all interactions up to the current round. They are always truthful in their statements. Rumormongers follow the same target selection logic as Investigators, but the truthfulness of their statements is random, making their input unreliable. In contrast, Criminals and Lunatics adopt a different strategy for choosing targets and deliberately introduce uncertainty by making deceptive statements with a certain probability, aiming to mislead others and conceal their true roles. To ensure each scenario is logically solvable, we design a search algorithm that verifies whether a unique solution exists to identify both the true Criminal and the LLM player's actual role. Additionally, the full reasoning chain leading to the solution is preserved and distilled into clear natural language, providing high-quality, curated examples of social reasoning that can be leveraged for targeted fine-tuning of language models. See \autoref{appendix:roleplaying_rules} for details.

\subsection{Find the Spy}
\label{subsec:task2}

This task is inspired by the classic word-based social deduction game \textit{Who Is The Spy} \citep{wei2025exploring} and aims to evaluate the LLM's ability to identify subtle deviations in communication within a group setting. The game involves $n$ players, where $n-1$ players (Civilians) receive the same secret word, and one player (the Spy) receives a related but different word. Over $T$ rounds, each player provides a description of their word. The LLM, observing the descriptions, must determine which player is the Spy. This task is designed to assess the model's capability to track and interpret information through interactions while minimizing the ambiguity of the presented information. The high level of interaction comes from the multi-round format, requiring the model to progressively piece together and evaluate each player's description. The low level of information uncertainty is maintained because, although the players are incentivized to be truthful, the task still involves the subtle challenge of discerning the Spy based on differences in description style and detail. Detailed task rules, the data generation process, and quality assurance procedures are provided in \autoref{appendix:whoisspy_rules}.

\subsection{Rating Estimation from Text}
\label{subsec:task3}

This task evaluates the ability of LLMs to predict a product's star rating (from 1 to 5) based on a set of $n$ textual reviews. These reviews may include genuine user feedback as well as promotional reviews written by shills. The task is designed to challenge the model's ability to deal with information uncertainty, where the reviews might include biased or misleading content, requiring the model to evaluate the credibility of each review. The LLM is tasked with estimating a single star rating for the product, based on the aggregated reviews from multiple users. The input is structured as a vertex-centric query, where the model must synthesize information from user reviews to predict the most likely rating for the product. Detailed task rules, the data generation process and quality assurance procedures are provided in \autoref{appendix:ecommerce_rules}.

\subsection{Social Graph Analysis} 
\label{subsec:task4}
This task aims to evaluate the ability of LLMs to analyze relationships within a social group. Given a description of the social network graph and pairwise relationship labels indicating whether two individuals are friends or have a bad relationship—with friendship being transitive—the model is required to perform reasoning such as: determining whether two individuals are friends, identifying the friend group of a given node, calculating the total number of distinct friend groups, and counting all relationships within the network. Detailed task rules, the data generation process that ensures logical consistency and solvability, and quality assurance procedures are provided in \autoref{appendix:interpersonal_rules}.

\subsection{Review Decision Prediction}
\label{subsec:task5}
This task aims to evaluate the ability of LLMs to predict the final acceptance outcome (Accepted/Rejected) of a research paper as they gradually receive more information throughout the academic review process. The model is required to make a prediction at each of three interaction stages, with the available context incrementally expanding: in the first stage, only the initial paper information is provided; in the second stage, reviewer comments (with numerical scores removed) are added and the model must reason over both the initial content and the reviews; in the third stage, the full author rebuttal is introduced, completing the review context. This task simulates how opinions evolve over time in real academic peer review. Detailed task rules, the data generation process using real-world OpenReview data, and quality assurance procedures are provided in \autoref{appendix:peerreview_rules}.

\subsection{User Profile Inference}
\label{subsec:task6}
This task aims to evaluate the ability of LLMs to recover demographic cues (age group and gender) that the generator pool has encoded into user-generated textual reviews. Specifically, we construct a large number of synthetic users with known demographic attributes using LLMs, and generate their reviews for various products they have purchased; the evaluated LLM then attempts to recover those demographic attributes from the resulting text. The inference tasks are twofold: (1) predicting the dominant user profile associated with the reviews of a specific product, and (2) identifying the profile of an individual user based on their reviews across multiple products.

\rev{We score predictions as joint age $\times$ gender 9-class accuracy on the 500-instance set (random-guess floor $1/9 \approx 11.1\%$). Because the demographic ground truth is the generator's input prompt rather than a real user identity, this task probes evaluator-vs-generator-pool stereotype alignment rather than inference about real demographic populations; see \autoref{subsec:realism_transfer}.}

Detailed task rules, the LLM-based data generation method used to embed subtle demographic cues, and human validation results are provided in \autoref{appendix:userprofile_rules}.

\section{Experiments and Analysis}
We conduct extensive experiments on SocialMaze to evaluate the social reasoning capabilities of LLMs. Specifically, we test \rev{five} proprietary LLMs and \rev{seven} open-weight LLMs across the six tasks. We additionally evaluate six reasoning workflows to assess their impact on model performance.

\rev{A condensed overview of model performance across the six tasks is reported in \autoref{tab:overall_summary}.} We observe that different social reasoning tasks impose distinct demands on the models. For example, tasks like \textit{Hidden Role Deduction}, which require \textit{Deep Reasoning}, are best tackled by models such as DeepSeek-R1 and Gemini-2.5-Pro. In contrast, DeepSeek-V3 and GPT-4o excel on tasks like \textit{Review Decision Prediction}, where nuanced understanding of reviewer attitudes is critical. We conducted extensive case studies and report full results and settings in \autoref{appendix:case_study}, \autoref{experient_details}, and \autoref{sec:supplementary_discussion}.

\subsection{The Impact of Deep Reasoning}
\label{subsec:results_deep_reasoning}
We compare Long CoT models (e.g., o1, DeepSeek-R1) against the remaining Short CoT models, both using identical prompts, on the two \textit{Deep Reasoning} tasks (\textit{Hidden Role Deduction}, \textit{Social Graph Analysis}) and the four shallower tasks listed in \autoref{tab:task_overview}.

\begin{figure}[htb]
    \centering
    \includegraphics[width=\linewidth]{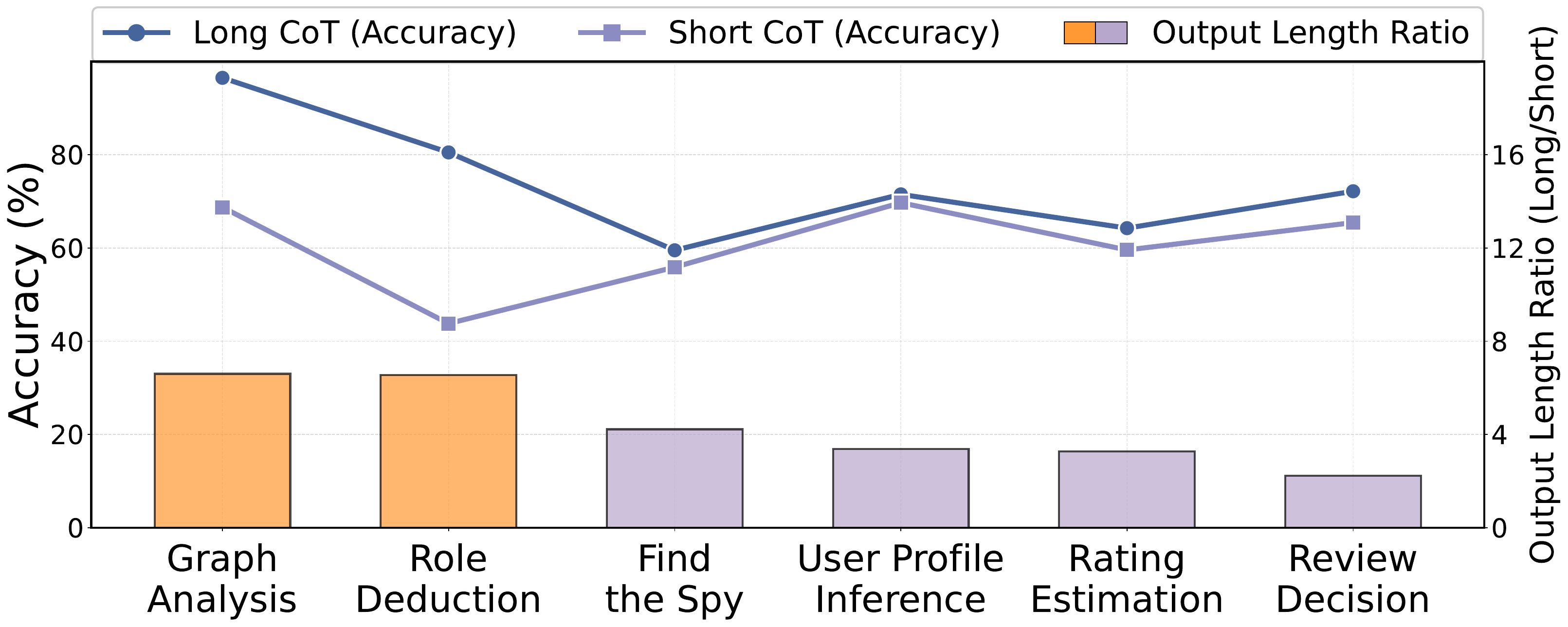}
    \caption{Performance comparison of Long vs Short CoT models. The line plot shows average accuracy; the bar plot shows the output length ratio (Long / Short). Orange bars indicate tasks with high deep reasoning demand, purple bars indicate low deep reasoning demand.}
    \label{fig:deep_reason}
\end{figure}

\textbf{Long CoT models achieve substantially higher accuracy on tasks requiring \textit{Deep Reasoning}.} \autoref{fig:deep_reason} (line plot) shows the gap is pronounced on graph analysis and role deduction and narrows on shallower tasks.

\rev{A pilot judge evaluation on four Short-CoT models (\autoref{subsec:faithfulness}) finds correctly-answered chains internally consistent and complete while incorrect-answer chains score substantially lower; the analogous Long-CoT evaluation (3{,}000--4{,}200 tokens, different judge protocol) is left to future work. A budget-matched control rules out raw output-cap as the Long-vs-Short driver (\autoref{subsec:model_centric}).}

\textbf{The improved performance comes with substantial computational cost.} The bar plot in \autoref{fig:deep_reason} shows Long CoT models use nearly eight times more output tokens on average on \textit{Deep Reasoning} tasks, narrowing on shallower tasks where the accuracy gains are smaller.

\begin{figure*}[t]
    \centering
    \vspace{-0em}
    \includegraphics[width=\linewidth]{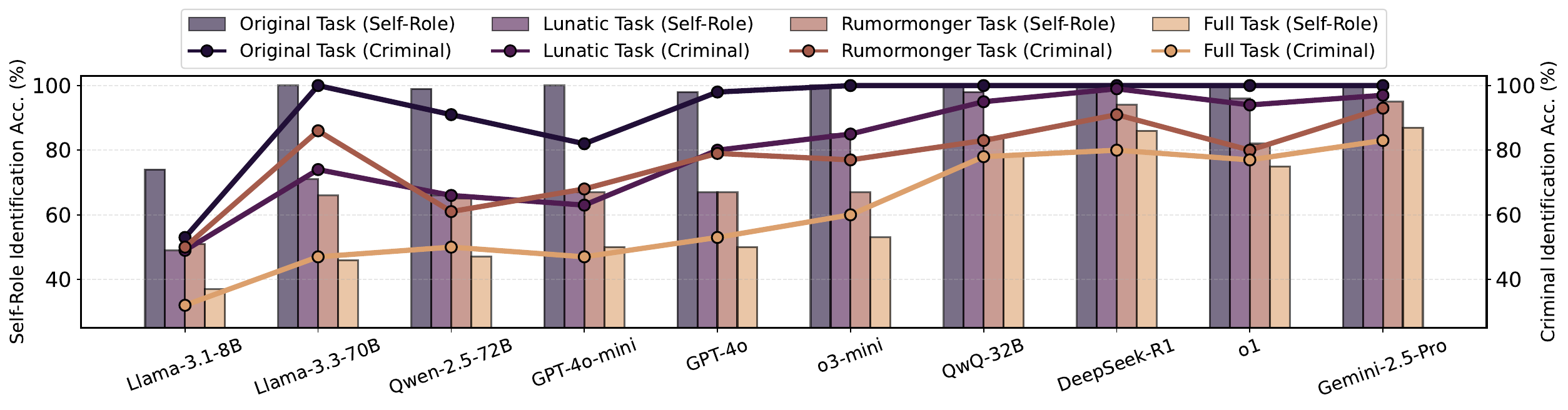}
    \vspace{-2em}
    \caption{Model performance in \textit{Hidden Role Deduction} across four task variants with increasing information uncertainty. Accuracy is shown after 3 rounds.}
    \label{fig:task_accuracy}
    \vspace{-1em}
\end{figure*}

\subsection{The Impact of Dynamic Interaction}
\label{subsec:results_dynamic_interaction}

In certain scenarios, social interactions unfold sequentially, requiring models to integrate and reason over information accumulated across multiple turns. We analyze how model performance evolves in tasks with high \textit{Dynamic Interaction}: \textbf{model accuracy generally improves with quantitatively increasing interaction, but the trajectory of performance evolution and sensitivity to dynamic information vary significantly across tasks and models.}

\begin{table}[htb]
\small
\centering
\setlength{\tabcolsep}{6pt}
\renewcommand{\arraystretch}{1.15}
\caption{\rev{Criminal identification accuracy across rounds in the 6-player \textit{Hidden Role Deduction} task; cells report 5-seed mean $\pm$ 95\% binomial CI half-width at $n=500$ per round. Models vary in leveraging \textit{Dynamic Interaction}.}}
\label{tab:model_rounds}
\vspace{-0.5em}
\rowcolors{2}{tabbluerowa}{tabbluerowb}
\resizebox{\columnwidth}{!}{%
\begin{tabular}{lccc}
\rowcolor{tabbluehead}
\textcolor{white}{\textbf{Model}} & \textcolor{white}{\textbf{Round 1}} & \textcolor{white}{\textbf{Round 2}} & \textcolor{white}{\textbf{Round 3}} \\
\toprule
\rev{Llama-3.1-8B}    & \rev{$31.4 \pm 4.1$}  & \rev{$38.6 \pm 4.3$}  & \rev{$41.2 \pm 4.3$} \\
\rev{Phi-4}           & \rev{$32.8 \pm 4.1$}  & \rev{$40.4 \pm 4.3$}  & \rev{$45.6 \pm 4.4$} \\
Llama-3.3-70B   & \rev{$37.6 \pm 4.2$}  & \rev{$46.7 \pm 4.4$}  & \rev{$46.5 \pm 4.4$} \\
Qwen-2.5-72B    & \rev{$31.3 \pm 4.1$}  & \rev{$42.6 \pm 4.3$}  & \rev{$50.3 \pm 4.4$} \\
GPT-4o-mini     & \rev{$33.5 \pm 4.1$}  & \rev{$38.4 \pm 4.3$}  & \rev{$46.5 \pm 4.4$} \\
GPT-4o          & \rev{$39.5 \pm 4.3$}  & \rev{$53.3 \pm 4.4$}  & \rev{$53.5 \pm 4.4$} \\
o3-mini         & \rev{$\mathbf{45.8} \pm 4.4$}  & \rev{$56.4 \pm 4.4$}  & \rev{$65.4 \pm 4.2$} \\
QwQ-32B         & \rev{$41.4 \pm 4.3$}  & \rev{$63.5 \pm 4.2$}  & \rev{$78.4 \pm 3.6$} \\
\rev{DeepSeek-V3}     & \rev{$22.4 \pm 3.7$}  & \rev{$33.6 \pm 4.1$}  & \rev{$45.0 \pm 4.4$} \\
DeepSeek-R1     & \rev{$44.3 \pm 4.4$}  & \rev{$72.3 \pm 3.9$}  & \rev{$80.4 \pm 3.5$} \\
o1              & \rev{$42.5 \pm 4.4$}  & \rev{$67.5 \pm 4.1$}  & \rev{$76.6 \pm 3.7$} \\
Gemini-2.5-Pro  & \rev{$43.3 \pm 4.4$}  & \rev{$\mathbf{74.3} \pm 3.8$}  & \rev{$\mathbf{87.6} \pm 2.9$} \\
\bottomrule
\end{tabular}%
}
\vspace{-1em}
\end{table}
This contrast is particularly evident in two tasks. In \textit{Hidden Role Deduction}, we track model accuracy in identifying the Criminal within the Full Task setting involving six players (\autoref{tab:model_rounds}). \textbf{Accuracy tends to increase as more rounds of interaction are observed}, reflecting the expected benefit of accumulating contextual evidence over time. However, the slope of improvement varies substantially across models, suggesting differing abilities to process and integrate newly revealed statements within the game's evolving context. Some models are more effective than others in leveraging additional rounds to refine their hypotheses. One notable insight is the large performance gap between Rumormonger and Lunatic. This is because once an Investigator correctly checks the Lunatic as “not the Criminal,” it provides a strong signal that helps the Lunatic realize their true role—making awakening easier than for the Rumormonger. \rev{Round-1 accuracy on \autoref{tab:model_rounds} is jointly determined by the model's exploitation of the prompt-supplied role label and role-distribution enumeration (which narrows the Criminal-candidate set) and the single round of public statements; the Round-1-to-Round-3 lift should therefore be read as ``rounds-accumulated reasoning over the Round-1 prior baseline'' rather than ``pure evidence accumulation from zero''.}

By contrast, \textit{Review Decision Prediction} presents a non-linear performance trajectory across the stages of the peer review process, as shown in \autoref{tab:multiturn_debate}. \rev{The three stages provide title+abstract+keywords (Info), reviewer comments with scores removed (Reviews), and the author rebuttal (Rebuttal); cells report 5-seed mean accuracy with 95\% binomial CI half-width at $n=500$ on a 67\% reject / 33\% accept test set.}

\rev{Round-1 accuracy falls below the 67\% reject-majority baseline for all evaluated models, indicating an over-accept anti-prior. The Round-2 to Round-3 trajectory splits into ten down-trending models and two up-trending evaluators (GPT-4o, DeepSeek-V3); the up-trends do not survive Bonferroni correction at the binomial CI floor, and the perturbed-subset analysis in \autoref{tab:rdp_perturbed} shows the Round-3 differences are not driven by OpenReview-text memorization. See \autoref{appendix:peerreview_rules} blocks 5--6.}

We observe that \rev{the Round-3 down-trend on the ten-evaluator majority} is frequently driven by the model being swayed by the author's sincere defense, which may not align with the judgment of human reviewers or area chairs. A more detailed analysis is in \autoref{appendix:case_study}.

\begin{figure*}[t]
    \centering
    \vspace{-1em}
    \includegraphics[width=\linewidth]{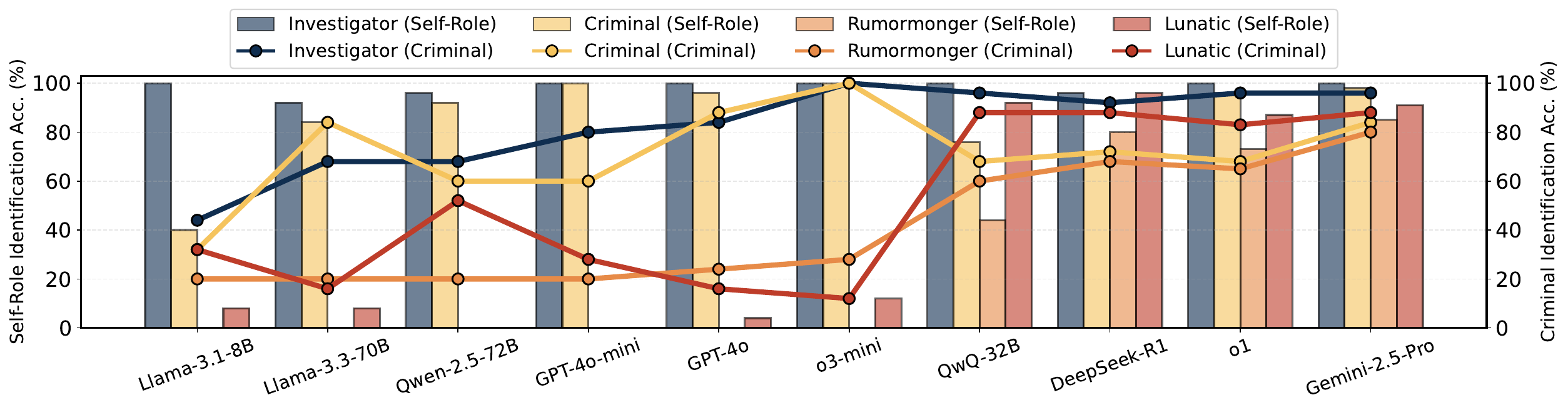} 
    \vspace{-2em}
    \caption{\rev{Full-task accuracy in \textit{Hidden Role Deduction} by model-assigned role. Each bar is the 5-seed bootstrap mean over $n=125$ per role with per-bin 95\% half-width $\approx \pm 3$ to $4$ pp (exact values in \autoref{tab:per_role_decomposition}). Investigator and Criminal bars have displayed role equal to true role (trivial echo of the prompt label); Rumormonger and Lunatic bars are the override-required regime where the model must revise the prompt label under indirect evidence.}}
    \label{fig:role_accuracy}
    \vspace{-1em}
\end{figure*}

\subsection{The Impact of Information Uncertainty}
\label{subsec:results_information_uncertainty}

\begin{table}[htb]
\small
\centering
\setlength{\tabcolsep}{6pt}
\renewcommand{\arraystretch}{1.15}
\caption{\textit{Review Decision Prediction} accuracy across the three sequential stages (Info / Reviews / Rebuttal). \rev{See §4.2 narrative and \autoref{appendix:peerreview_rules} for column definitions, test-set composition, and trajectory discussion.}}
\label{tab:multiturn_debate}
\vspace{-0.5em}
\rowcolors{2}{tablavrowa}{tablavrowb}
\resizebox{\columnwidth}{!}{%
\begin{tabular}{lccc}
\rowcolor{tablavhead}
\textcolor{white}{\textbf{Model}} & \textcolor{white}{\textbf{Info.}} & \textcolor{white}{\textbf{Reviews}} & \textcolor{white}{\textbf{Rebuttal}} \\
\toprule
Llama-3.1-8B    & \rev{$37.0 \pm 4.2$}  & \rev{$79.6 \pm 3.5$} & \rev{$62.0 \pm 4.3$} \\
Llama-3.3-70B   & \rev{$26.2 \pm 3.9$}  & \rev{$87.4 \pm 2.9$} & \rev{$72.2 \pm 3.9$} \\
Qwen-2.5-72B    & \rev{$23.6 \pm 3.7$}  & \rev{$82.2 \pm 3.4$} & \rev{$65.8 \pm 4.2$} \\
Phi-4           & \rev{$23.5 \pm 3.7$}  & \rev{$77.6 \pm 3.6$} & \rev{$61.4 \pm 4.3$} \\
GPT-4o-mini     & \rev{$38.8 \pm 4.3$}  & \rev{$85.2 \pm 3.1$} & \rev{$85.0 \pm 3.1$} \\
GPT-4o          & \rev{$\mathbf{55.3} \pm 4.4$} & \rev{$86.2 \pm 3.0$} & \rev{$\mathbf{90.2} \pm 2.6$} \\
o3-mini         & \rev{$40.2 \pm 4.3$}  & \rev{$86.4 \pm 3.0$} & \rev{$78.6 \pm 3.6$} \\
QwQ-32B         & \rev{$49.8 \pm 4.4$}  & \rev{$83.8 \pm 3.2$} & \rev{$79.6 \pm 3.5$} \\
\rev{DeepSeek-V3}     & \rev{$38.2 \pm 4.3$}  & \rev{$86.0 \pm 3.0$} & \rev{$\mathbf{91.4} \pm 2.5$} \\
DeepSeek-R1     & \rev{$50.2 \pm 4.4$}  & \rev{$88.0 \pm 2.9$} & \rev{$82.0 \pm 3.4$} \\
o1              & \rev{$52.2 \pm 4.4$}  & \rev{$\mathbf{88.6} \pm 2.8$} & \rev{$78.2 \pm 3.6$} \\
Gemini-2.5-Pro  & \rev{$47.4 \pm 4.4$}  & \rev{$87.6 \pm 2.9$} & \rev{$77.6 \pm 3.7$} \\
\bottomrule
\end{tabular}%
}
\vspace{-1em}
\end{table}

A defining characteristic of complex social environments is the prevalence of unreliable information. We evaluate the impact of Information Uncertainty using the \textit{Hidden Role Deduction} task by systematically introducing actors who generate distinct forms of unreliable information: intentional deception (Criminal) and noise stemming from flawed self-perception (Rumormonger and Lunatic).

\textbf{Increased information uncertainty significantly elevates the difficulty of social reasoning for LLMs.} \autoref{fig:task_accuracy} illustrates this across four task configurations where uncertainty levels are quantitatively controlled by varying the number of unreliable actors (Rumormongers, Lunatics). Progressing from the baseline Original setting through scenarios introducing these noise sources to the complex Full setting, both the accuracy in identifying the Criminal (line plot) and the model's own role (bar plot) demonstrate a marked decline. This degradation underscores the substantial challenge posed by noise and deception.

\textbf{Reasoning becomes particularly challenging when the model's own perceived role or information source is compromised.} \autoref{fig:role_accuracy} delves into the Full task configuration, analyzing final accuracy based on the specific role assigned to the LLM.

Models exhibit considerably lower accuracy, especially in identifying their own role (bars), when assigned as a Rumormonger or Lunatic compared to being an Investigator or Criminal.
This suggests a significant difficulty in reconciling internal beliefs (e.g., ``I think I am an Investigator/Criminal'') with conflicting external evidence or the unreliable nature of one's own information.

These experiments also highlight critical differences between model capabilities. \textbf{Existing Short CoT models demonstrate severe limitations in handling complex scenarios with high information uncertainty}, often failing to perform reliably in the Rumormonger and Full task settings (\autoref{fig:task_accuracy}). In contrast, Long CoT models, while still impacted, exhibit significantly better resilience to uncertainty. Furthermore, the challenge of self-assessment under uncertainty exposes a stark gap: \textbf{Short CoT models are almost entirely unable to deduce their true identity when assigned as a Rumormonger or Lunatic} (\autoref{fig:role_accuracy}), suggesting a profound lack of capacity for self-doubt and meta-reasoning necessary to overcome compromised initial information.

\rev{Part of this gap reflects indirect-evidence availability rather than capability alone (\autoref{tab:lunatic_solvability}).}

\subsection{Enhancing Social Reasoning Capabilities}
\label{subsec:enhancing_capabilities}
To explore strategies for improving social reasoning in LLMs, we conduct focused experiments on the \textit{Hidden Role Deduction} task. This task is not only uniquely representative—combining Deep Reasoning, Dynamic Interaction, and Information Uncertainty—but also well-suited for generating diverse reasoning examples at scale, providing valuable supervision for model learning.

\rev{\textbf{Reasoning agents and workflows produce regime-dependent gains for social reasoning.}} Recent approaches improve complex reasoning through dynamic workflow adaptation \citep{wang2025dyflow}, adaptive allocation of reasoning effort \citep{NEURIPS2025_6dc129f0}, context-oriented reasoning synthesis \citep{jin2026contextcot}, and structured context self-auditing \citep{jin2026contextguard}. Agent systems additionally use feedback-aware multi-agent orchestration \citep{song2025quite}, co-active preference learning \citep{xu-etal-2026-coact}, and execution-grounded trajectories \citep{wu2026terminaltraj}. We assess whether such task decomposition and planning mechanisms can improve performance on \textit{Hidden Role Deduction}. \rev{Six workflow methods on the weak Short-CoT backbone (\autoref{tab:workflow}) yield Both-Correct lifts of $+4$ to $+8$~pp, with the largest gains (DyFlow, MaAS, AFlow) clearly above the binomial $\Delta$-CI of $\approx \pm 6$ pp at $p \approx 0.30$ and the smaller gains (Self-Refine, LLM-Debate) inside it; on the DeepSeek-R1 backbone all four methods land within $\pm 2.0$ pp of the Base, indicating saturation. See \autoref{subsec:workflow_dsr1} for per-method $\Delta$ and noise-floor analysis.}

\begin{table*}[t]
\centering
\caption{Overall Accuracy (\%) on SocialMaze Tasks. \rev{Each cell reports the mean accuracy and 95\% binomial CI half-width at $n=500$ over five independent runs at temperature 0.7. The HRD column averages over the 1:1:1:1 role mixture (Investigator / Criminal / Rumormonger / Lunatic); the per-role decomposition is reported in \autoref{tab:per_role_decomposition} and shows that the Both-Correct headline aggregates a regime with displayed=true self-role (Investigator, Criminal) and a regime with displayed$\neq$true self-role (Rumormonger, Lunatic).} \rev{UPI: joint age $\times$ gender 9-class accuracy (floor $1/9 \approx 11.1\%$); see \autoref{appendix:userprofile_rules} and \autoref{subsec:realism_transfer} for the construct-scope note that UPI scores on the FtS / Rating / UPI columns are confounded with within-family alignment to the generator pool. Rating Estimation: constant-predict-majority baseline = 73\% (see \autoref{subsec:rating_distribution}); cells below this floor indicate anti-calibration toward minority classes.} See text for further metric details.}
\label{tab:overall_summary}
\small
\renewcommand{\arraystretch}{1.1}
\setlength{\tabcolsep}{4pt}
\resizebox{\textwidth}{!}{%
\begin{tabular}{@{}lcccccc@{}}
\toprule
\textbf{Model} & \textbf{\makecell{Hidden Role \\ Deduction}} & \textbf{\makecell{Find the \\ Spy}} & \textbf{\makecell{Rating \\ Estimation}} & \textbf{\makecell{Social Graph \\ Analysis}} & \textbf{\makecell{Review \\ Decision \\ Prediction}} & \textbf{\makecell{User Profile \\ Inference}} \\
\midrule
Llama-3.1-8B    & \rev{$15.4 \pm 3.2$} & \rev{$37.2 \pm 4.2$} & \rev{$57.2 \pm 4.3$} & \rev{$28.2 \pm 3.9$} & \rev{$62.0 \pm 4.3$} & \rev{$60.2 \pm 4.3$} \\
Llama-3.3-70B   & \rev{$25.2 \pm 3.8$} & \rev{$60.0 \pm 4.3$} & \rev{$74.8 \pm 3.8$} & \rev{$81.0 \pm 3.4$} & \rev{$72.2 \pm 3.9$} & \rev{$\mathbf{78.6} \pm 3.6$} \\
Phi-4           & \rev{$20.4 \pm 3.5$} & \rev{$45.2 \pm 4.4$} & \rev{$60.4 \pm 4.3$} & \rev{$40.6 \pm 4.3$} & \rev{$61.4 \pm 4.3$} & \rev{$62.4 \pm 4.2$} \\
Qwen-2.5-72B    & \rev{$22.6 \pm 3.7$} & \rev{$48.9 \pm 4.4$} & \rev{$72.2 \pm 3.9$} & \rev{$80.6 \pm 3.5$} & \rev{$65.8 \pm 4.2$} & \rev{$68.0 \pm 4.1$} \\
QwQ-32B         & \rev{$52.0 \pm 4.4$} & \rev{$50.2 \pm 4.4$} & \rev{$74.4 \pm 3.8$} & \rev{$95.0 \pm 1.9$} & \rev{$79.6 \pm 3.5$} & \rev{$72.2 \pm 3.9$} \\
GPT-4o-mini     & \rev{$26.4 \pm 3.9$} & \rev{$61.2 \pm 4.3$} & \rev{$75.8 \pm 3.8$} & \rev{$53.0 \pm 4.4$} & \rev{$\mathbf{85.0} \pm 3.1$} & \rev{$74.4 \pm 3.8$} \\
GPT-4o          & \rev{$28.8 \pm 4.0$} & \rev{$69.2 \pm 4.0$} & \rev{$\mathbf{76.0} \pm 3.7$} & \rev{$83.2 \pm 3.3$} & \rev{$90.2 \pm 2.6$} & \rev{$\mathbf{79.2} \pm 3.6$} \\
o3-mini         & \rev{$55.2 \pm 4.4$} & \rev{$74.0 \pm 3.8$} & \rev{$71.2 \pm 4.0$} & \rev{$99.0 \pm 0.9$} & \rev{$78.6 \pm 3.6$} & \rev{$71.4 \pm 4.0$} \\
o1              & \rev{$60.8 \pm 4.3$} & \rev{$\mathbf{78.4} \pm 3.6$} & \rev{$\mathbf{76.2} \pm 3.7$} & \rev{$99.2 \pm 0.8$} & \rev{$78.2 \pm 3.6$} & \rev{$77.0 \pm 3.7$} \\
\rev{DeepSeek-V3}   & \rev{$30.2 \pm 4.0$} & \rev{$58.6 \pm 4.3$} & \rev{$75.8 \pm 3.8$} & \rev{$92.4 \pm 2.3$} & \rev{$\mathbf{91.4} \pm 2.5$} & \rev{$65.4 \pm 4.2$} \\
DeepSeek-R1     & \rev{$65.4 \pm 4.2$} & \rev{$70.2 \pm 4.0$} & \rev{$71.0 \pm 4.0$} & \rev{$98.6 \pm 1.0$} & \rev{$82.0 \pm 3.4$} & \rev{$74.6 \pm 3.8$} \\
Gemini-2.5-Pro  & \rev{$\mathbf{73.6} \pm 3.9$} & \rev{$76.6 \pm 3.7$} & \rev{$73.6 \pm 3.9$} & \rev{$\mathbf{100.0} \pm 0.4$} & \rev{$77.6 \pm 3.7$} & \rev{$73.0 \pm 3.9$} \\
\midrule
\textbf{Human (avg.)} & \rev{$64.2 \pm 4.2$} & \rev{$84.4 \pm 3.2$} & \rev{$75.2 \pm 3.8$} & - & \rev{$78.0 \pm 3.6$} & \rev{$73.9 \pm 3.9$} \\
\bottomrule
\end{tabular}%
}
\end{table*}

\textbf{Fine-tuning on curated reasoning traces substantially improves performance.} We apply Supervised Fine-Tuning (SFT) \citep{ouyang2022training} and Direct Preference Optimization (DPO) \citep{rafailov2023direct} to Llama-3.1-8B and Phi-4 on high-quality reasoning examples from our dataset. \autoref{tab:dpo} reports HRD gains, and \autoref{tab:dpo_cross_task} reports cross-task transfer on the five non-HRD tasks. \rev{Transfer to \textit{Social Graph Analysis} is robust ($+9.2$ to $+9.6$ pp under SFT, clearly above the $\pm 6$~pp paired-$\Delta$ CI); \textit{Find the Spy} is borderline ($+5.8$ / $+6.4$ pp under SFT) and partly consistent with sampling noise; the linguistic family (\textit{Rating Estimation}, \textit{Review Decision Prediction}, \textit{User Profile Inference}) sits inside the $\Delta$-CI at $+0.4$ to $+2.4$ pp. Thus, transfer to the structured-graph task is supported, transfer to Find the Spy is borderline, and transfer to the language-aggregation tasks remains undetermined (\autoref{subsec:dpo_cross_task}).}

\paragraph{Cross-task comparison.}
\autoref{tab:overall_summary} reveals two complementary regimes. Reasoning-oriented models form the strongest group on \textit{Hidden Role Deduction}, while Gemini-2.5-Pro, o1, o3-mini, and DeepSeek-R1 all exceed 98\% on \textit{Social Graph Analysis}. Conversely, language-aggregation and judgment tasks favor strong generalists: DeepSeek-V3 and GPT-4o lead \textit{Review Decision Prediction}, the leading \textit{Rating Estimation} results occupy a narrow 74.4--76.2\% band, and GPT-4o, Llama-3.3-70B, and o1 are similarly close on \textit{User Profile Inference}. \textit{Find the Spy} lies between these regimes, led by o1, Gemini-2.5-Pro, o3-mini, and DeepSeek-R1 among the evaluated models.

These task-dependent reversals are more informative than a single aggregate ranking. Most within-tier gaps are comparable to the $n=500$ binomial half-width and should not be read as ordered separations; detailed family-wise comparisons appear in \autoref{sec:supplementary_discussion}. The human baseline likewise varies by task, exceeding every model on \textit{Find the Spy} while remaining competitive on \textit{Hidden Role Deduction}. Because the HRD headline mixes prompt-label echo with diagnostic role-override cases, \autoref{tab:per_role_decomposition} provides the more direct measure of belief revision.

\section{Conclusion}

We introduced SocialMaze, a benchmark designed to rigorously evaluate the social reasoning capabilities of LLMs by capturing the challenges of deep reasoning, dynamic interaction, and information uncertainty. Experiments across six diverse tasks reveal notable weaknesses in current models, particularly in handling evolving contexts and reasoning under uncertainty. \rev{Targeted fine-tuning on curated reasoning examples significantly improves performance on structured social-reasoning tasks; transfer to language-aggregation tasks is inconclusive at this sample size (\S4.4), highlighting the scope as well as the value of domain-specific adaptation.}

As future work, we plan to expand SocialMaze with more real-world data and new social scenarios and task types.

\section*{Limitations}
\label{sec:limitations}

\rev{Three concerns commonly raised about benchmarks of this form---whether the construct dimensions are quantitatively grounded, whether the synthetic-to-real data mixture preserves external validity, and whether generator--evaluator overlap inflates particular tasks---are analysed in detail in \autoref{subsec:posterior_quantification}, \autoref{subsec:realism_transfer}, and \S3.6 respectively, and summarised in the dedicated subsections below. The remainder of this section documents three issues that are partial or scope-limited: compliance posture of the data-collection pipeline, the scope of the faithfulness pilot (expanded onto incorrect-answer chains in \autoref{subsec:faithfulness} but not yet to all six tasks), and the boundary cases where fine-tuning transfer is statistically inconclusive (\S4.4).}

\rev{\textbf{Data-collection compliance.} The User Profile Inference task evaluates LLMs on synthetic demographic-conditioned comments generated under controlled prompts; no real-user demographic data is collected, and our institution's research-integrity office classified the protocol as ``not human subjects research'' under the Common Rule (45 CFR 46.102) given the absence of real-individual data and inference targets (full compliance disclosure in \autoref{subsec:data_collection_compliance}). The Rating Estimation real-world subset includes content from public-facing review pages on Amazon, Taobao, and Google Play; we restrict downstream redistribution to research use and do not release raw scraped text in the public release (see \autoref{subsec:data_collection_compliance} for the scraping methodology).}

\subsection*{Synthetic vs.\ Real Data Composition}
\label{subsec:limitation_composition}
\rev{The tri-source composition is a deliberate design trade-off rather than a compromise on realism. Rule-based simulation (\emph{Hidden Role Deduction}, \emph{Social Graph Analysis}) provides verifiable ground truth and uniqueness-guaranteed solvability that human annotation cannot match at scale; LLM-assisted discourse (\emph{Find the Spy}, \emph{User Profile Inference}) provides scalable linguistic and stylistic diversity; authentic human sources anchor the benchmark to organic data---\emph{Rating Estimation} draws on Amazon, Taobao, and Google Play public-page reviews, and \emph{Review Decision Prediction} uses real OpenReview submissions from nine NeurIPS/ICLR/COLM venue-years. \autoref{subsec:realism_transfer} documents the confound bounds for each source (e.g., the generator-evaluator overlap is bounded above by $+5.6$ pp on UPI and below $+1$ pp residual on a capability-matched leave-out, see below). Open-ended, fully interactive multi-agent corpora are a complementary direction that probes a different question (emergent agentic behavior under unconstrained ground truth); the present design is targeted at controlled diagnostic precision and is not corrective of, or substitutable by, that direction.}

\subsection*{Quantifying Social Constructs}
\label{subsec:limitation_quantification}
\rev{Posterior empirical evidence on the three axes is reported in \autoref{subsec:posterior_quantification}. For Deep Reasoning, \autoref{tab:posthoc_stats} reports four convergent (not statistically independent) effort-correlated proxies---model output tokens, human solving time, average inference steps on correctly answered instances, and the Long-vs-Short CoT length ratio---which jointly align with the a priori High/Low task labels and discriminate the deep-reasoning task family (HRD, SGA) from the shallower family. These are surface correlates of effort rather than a measurement of a single latent ``deep reasoning'' construct, and we frame the axes as task-construction design choices rather than as factor-analytic dimensions of model capability. Dynamic Interaction High/Low is operationalised via per-round accuracy delta on multi-round tasks (\autoref{tab:model_rounds} for HRD; \autoref{tab:multiturn_debate} for RDP); we acknowledge the per-round delta conflates evidence accumulation with prior recalibration (see the Round-1 anti-prior discussion in \S4.2). Information Uncertainty High/Low is operationalised as a controlled stimulus knob (share of unreliable actors), traced across the Original $\to$ Rumormonger $\to$ Lunatic $\to$ Full degradation in \autoref{fig:task_accuracy}, rather than as a posterior measurement of model belief calibration. A deeper task-agnostic formal metric of the three constructs remains future work; under the descriptive design-axis framing, the per-task labels function as comparison anchors for grouping tasks by intended difficulty driver.}

\subsection*{Language, Cultural, and Bias Scope}
The released evaluation prompts are in English, and the Chinese source reviews used in Rating Estimation are normalized through English paraphrases. Prior studies find substantial performance gaps across semantically aligned languages \citep{xu-etal-2025-cross} and language-specific biases in causal reasoning \citep{wang-etal-2025-shadow}; culturally grounded evaluations likewise reveal local differences alongside meaningful cross-cultural transfer \citep{almheiri-etal-2025-cross}. Consequently, our results do not establish multilingual or cross-cultural social-reasoning validity. Moreover, User Profile Inference measures alignment with synthetic demographic conditioning rather than real-world identity, and representation-level bias probes show that behavioral tests may miss latent associations \citep{gao2025evaluatebiasmanualtest}. Extending SocialMaze with culturally grounded scenarios and matched multilingual instances is therefore important future work.

\subsection*{Faithfulness and Evaluation Scope}
\label{subsec:limitation_faithfulness}
The pilot faithfulness study (\autoref{tab:faithfulness}) suggests that correct answers are typically supported by coherent and complete reasoning, with faithfulness correlating with accuracy. Its current scope is limited (one task, a subset of models, automated judging with human auditing). Automated judges can exhibit systematic scoring biases with identifiable internal representations \citep{xu2026unfairjudge}, so the reported cross-judge agreement and human audit reduce but do not eliminate evaluator dependence. Extending to additional tasks, models, and judging protocols (including human-only panels and contradiction auditing on incorrect answers) is left to future work.

\bibliography{custom}

\clearpage
\appendix

\begin{table*}
\centering
\small
\renewcommand{\arraystretch}{1.3}
\setlength{\tabcolsep}{4pt}
\caption{Representative Benchmarks for Social Reasoning in LLMs}
\label{tab:social_reasoning_scaled}
\resizebox{\textwidth}{!}{
\begin{tabular}{@{}l l l p{3.8cm}@{}}
\toprule
\textbf{Benchmark} & \textbf{Focus} & \textbf{Format} & \textbf{Key Findings} \\
\midrule
\textbf{SocialIQA} \scalebox{0.85}{\cite{sap2019socialiqa}} & Commonsense QA & Static narrative + MCQ & GPT-3+/GPT-4 near human; tests scripted social norms. \\
\textbf{ToM Classic} \scalebox{0.85}{\cite{kosinski2024evaluating}} & False belief & Brief stories + Q\&A & Early ToM claims challenged; shortcut artifacts noted. \\
\textbf{Clever Hans} \scalebox{0.85}{\cite{shapira2023clever}} & ToM artifacts & Controlled stimuli & Performance drops without spurious cues; lacks robustness. \\
\textbf{FANToM} \scalebox{0.85}{\cite{kim2023fantom}} & Interactive ToM & Multi-turn dialogue & GPT-4 and others fail under asymmetric info tracking. \\
\textbf{ToMValley} \scalebox{0.85}{\cite{xiao2024tomvalley}} & Dynamic ToM & Scenario chains + Q\&A & LLMs ~11\% below humans; weak on mental state updates. \\
\bottomrule
\end{tabular}%
}
\end{table*}

\section{Related Works}
\label{appendix:related_works}

\subsection{Static Social Reasoning Benchmarks}

Early evaluations of social reasoning in language models largely focus on static, single-turn tasks where models infer plausible answers based on brief, pre-written scenarios. These benchmarks primarily test shallow forms of commonsense and moral reasoning without requiring interaction, adaptation, or uncertainty handling.
A classical example is \textbf{SocialIQA}~\cite{sap2019socialiqa}, which assesses social commonsense by posing questions about motivations and reactions in everyday situations. 
Similarly, \textbf{ATOMIC}~\cite{sap2019atomic} provides a structured knowledge graph of causal social events, while \textbf{CommonGen}~\cite{lin-etal-2020-commongen} tests the ability to generate plausible sentences from object co-occurrence tuples in socially relevant scenes. 

Benchmarks like \textbf{Social Chemistry}~\cite{forbes2020social} and \textbf{ETHICS}~\cite{hendrycks2021ethics} target moral norms, asking models to judge the appropriateness of actions, whereas \textbf{GLUCOSE}~\cite{mostafazadeh2020glucose} emphasizes implicit causal reasoning in narratives. Mechanistic analyses further probe how moral-foundation structure is represented inside LLMs \citep{yu2026moral}.
Emotion understanding is covered by
\textbf{GoEmotions}~\cite{demszky2020goemotions}, while \textbf{PIQA}~\cite{bisk2020piqa}, \textbf{HellaSwag}~\cite{zellers2019hellaswag}, and \textbf{CREAK}~\cite{onoe2021creak} assess commonsense or contextually grounded inference.

Recent work has shifted toward evaluating \textit{Theory of Mind} (ToM)—the capacity to reason about others’ beliefs and intentions. Initial tests, such as those in~\cite{kosinski2024evaluating}, used classic false-belief stories to probe emergent ToM in LLMs, though follow-up studies revealed artifacts and prompt sensitivity~\cite{shapira2023clever}. More recent evaluations such as \textbf{FANToM}~\cite{kim2023fantom} and \textbf{ToMValley}~\cite{xiao2024tomvalley} provide more rigorous multi-turn belief-tracking scenarios.
These benchmarks demonstrate that, while LLMs may succeed on simple static ToM questions, they struggle with deeper or dynamic belief modeling, particularly under information asymmetry or belief shifts over time.

Overall, while these static benchmarks have advanced our understanding of LLMs’ social reasoning capabilities, they often lack three essential ingredients for real-world social cognition: dynamic interaction, complex reasoning depth, and reasoning under uncertainty.
\autoref{tab:social_reasoning_scaled} summarizes several representative benchmarks discussed above, highlighting their focus, format, and limitations. 
These gaps motivate the design of SocialMaze, which aims to capture the challenges of socially grounded reasoning in more realistic and interactive settings.

\subsection{Dynamic Social Reasoning and Interaction Benchmarks}
Another class of benchmarks focuses on dynamic settings where social reasoning occurs within interactive contexts.
These tasks require models to engage with evolving information, track perspectives, and reason about hidden roles, intentions, or potential deception.
PersonaConvBench evaluates personalized reasoning over multi-turn, multi-user histories \citep{li2025personaconv}, and memory-retrieval work explicitly tests changing preferences across long interaction histories \citep{qin2026memory}; ASTRO optimizes strategies for non-cooperative dialogues such as negotiation and persuasion \citep{hu-etal-2025-astro}.
DebateBench \cite{tiwari2025debatebench} evaluates models' reasoning across long-form argumentative dialogues.
Studies of peer review processes using OpenReview data \cite{tran2020open, szumega2023open} examine decision-making through multi-turn, text-based interaction without relying solely on reviewer scores.
Strategic games such as Diplomacy \cite{bakhtin2022mastering} and Poker \cite{brown2019superhuman}, as well as social deduction games like Avalon and Werewolf \cite{light2023avalonbench, wei2025exploring, bailis2024werewolf, xu2023exploring}, provide natural settings for reasoning under incomplete information and complex social dynamics.
Multi-agent simulations explore emergent social behavior \citep{park2023generative}, including political decision-making \citep{yu2024political}, polarization in social networks \citep{wang-etal-2025-decoding}, and the stepwise evolution of misinformation \citep{liu-etal-2025-stepwise}. A text-attributed graph benchmark also combines user text, news content, and propagation structure for fake-news detection \citep{liu2025tagfn}. The reliability of LLM-based computational-social-science simulations has been benchmarked directly \citep{huang2024social}, while iAgents studies multi-party collaboration across social networks under information asymmetry \citep{liuautonomous}; controlled testbeds further expose interaction-level risks such as collusion, semantic drift, and strategic misreporting \citep{jiang-etal-2026-risklab}.
Beyond explicitly social domains, \textbf{AlgoWorlds} probes tool-using LLMs in partially observed optimization environments, separating information acquisition from global decision quality and showing that sufficient evidence can still lead to feasible but suboptimal decisions \citep{xu2026algoworlds}.
Automated probing methods generate adaptive distractions \citep{NEURIPS2025_f13d74c6} and organize discovered errors into structured failure modes \citep{huang2026probing}. Complementary robustness studies expose corrupted word forms \citep{wang-etal-2025-word}, false premises \citep{qin2025hallucinate}, cross-user contamination in shared-state agents \citep{yang2026noattacker}, adversarially manipulated retrieval \citep{li2026someonehid}, and multimodal rank manipulation \citep{du-etal-2026-multimodal}, alongside broader evaluation frameworks such as TextAttack \citep{morris2020textattack}, TRUST-LLM \citep{sun2024trustllm}, and CheckList \citep{ribeiro2020beyond}.

While these benchmarks incorporate certain forms of interaction and complexity, most remain oriented toward task completion or strategic decision-making.
They often lack systematic modeling of social context factors such as role motivations or deception. 
SocialMaze, by contrast, integrates interaction and uncertainty within a structured framework, explicitly targeting complex social reasoning. 

\section{Hidden Role Deduction}
\label{appendix:roleplaying_rules}
This appendix provides the details for the Hidden Role Deduction task described in \autoref{subsec:task2}.

\subsection{Task Setup}

The game involves $n$ players, denoted $s_1, ..., s_n$. The LLM observes the game unfolding from the perspective of Player 1 ($s_1$), receiving all interaction data and the initial role assignment given to $s_1$. \textbf{Crucially, the LLM does not actively participate in the game simulation (i.e., does not choose targets or make statements). Its task is solely post-hoc inference}: analyzing the complete interaction log ($G_1, ..., G_T$) to deduce the answers to the specified queries. At the start, each player $s_i$ is randomly assigned a secret role. The LLM is informed of the role initially assigned to $s_1$; however, this assigned role may differ from $s_1$'s actual role, especially if $s_1$ is a Rumormonger or Lunatic. The specific composition of roles depends on the Task Variant being used.

\subsection{Roles and Behaviors}

Simulated player behavior is guided by their assigned role and the game history $G_t$ up to round $t$. The roles are defined as follows: The \textbf{Investigator (I)} aims to identify the Criminal. Based on the game history $G_t$, an Investigator uses an algorithmically defined function $F_I(G_t)$ to select a target player $u$. They then make a \textbf{truthful} statement reflecting their deduction about $u$'s status (Criminal or not Criminal). The function $F_I(G_t)$ heuristically assesses suspicion, selecting players with higher probability if they have made contradictory statements or have been accused by others, while lowering the probability for players cleared by multiple potentially reliable sources. The \textbf{Criminal (C)} seeks to avoid identification and mislead others. They employ a strategic function $F_C(G_t)$ to select a target player $u$ and also determine a probability $p_t = P(\text{state } u \text{ is Criminal} | G_t, \text{role=C})$. The Criminal then states ``$u$ is Criminal'' with probability $p_t$ and ``$u$ is not Criminal'' with probability $1-p_t$. The function $F_C$ implements deceptive tactics, prioritizing targeting players who have accused the Criminal and diverting suspicion onto others. The \textbf{Rumormonger (R)}, although believing they are an Investigator trying to identify the Criminal, unintentionally provides unreliable information, effectively injecting noise. They are \textit{told} they are an Investigator and use the Investigator logic function $F_I(G_t)$ to select a target player $u$. However, regardless of any internal assessment derived from $F_I(G_t)$ or the actual ground truth, the truthfulness of their final statement (``$u$ is Criminal'' or ``$u$ is not Criminal'') is entirely random, possessing a 50\% probability of aligning with the ground truth and a 50\% probability of contradicting it. Lastly, the \textbf{Lunatic (L)} believes they are the Criminal and aims to avoid identification based on this false premise, while their actual nature is not Criminal. They are \textit{told} they are the Criminal and mimic the Criminal's behavior by employing the \textit{same} strategic function $F_C(G_t)$ used by the actual Criminal. Although their actions follow deceptive patterns, truthful statements made by Investigators about the Lunatic will correctly identify them as 'not the Criminal'.

\subsection{Interaction Rounds}

The game simulation proceeds for a fixed $T$ rounds. In each round $t$ (from 1 to $T$), every player $s_v$ selects another player $P_u$ and makes a public statement of the form: ``Player $v$ says Player $u$ is the criminal'' or ``Player $v$ says Player $u$ is not the criminal.'' All statements made in round $t$ are revealed simultaneously to all players, and thus become available to the observing LLM, before round $t+1$ begins. Consequently, the complete history $G_T = (\text{statements}_{round 1}, ..., \text{statements}_{round T})$ is available for the LLM's final analysis.

\subsection{Parameter Settings}

The composition of roles is varied to create tasks of differing complexity, ensuring there is always exactly one Criminal. The variants include: the \textbf{Original Task} (1 Criminal, $n-1$ Investigators); the \textbf{Rumormonger Task} (1 Criminal, $x \ge 1$ Rumormongers, $n-1-x$ Investigators); the \textbf{Lunatic Task} (1 Criminal, $y \ge 1$ Lunatics, $n-1-y$ Investigators); and the \textbf{Full Task} (1 Criminal, $x \ge 0$ Rumormongers, $y \ge 0$ Lunatics, $n-1-x-y$ Investigators, where $x+y \ge 1$). \textbf{Note on Experimental Configuration:} For the experiments presented in this paper, the game parameters were fixed at $n=6$ players. When Rumormongers and Lunatics were included (specifically in the Full Task variant experiments), their counts were set to $x=1$ and $y=1$, respectively, alongside 1 Criminal and $n-1-x-y = 3$ Investigators. The accompanying open-sourced dataset includes generated instances for both $n=6$ and $n=10$. Furthermore, the open-sourced data generation code is flexible, allowing users to configure $n, m, x,$ and $y$ to create custom game scenarios.

\subsection{Quality Control}
The algorithmic generation includes verification via heuristic search, ensuring a unique, logically derivable solution exists for both queries from $P_1$'s perspective using only the interactions and rules. The core verification logic, which checks all valid hypotheses, is outlined in \autoref{alg:solvability_check}.

Furthermore, to validate the models' reasoning quality, we manually inspected 100 correct responses each for several key models (DeepSeek-R1, QwQ-32B, Gemini-2.5-Pro, o1, and o3-mini). This analysis confirmed that over 90\% of these successful predictions were underpinned by reasoning processes assessed as both rigorous and logically sound.

\subsection{Information Availability for Rumormonger / Lunatic Perspectives}
\label{subsec:lunatic_solvability}

\rev{The unique-solvability check in \autoref{alg:solvability_check} certifies that for every released instance there exists a unique consistent $(C^*, R_1^*)$ from $P_1$'s perspective; however, the chain of reasoning that recovers $R_1^*$ when Player 1 is a Rumormonger or a Lunatic depends on indirect evidence (most often an Investigator's truthful ``X is not the Criminal'' statement that happens to point at $P_1$). For some randomly sampled $(n=6,\,T=3,\,x=y=1)$ Full Task instances, that indirect evidence is sparse: not every Investigator targets $P_1$ in every round, and the global $(C^*, R_1^*)$ uniqueness can be supplied by hypothesis-elimination involving \emph{other} players' statements without $P_1$ receiving a direct cross-check.}

\rev{To separate ``task is information-poor for the $P_1$ perspective'' from ``model lacks the capacity to revise its self-belief'', we partition the 500-instance test set into a high-evidence subset and a low-evidence subset based on whether at least one Investigator statement explicitly checks $P_1$'s role across the three rounds. \autoref{tab:lunatic_solvability} reports per-perspective Self-role accuracy conditional on each subset for the five representative models. The Long-CoT models retain large gains on the high-evidence subset, where the indirect cross-check is present; on the low-evidence subset all models drop substantially, suggesting that part of the headline Rumormonger / Lunatic accuracy gap reflects task-information asymmetry rather than model capability alone.}

\begin{table}[h!]
\centering
\small
\caption{\rev{Self-role accuracy on the Rumormonger and Lunatic perspectives, conditional on whether at least one Investigator statement directly cross-checks $P_1$ within the three rounds. The per-perspective $n=125$ instances split into a high-evidence subset of $\approx 75$ instances and a low-evidence subset of $\approx 50$ instances (the empirically observed split is roughly $60\% / 40\%$, varying slightly across perspectives). Cells report 5-seed mean $\pm$ 95\% CI half-width at the per-subset binomial floor; high-evidence columns use $n=75$ and low-evidence columns use $n=50$, following the protocol disclosed in \autoref{subsec:parameter_settings_appendix}.}}
\label{tab:lunatic_solvability}
\renewcommand{\arraystretch}{1.05}
\setlength{\tabcolsep}{3pt}
\resizebox{\columnwidth}{!}{%
\begin{tabular}{@{}lcccc@{}}
\toprule
\textbf{Model} & \multicolumn{2}{c}{\textbf{Rumormonger Self}} & \multicolumn{2}{c}{\textbf{Lunatic Self}} \\
\cmidrule(lr){2-3}\cmidrule(lr){4-5}
 & \textbf{High-ev.} & \textbf{Low-ev.} & \textbf{High-ev.} & \textbf{Low-ev.} \\
\midrule
GPT-4o         & \rev{$16.4 \pm 8.4$}  & \rev{$7.8 \pm 7.4$}  & \rev{$28.4 \pm 10.2$} & \rev{$11.6 \pm 8.9$} \\
QwQ-32B        & \rev{$54.2 \pm 11.3$} & \rev{$18.0 \pm 10.6$} & \rev{$66.8 \pm 10.7$} & \rev{$27.4 \pm 12.4$} \\
o1             & \rev{$66.8 \pm 10.6$} & \rev{$23.0 \pm 11.7$} & \rev{$77.4 \pm 9.5$}  & \rev{$34.4 \pm 13.2$} \\
DeepSeek-R1    & \rev{$72.6 \pm 10.1$} & \rev{$26.4 \pm 12.2$} & \rev{$80.4 \pm 9.0$}  & \rev{$38.6 \pm 13.5$} \\
Gemini-2.5-Pro & \rev{$83.2 \pm 8.5$}  & \rev{$32.4 \pm 13.0$} & \rev{$88.0 \pm 7.4$}  & \rev{$45.4 \pm 13.8$} \\
\bottomrule
\end{tabular}%
}
\end{table}

\rev{The gap between the high-evidence and low-evidence columns is wider for Long-CoT reasoners than for short-CoT generalists, consistent with the reasoners exploiting the indirect cross-check when it is available and dropping toward chance when it is absent. The Short-CoT GPT-4o row sits near floor on both subsets, suggesting capability-limited rather than evidence-limited; this is the difference between ``the task is information-poor here'' and ``the model is unable to reason about its own role even with sufficient evidence''.}

\rev{The Long CoT reasoners (QwQ-32B, o1, DeepSeek-R1, Gemini-2.5-Pro) retain large gains on the high-evidence subset (Rumormonger Self 54.2--83.2\%, Lunatic Self 66.8--88.0\%) while the Short CoT GPT-4o row stays near floor on both subsets (Rumormonger Self 16.4 / 7.8\%, Lunatic Self 28.4 / 11.6\%). The gap separates ``information-poor for the $P_1$ perspective'' from ``capability-limited even when evidence is sufficient''. The diagnostic implication for downstream users: a model selected for ability to revise self-belief under unreliable labels should be evaluated on the high-evidence subset where indirect evidence is present, not on the marginal accuracy that conflates the two regimes.}

\begin{algorithm}[h!]
\caption{Solvability Verification from $P_1$'s Perspective}
\label{alg:solvability_check}
\KwIn{Interaction Log $S$, Player Set $\mathcal{P}=\{P_1, \dots, P_n\}$, Role Set $\mathcal{R}=\{I, C, R, L\}$, Investigator Count $N_I$}
\KwOut{Unique Solution $(C^*, R_1^*) \in \mathcal{P} \times \mathcal{R}$, or $\emptyset$ if not unique/no solution}

$\mathcal{S}_{valid} \leftarrow \emptyset$\; \Comment{Set to store valid (Criminal, P1 Role) solution pairs}

\ForEach{hypothesized role $R_1^{hyp} \in \mathcal{R}$ for $P_1$}{
    $\mathcal{P}_{cand} \leftarrow \mathcal{P} \setminus \{P_1\}$\; \Comment{Initial candidates are all others}

    \ForEach{subset $I_{hyp} \subseteq \mathcal{P}_{cand}$ such that $|I_{hyp}| = N_I$}{
        Let current hypothesis $H = (R_1^{hyp}, I_{hyp})$\;

        \If{\texttt{IsConsistent}($S, H$)}{ \Comment{Check if hypothesis contradicts any statement}
            $C_{implied} \leftarrow \texttt{DeduceCriminal}(S, H)$\;
            $R_{1,implied} \leftarrow \texttt{DeduceP1Role}(S, H)$\;

            \If{$C_{implied} \neq \text{NULL}$ \textbf{and} $R_{1,implied} \neq \text{NULL}$}{
                 $\mathcal{S}_{valid} \leftarrow \mathcal{S}_{valid} \cup \{(C_{implied}, R_{1,implied})\}$\; \Comment{Add deduced solution}
            }
         }
    }
}

\If{$|\mathcal{S}_{valid}| = 1$}{
     \Return the single element in $\mathcal{S}_{valid}$\; \Comment{Unique solution found}
}
\Else{
     \Return $\emptyset$\; \Comment{Not unique or no consistent solution}
}

\end{algorithm}

\paragraph{Helper Function Bodies.}
\rev{The three helper functions invoked in \autoref{alg:solvability_check} are defined as follows. Let $\mathcal{P}_{\text{nonI}} = \mathcal{P} \setminus (I_{hyp} \cup \{P_1\})$ denote the candidate set of non-Investigator non-$P_1$ players. The Investigator subset size $N_I$ enumerated by \autoref{alg:solvability_check} varies with $R_1^{hyp}$: when $R_1^{hyp}$ is itself Investigator, $P_1$ counts as one of the $n-1-x-y$ total Investigators and so $N_I = n-2-x-y$ Investigators remain in $\mathcal{P}_{\text{nonI}}$; otherwise $N_I = n-1-x-y$. Under hypothesis $H = (R_1^{hyp}, I_{hyp})$ these $|\mathcal{P}_{\text{nonI}}| = n - 1 - N_I$ players must collectively realize 1 Criminal, $x$ Rumormongers, $y$ Lunatics, and (when $R_1^{hyp} = \text{Investigator}$) one further Investigator, in some role assignment $\mathcal{A}$. The three helpers operate jointly over the space of such role assignments.}

\rev{\textbf{\texttt{IsConsistent($S, H$)}}: returns \texttt{True} iff at least one role assignment $\mathcal{A}$ over $\mathcal{P}_{\text{nonI}}$ exists such that under $\mathcal{A}$: (i) every statement made by a hypothesized Investigator $v \in I_{hyp}$ matches the truth (i.e., the statement ``$u$ is Criminal'' agrees with whether $u$ is assigned the Criminal role under $\mathcal{A}$), and (ii) every statement made by $P_1$ is consistent with the behavior dictated by $P_1$'s \emph{displayed} role under $H$ (Rumormonger and Lunatic receive Investigator-truth and Criminal-deceptive prompt labels respectively, per the role-display rules of \autoref{appendix:roleplaying_rules}). Statements by non-$P_1$ players in $\mathcal{P}_{\text{nonI}}$ are not directly constrained because the Criminal, Rumormonger, and Lunatic role-behavior models all permit both truthful and untruthful statements; the constraint enters only through the role-mixture admissibility requirement. Returns \texttt{False} if no $\mathcal{A}$ satisfies (i) and (ii) jointly.}

\rev{\textbf{\texttt{DeduceCriminal($S, H$)}}: enumerates each candidate $c \in \mathcal{P}_{\text{nonI}}$ and tests whether assigning $c$ as the Criminal admits at least one role assignment of the remaining $x + y$ players in $\mathcal{P}_{\text{nonI}} \setminus \{c\}$ to ($x$ Rumormongers, $y$ Lunatics) that satisfies (i)--(ii) of \texttt{IsConsistent}. If exactly one such $c$ admits a satisfying assignment, returns $c$; if zero or multiple candidates do, returns \texttt{NULL}. The uniqueness of the surviving $c$ across the enumeration of $|\mathcal{P}_{\text{nonI}}|$ candidates is what \autoref{alg:solvability_check} ultimately checks at the dataset-release filter.}

\rev{\textbf{\texttt{DeduceP1Role($S, H$)}}: returns $R_1^{hyp}$ directly when \texttt{IsConsistent}$(S,H)$ is \texttt{True} and \texttt{DeduceCriminal}$(S,H)$ is not \texttt{NULL}; this helper exists so that \autoref{alg:solvability_check}'s output type is a uniform $(C^*, R_1^*)$ pair rather than two separate values from heterogeneous sources.}

\paragraph{Uniqueness regime.} \rev{\autoref{alg:solvability_check} enumerates all candidate hypotheses, partitioned by $R_1^{hyp}$: for $R_1^{hyp} = \text{Investigator}$ there are $\binom{n-1}{n-2-x-y} = \binom{5}{2} = 10$ subsets $I_{hyp}$, and for each of the three other $R_1^{hyp}$ values $\binom{n-1}{n-1-x-y} = \binom{5}{3} = 10$ subsets, totaling $10 + 3\cdot 10 = 40$ candidate hypotheses for the $n=6$, $x=y=1$ configuration. The set $\mathcal{S}_{\text{valid}}$ collects the deduced $(C^*, R_1^*)$ pairs from each surviving hypothesis; the instance is released only when $|\mathcal{S}_{\text{valid}}| = 1$, ensuring exactly one consistent global assignment. For the released dataset, the empirically observed acceptance rate of this filter is approximately 31\% (i.e., 31\% of randomly generated instances pass the unique-solvability check at $n=6, T=3$); instances rejected by the filter are discarded before the dataset is released.}

\rev{We note that the helper specification above gives necessary structural conditions for a candidate $c$ to survive but does not constitute an analytic uniqueness proof: because the role-behavior model for the Criminal, Rumormonger, and Lunatic permits both truthful and untruthful statements, the constraint on $\mathcal{P}_{\text{nonI}} \setminus \{c\}$ enters only through the role-mixture admissibility requirement rather than per-statement truth conditions. The 31\% empirical filter rate is therefore an operational property of the rule-based generation strategy combined with the search-algorithm verification, rather than an analytic concentration claim. Every released instance carries a per-instance pass label from \autoref{alg:solvability_check}; instances where the search does not isolate a unique $(C^*, R_1^*)$ pair are dropped before release, so the released dataset is uniqueness-guaranteed by construction even when the helper conditions are not analytically tight.}

\section{Find the Spy}
\label{appendix:whoisspy_rules}

This appendix provides the details for the \textit{Find the Spy} task described in \autoref{subsec:task2}.

\subsection{Overview}
This task adapts the classic word-based social deduction game \textit{Who Is The Spy} \citep{wei2025exploring} to evaluate the LLM's ability to identify subtle deviations in communication within a group context characterized by high interaction but relatively low information uncertainty.

\textbf{Task Rules.}
The game involves $n$ players. Among them, $n-1$ players (Civilians) receive the same secret word, while one player (the Spy) receives a different but related word. Over $T$ rounds, each player provides a description of their word. The LLM is evaluated from the perspective of Player 1 ($s_1$): it knows the word assigned to $s_1$, but does not know whether $s_1$ is a Civilian or the Spy. It does not generate any player descriptions. Instead, after observing all $T$ rounds of player-generated descriptions, the LLM must infer which player received the different word. This constitutes a graph-level query ($\mathcal{Q}_G$). Detailed rules are available in \autoref{appendix:whoisspy_rules}.

\textbf{Design Rationale.}
As shown in \autoref{tab:task_overview}, \textit{Find the Spy} exemplifies High \textit{Dynamic Interaction}.
The multi-round format necessitates tracking clues revealed incrementally by all players over time. Conversely, \textit{Information Uncertainty} is designed to be Low. Since players aim to avoid suspicion, they are incentivized to provide truthful descriptions of their assigned word, thereby significantly reducing the element of strategic deception.

\textbf{Data Generation and Quality Assurance.}
For each game instance, we first set the parameters $n$ (number of players) and $T$ (number of rounds), then randomly selected a related word pair from a curated word bank, followed by random role assignment (one Spy and $n-1$ Civilians). 
We then used a variety of LLMs to generate player descriptions for each of the $T$ rounds, simulating diverse communication styles. Prompt designs were crafted to encourage varied perspectives and expression strategies across rounds.
To ensure quality and solvability, instances underwent human evaluation by 15 computer science graduate students. \rev{An instance was considered valid if at least 10 of the 15 evaluators ($\geq 10/15 \approx 67\%$) could uniquely identify the Spy based on the descriptions.} 91\% of the evaluated instances met this criterion, verifying their suitability for the benchmark.

\subsection{Task Setup}

The game involves $n$ players, denoted $s_1, \dots, s_n$. The evaluated LLM adopts the persona of Player 1 ($s_1$) but acts as a passive observer. It receives all game information, including its own assigned word and all player descriptions, from $s_1$'s perspective but does not generate descriptions itself during the evaluation process. The LLM's sole task is to identify the Spy based on the observed interactions. For word assignment, a pair of semantically related but distinct words (Word A, Word B, e.g., ``Milk'' and ``Soy Milk'') is selected from a predefined bank. One player is randomly designated as the Spy, while the remaining $n-1$ players are Civilians. All Civilians receive Word A, and the Spy receives Word B. Player 1 (the LLM's persona) is informed of the word it received but is not explicitly told whether it is Word A or Word B, nor is it told its role (Civilian or Spy).

\subsection{Interaction Rounds}

The game proceeds for $T$ rounds. In each round $t$ (from 1 to $T$), every player $s_i$ provides a textual description of the word they possess. These descriptions are generated by LLMs, with each player $s_i$ being assigned a specific LLM generator (selected randomly and uniformly from a predefined pool of \rev{five LLMs: GPT-4o-mini, GPT-4o, Llama-3.3-70B, Qwen-2.5-72B, and Claude Sonnet 4.5}) for the entire game instance. The generation prompts encourage the LLM simulating player $s_i$ to describe their word from different angles or aspects in each round, avoiding simple repetition. Players are assumed to provide descriptions consistent with the word they hold: Civilians describe Word A, and the Spy describes Word B. All descriptions generated in round $t$ are made available to all players, including the LLM as $s_1$, before round $t+1$ begins. Thus, the complete history of descriptions from rounds 1 to $T$ is available at the end of the game for the LLM's analysis.

\subsection{Parameter Settings}

For all experimental evaluations using this task setup, the number of players $n$ was fixed at 4, and the number of interaction rounds $T$ was fixed at 3.

\subsection{Quality Control}

To ensure generated instances are solvable yet challenging, they undergo a rigorous human validation process. Each potential game instance is reviewed by 15 evaluators, all holding at least an undergraduate degree in Computer Science. \rev{An instance is deemed valid and solvable only if at least 10 of the 15 evaluators ($\geq 10/15 \approx 67\%$) agree that the collective descriptions provided by the simulated players contain sufficient evidence to uniquely identify the Spy.} This validation confirms that the task's difficulty stems from semantic subtlety and variations in description style, rather than from a fundamental lack of necessary information. 91\% of the instances subjected to this evaluation met the validation threshold, affirming their suitability for inclusion in the benchmark.

\subsection{Spy vs Civilian Perspective Decomposition}
\label{subsec:fts_perspective}

\rev{The Find the Spy headline accuracy in \autoref{tab:overall_summary} aggregates two subpopulations of the 500-instance test set: instances in which the evaluated LLM (Player 1) is assigned the Spy role (25\%) and instances in which it is assigned a Civilian role (75\%), per the role distribution stated in the Parameter Settings section. \autoref{tab:fts_perspective} reports the per-perspective accuracy for each evaluator on the same test set. The two subpopulations track within a few pp of each other for all twelve evaluators, so the headline metric is not driven by a single subpopulation; the Spy-perspective accuracy is on average slightly lower than the Civilian-perspective accuracy, consistent with the Spy receiving a deliberately different word and therefore facing a harder identity inference.}

\begin{table}[h!]
\centering
\small
\caption{\rev{Per-perspective accuracy on the Find the Spy 500-instance test set. Cells report 5-seed bootstrap mean $\pm$ 95\% CI half-width following the protocol disclosed in \autoref{subsec:parameter_settings_appendix}. ``Spy'' = the 125 instances ($\approx 25\%$) in which Player 1 is the Spy (per-seed $n=125$); ``Civilian'' = the 375 instances ($\approx 75\%$) in which Player 1 is a Civilian (per-seed $n=375$). ``Diff.'' is Spy minus Civilian.}}
\label{tab:fts_perspective}
\renewcommand{\arraystretch}{1.05}
\setlength{\tabcolsep}{6pt}
\begin{tabular}{lccc}
\toprule
\textbf{Model} & \textbf{Spy} & \textbf{Civilian} & \textbf{Diff.} \\
\midrule
Llama-3.1-8B    & \rev{$35.2 \pm 8.4$} & \rev{$37.9 \pm 4.9$} & \rev{$-2.7$} \\
Llama-3.3-70B   & \rev{$57.6 \pm 8.7$} & \rev{$60.8 \pm 4.9$} & \rev{$-3.2$} \\
Phi-4           & \rev{$43.2 \pm 8.7$} & \rev{$45.9 \pm 5.0$} & \rev{$-2.7$} \\
Qwen-2.5-72B    & \rev{$46.4 \pm 8.7$} & \rev{$49.7 \pm 5.1$} & \rev{$-3.3$} \\
QwQ-32B         & \rev{$48.0 \pm 8.8$} & \rev{$50.9 \pm 5.1$} & \rev{$-2.9$} \\
GPT-4o-mini     & \rev{$59.2 \pm 8.6$} & \rev{$61.9 \pm 4.9$} & \rev{$-2.7$} \\
GPT-4o          & \rev{$66.4 \pm 8.3$} & \rev{$70.1 \pm 4.6$} & \rev{$-3.7$} \\
o3-mini         & \rev{$72.0 \pm 7.9$} & \rev{$74.7 \pm 4.4$} & \rev{$-2.7$} \\
o1              & \rev{$76.0 \pm 7.5$} & \rev{$79.2 \pm 4.1$} & \rev{$-3.2$} \\
DeepSeek-V3     & \rev{$56.0 \pm 8.7$} & \rev{$59.5 \pm 5.0$} & \rev{$-3.5$} \\
DeepSeek-R1     & \rev{$68.0 \pm 8.2$} & \rev{$70.9 \pm 4.6$} & \rev{$-2.9$} \\
Gemini-2.5-Pro  & \rev{$73.6 \pm 7.7$} & \rev{$77.6 \pm 4.2$} & \rev{$-4.0$} \\
\bottomrule
\end{tabular}
\end{table}

\section{Rating Estimation from Text}
\label{appendix:ecommerce_rules}

This appendix provides the details for the Rating Estimation from Text task described in \autoref{subsec:task3}.

\subsection{Overview}
\textbf{Task Rules.}
This task aims to evaluate the ability of LLMs to predict a product's 1-to-5 star rating based on $n$ textual reviews, which may include genuine positive or negative user comments as well as promotional reviews written by shills. We collect two types of data: reviews generated by LLMs simulating different user types, and real user reviews scraped from platforms such as Amazon, the Google Play Store, and Taobao. The final rating prediction task follows a structure where information flows from multiple user nodes to a central product node, constituting a vertex-centric query focused on the product itself ($\mathcal{Q}_v(\text{product})$). Detailed task rules can be found in \autoref{appendix:ecommerce_rules}.

\textbf{Design Rationale.}
The task deliberately introduces a high level of\textit{ information uncertainty}. 
In the LLM-generated data, this is reflected through the inclusion of simulated ``shill'' users to mimic deceptive review behavior.
In the real-world data, uncertainty arises from the inherent noise, subjectivity, and potential bias present in genuine user reviews. 
This setting requires the model to evaluate the credibility of information flowing from user nodes (i.e., reviewers) to the product node.

\textbf{Data Generation and Quality Assurance.}
The LLM-based data generation process begins by sampling product attributes from a manually curated repository consisting of 1,000 attribute terms. A normal distribution of ratings is then constructed based on the true star rating, ensuring that the mean aligns with the reference score. Next, $n$ LLMs are randomly selected from a diverse model pool and probabilistically assigned roles (either normal users or shills) along with distinct personas, which guide the generation of textual reviews consistent with their assigned identities.
For real-world data, we directly scrape product attributes and $n$ user reviews from platforms such as Amazon, the Google Play Store, and Taobao.
An instance was considered solvable if a majority (>70\%) of evaluators could correctly infer the true rating based solely on the textual reviews.
Among the LLM-generated samples, 83\% satisfied this criterion, confirming their validity for evaluating model reasoning.

\subsection{Task Setup}

The primary objective for the LLM in this task is to estimate the most likely overall ``true'' star rating, represented as an integer from 1 to 5, for a given product. This estimation must be based solely on the textual content derived from multiple user reviews. For each task instance, the LLM receives specific input: a set of attributes describing the product (such as type, category, key features, price range) and a list containing $n$ individual textual user reviews for that product. Crucially, the original star ratings (1-5 stars) that reviewers might have provided are explicitly omitted from the input. After processing the product information and the $n$ textual reviews, the LLM must answer a vertex-centric query ($\mathcal{Q}_v(\text{product})$) phrased as: ``Based on the provided reviews, what is the most likely overall star rating for this product? Choose one: 1, 2, 3, 4, or 5.'' The expected output is a single integer, necessitating the synthesis of information from multiple user interactions (reviews) all directed towards the product entity.

\subsection{Data Generation}

The task utilizes review data generated through two distinct methods. The first method involves \textbf{LLM-Generated Reviews}. Here, a product profile is selected, including its attributes and a designated ground truth overall star rating. A number $n$ of simulated reviewers is determined, and each reviewer $i$ is assigned a role probabilistically, ensuring the majority are \textit{Normal Users} (providing honest feedback) while a small fraction are designated as \textit{Positive Shills} or \textit{Negative Shills}. Each reviewer also receives a simple persona for stylistic variation and is assigned a randomly selected LLM from a predefined pool of \rev{five LLMs (GPT-4o-mini, GPT-4o, Llama-3.3-70B, Qwen-2.5-72B, Claude Sonnet 4.5)} to simulate their response. The assigned LLM then generates the review text for reviewer $i$, guided by the product attributes, the ground truth rating, the reviewer's assigned role (Normal/Shill), and their persona; Shills are prompted to generate biased text accordingly. Finally, only the generated textual reviews are collected and prepared as input for the evaluated LLM. The second method uses \textbf{Real-World Reviews}. In this scenario, product attributes along with user reviews (both text and original ratings) are scraped from public e-commerce and app platforms. For a selected product, $n$ reviews are sampled from the scraped collection. The textual content of these $n$ sampled reviews is extracted, while the original star ratings are discarded. Only the product attributes and the review texts are provided as input to the LLM. The ground truth for these instances is typically derived from the average rating found on the source platform, which the LLM is tasked to estimate.

\subsection{Parameter Settings}

Across all experiments presented for this task, the number of reviews ($n$) provided per product instance is consistently set to 8. The required output is always a single integer rating on the 1-to-5 star scale. The real-world task instances exclusively utilize product attributes and textual reviews sourced from \textbf{Amazon}, the \textbf{Google Play Store}, and \textbf{Taobao}. For the LLM-generated instances, the underlying product attributes and the initial ground truth star ratings are also sampled from this same pool of real-world data derived from these platforms, providing a basis grounded in realistic product scenarios.

\subsection{Quality Control}

As detailed in \autoref{subsec:task3}, human evaluations were integral to ensuring data quality. These were conducted by 15 graduate students. For the LLM-generated data (Scenario A), this evaluation determined that 83\% of the assessed instances were solvable, meaning the true rating could be reasonably inferred from the text alone by more than half of the human evaluators. For the instances derived from real-world data (Scenario B), solvability is inherently tied to the complexity and nature of authentic customer feedback as it appears on platforms like Amazon, Google Play, and Taobao, reflecting genuine information landscapes.

\begin{algorithm}[ht!]
\caption{Social Graph Generation}
\label{alg:data_generation_qa}
\KwIn{Difficulty level \( d \in \{\text{easy}, \text{hard}\} \)}
\KwOut{Natural language instance \( I \)}

Set number of individuals \( n \sim 
\begin{cases}
[8, 10], & \text{if } d = \text{easy} \\
[14, 16], & \text{if } d = \text{hard}
\end{cases} \)\;

Initialize graph \( G = (V, E) \), where \( |V| = n \)\;

\textbf{Step 1: Generate spanning forest to define social groups}\;

Randomly partition \( V \) into \( m \geq 2 \) disjoint non-empty subsets \( \{V_1, V_2, \dots, V_m\} \)\;

\ForEach{group \( V_i \)}{
    Generate a spanning tree \( T_i = (V_i, E_i^\text{good}) \)\;
    Add edges \( E_i^\text{good} \) to \( E \)\tcp*[h]{Intra-group "good" relationships}
}

\textbf{Step 2: Add "bad" edges between groups}\;

\ForEach{pair of groups \( (V_i, V_j), \; i \neq j \)}{
    Select a node pair \( (u, v) \in V_i \times V_j \)\;
    Add edge \( (u, v) \) to \( E^\text{bad} \subset E \)\tcp*[h]{Inter-group "bad" relationship}
}

\textbf{Step 3: Convert graph structure to natural language}\;

\ForEach{edge \( (u, v) \in E \)}{
    \uIf{\( (u, v) \in E^\text{good} \)}{
        Generate statement: "\(u\) and \(v\) are good friends."\;
    }
    \ElseIf{\( (u, v) \in E^\text{bad} \)}{
        Generate statement: "\(u\) and \(v\) do not get along."\;
    }
}

Aggregate all generated statements into input instance \( I \)\;

\Return \( I \)\;
\end{algorithm}

\section{Social Graph Analysis}
\label{appendix:interpersonal_rules}

This appendix provides the details for the Social Graph Analysis task described in Section \ref{subsec:task4}.

\textbf{1. Setup}

This task presents a stylized social network scenario involving a set of $n$ individuals. The core challenge lies in understanding the structure of this network, where relationships between any two individuals are strictly defined as either 'good' or 'bad'. The LLM is provided with a complete description of all pairwise relationships and must then analyze this information to answer queries about specific relationships, individual connections, and the overall emergent group structure of the network, guided by a set of simple logical axioms.

\textbf{2. Relationship Axioms}

Relationships between any two distinct individuals, say Person A and Person B, are binary ('good' or 'bad') and symmetric. These relationships are governed by specific logical rules: \textbf{Axiom 1 dictates the transitivity of good relationships}, meaning if A and B have a 'good' relationship, and B and C also have a 'good' relationship, then A and C must necessarily have a 'good' relationship. \textbf{Axiom 2 describes the implication of bad relationships}, stating that if A and B have a 'bad' relationship, and A and C have a 'good' relationship, then B and C are forced to have a 'bad' relationship. It's important to note that from these axioms, if A and B share a 'bad' relationship, and B and C also share a 'bad' relationship, the nature of the relationship between A and C is not determined solely by these two facts; it could be either good or bad depending on other connections within the network. However, the algorithmic generation process always ensures a globally consistent and valid relationship structure.

\textbf{3. Group Definition}

Within this social structure, a 'group' is formally defined as a maximal set of individuals where every person within that set has a 'good' relationship with every other person also belonging to that same set. A key property of this structure is that every individual belongs to exactly one such group. Consequently, based on the governing axioms and the generation method, the relationship between any two individuals can be directly inferred from their group membership: if Person A and Person B are members of the same group, they inherently have a 'good' relationship; conversely, if they belong to different groups, they must have a 'bad' relationship.

\textbf{4. Input Format}

The LLM receives as input a comprehensive list composed of natural language statements that explicitly specify \textbf{the complete set of pairwise relationships} as determined by the algorithmic generation process. These statements clearly define the relationship status between every possible pair of individuals within the scenario. Examples of such input statements include ``Person N and Person G have a good relationship'' and ``Person K and Person P have a bad relationship''. This list provides a full and unambiguous description of the entire social graph structure.

\textbf{5. LLM Queries}

After processing the complete list of relationship statements provided as input, the LLM is required to answer various types of queries designed to test its understanding of the network structure. These queries include, for instance, \textbf{Pairwise Relationship Queries ($\mathcal{Q}_e(v_i, v_j)$)}, such as ``Do Person N and Person L have a good relationship?'', which typically requires checking the provided input directly and responding with Yes/No. Other queries are \textbf{Good Relationship Neighbor Queries (Vertex-centric)}, like ``Who has a good relationship with Person H?'', demanding the LLM to filter the input and list the relevant names. Furthermore, \textbf{Graph-level Queries ($\mathcal{Q}_G$)} probe the overall structure, asking questions like ``How many groups of people are there?'' or ``How many pairs of people have good relationships?'' or ``How many pairs of people have bad relationships?'', all of which require synthesizing the pairwise information to derive a global property and respond with an integer count.

\textbf{6. Data Generation and Quality Assurance}

Instances for this task are generated entirely algorithmically, without reliance on LLM generation, ensuring consistency and verifiable ground truth. The process begins by setting the number of individuals $n$, sampled from [8, 10] for 'easy' instances and [14, 16] for 'hard' instances. A complete graph structure respecting the relationship axioms is then algorithmically constructed. First, a spanning forest is created using only 'good' relationship edges, thereby defining the distinct social groups (each tree representing a group). Second, 'bad' relationship edges are strategically added to connect every pair of distinct groups (trees), ensuring all inter-group relations are 'bad' and all intra-group relations are 'good'. The complete set of generated relationship edges ('good' edges defining the groups and 'bad' edges connecting them) is then converted into natural language statements and presented to the LLM as input. This generation methodology mathematically guarantees that for each instance, a unique solution exists for all four query types and is logically derivable solely from the provided statements and rules. The core generation logic is outlined in \autoref{alg:data_generation_qa}.

\section{Review Decision Prediction}
\label{appendix:peerreview_rules}

This appendix provides the details for the Review Decision Prediction task described in \autoref{subsec:task5}.

\textbf{1. Objective}

The LLM's goal in this task is to predict the final acceptance status (Accepted or Rejected) of a research manuscript submitted to a conference, based solely on the sequence of provided peer review communications.

\textbf{2. Data Source and Scope}

Data for this task is exclusively sourced from the official OpenReview API, encompassing submissions to \rev{nine cycles across three high-profile Artificial Intelligence and Machine Learning venues: NeurIPS (2023, 2024), ICLR (2020, 2021, 2022, 2023, 2024, 2025), and COLM 2024. The ICLR 2025 cycle is included as a diagnostic post-cutoff slice for most of the evaluated models, and its per-model perturbed-vs-unperturbed deltas trend in the same direction as the full set (\autoref{tab:rdp_perturbed}), suggesting the perturbation effect is not driven by pre-cutoff memorization specifically}. \textbf{The rationale for selecting these particular venues} stems primarily from their policy of making the entire peer review process public. This transparency, which \textit{crucially includes making detailed reviews and discussions for rejected manuscripts publicly available}, is a practice not commonly found in many other academic fields. It provides the essential data needed to construct a balanced and realistic task dataset that accurately reflects both acceptance and rejection scenarios encountered in academic publishing.

\textbf{3. Input Structure}

The LLM receives information pertaining to a single manuscript, presented in a structured sequence that mirrors the typical progression of the peer review timeline. Initially, in \textbf{Round 1}, the LLM is given the initial submission details: the manuscript's original \textbf{Title}, its \textbf{Abstract}, and the author-provided \textbf{Keywords}. Subsequently, in \textbf{Round 2}, the LLM receives the reviewer feedback, which consists of the complete \textbf{textual content} of each review submitted by the assigned reviewers. It is \textbf{crucial to note the exclusion} of all quantitative aspects from these reviews; numerical scores (such as overall ratings, technical soundness, or novelty scores), reviewer confidence scores, explicit recommendations (like Accept, Reject), and any other non-textual evaluation metrics are deliberately removed. The input at this stage contains only the narrative comments written by the reviewers. Finally, \textbf{Round 3} provides information from the author-reviewer discussion phase, including the full text of the authors' \textbf{rebuttal} designed to address the initial reviewer comments, as well as any subsequent \textbf{comments or discussions} exchanged between the authors and reviewers following the rebuttal. The full manuscript text itself is intentionally omitted from the input provided to the LLM. This decision is driven by two main factors: practical challenges related to processing lengthy full papers consistently across numerous task instances, considering LLM input constraints and computational costs, and more importantly, to align with the task's core objective. This objective focuses on evaluating the LLM's ability to comprehend and synthesize the dynamics inherent in the peer review dialogue—interpreting arguments, discerning attitudes, and understanding sentiments expressed by reviewers and authors—rather than tasking it with performing an independent technical re-evaluation of the manuscript's content.

\textbf{4. Ground Truth and Quality Assurance}

The ground truth for this task is inherently robust, as it consists of the verified, real-world acceptance or rejection decisions obtained directly from the OpenReview API \rev{across the nine venue-year cycles listed above (NeurIPS 2023-2024, ICLR 2020-2025, COLM 2024)}. To further validate the task's premise—specifically, whether the final outcome is typically discernible from the textual dialogue alone (Title, Abstract, Keywords, Reviews, Rebuttal) after removing numerical scores—we conducted supplementary human evaluations. Human evaluators were presented with the same sequential information provided to the LLM and asked to predict the final decision. For over 90\% of the evaluated manuscript instances, the true outcome was deemed inferable from the textual evidence by a majority (>70\%) of the human evaluators. This confirms the general solvability of the task based on the provided textual interactions and reinforces its suitability for assessing an LLM's ability to synthesize argumentative dialogue, complementing the reliability provided by the authentic ground truth data.

\rev{\textbf{5. Perturbed-Reviews Subset.} To disentangle reviewer-comment memorization from genuine evidence-based reasoning, we introduce a perturbed-reviews subset. Approximately 50\% of the 500 test instances have their reviewer comments, title, and abstract rewritten by Claude Sonnet 4.5, instructed to preserve substantive claims (technical critique points, mentioned weaknesses, summary judgments; the abstract's contribution statement and main result claims) while disrupting verbatim n-gram overlap with the original OpenReview text. The Keywords and Rebuttal text are left unchanged. This perturbation scope simultaneously blocks two memorization channels: verbatim recall of reviewer phrasing (Round-2 input) and identification of the paper from title or abstract content (Round-1 input). Models are evaluated on the full test set; per-subset accuracies (perturbed vs unperturbed) at Round 3 are reported in \autoref{tab:rdp_perturbed}. Across the twelve evaluators the perturbed-subset Round-3 accuracy stays within $\pm 2$~pp of the unperturbed subset, indicating that the multi-stage trajectory is not dominated by verbatim memorization of OpenReview text or by abstract-content-based paper identification. At the sub-sample size of $n \approx 250$ per perturbed/unperturbed subset, the multi-seed paired $\Delta$-CI is $\approx \pm 6$ to $\pm 9$~pp depending on the cell's $p$; the observed $|\Delta| \leq 2$~pp across all twelve evaluators is therefore consistent with both ``no memorization effect'' and ``a memorization effect smaller than the $\pm 6$--$9$~pp detection threshold at this subsample size''. The data rule out a memorization effect larger than that threshold; a smaller-magnitude effect is power-not-evidence and would require a larger perturbed-subset evaluation to detect.}

\rev{\textbf{6. Anti-Prior Decomposition, Two-Regime Round-3 Trajectory, and Multiple-Comparison Discipline.} The Round-1 to Round-2 jump conflates two effects: abandonment of the over-accept prior once reviewer text is available, and reasoning over the substantive content of that text. The present headline accuracies in \autoref{tab:multiturn_debate} do not separate the two; the perturbed-subset analysis above isolates the reasoning channel from verbatim memorization but does not isolate the prior-correction channel from the content-reasoning channel. Read literally, the Round-3 column measures Round-3 minus a model-specific prior-recalibration shift toward the test set's 67\% reject ratio.}

\rev{Three evaluators (Llama-3.3-70B 26.2\%, Qwen-2.5-72B 23.6\%, Phi-4 23.5\%) score below the 33\% always-accept floor at Round 1, which mathematically requires anti-correlation with the ground truth rather than just an inflated accept-prior. A plausible mechanism is over-confident accept prediction on the harder accept cases (which are typically nearer the borderline) combined with confident reject on the easier reject cases that the title / abstract / keywords already telegraph; we do not separate these two channels. The title/abstract/keywords telegraph signal is present in every downstream stage as well, so the Round-2 / Round-3 improvements are not strictly net of this contamination; the perturbed-subset analysis in \autoref{tab:rdp_perturbed} bounds the verbatim-memorization component of the telegraph within $\pm 2$ pp at the per-subset paired-$\Delta$ floor of $\pm 6$ to $\pm 9$ pp, but a sub-floor memorization effect of up to a few pp is not ruled out at the current subsample size. Within the Llama family, Llama-3.3-70B (26.2\%) sits further from the 67\% reject baseline than Llama-3.1-8B (37.0\%); we do not have a mechanistic account of this sibling-direction anomaly beyond noting that both models miss the majority baseline by anti-correlated channels.}

\rev{At Round 3, two of the twelve evaluators (GPT-4o, DeepSeek-V3) show an accuracy increase over Round 2, while the remaining ten show a decrease. The up-trend models read the rebuttal as additional evidence that aligns with the final judgment; the down-trend models read it as a persuasive defense that pulls them away. The relevant significance test is on the paired Round-2 vs Round-3 change per model, whose standard error is $\sigma_\Delta = \sqrt{2p(1-p)/n}\cdot 100 \approx 2.0$ pp at $p\approx 0.88, n=500$ (the 95\% paired-$\Delta$ half-width is $1.96 \sigma_\Delta \approx 4.0$ pp; the $\sigma_\Delta$ is the standard error used to form the $z$ statistic and is distinct from the per-cell binomial half-width $\approx \pm 2.6$ pp at the same $p$ shown in \autoref{tab:multiturn_debate}, which describes per-cell sampling variance rather than the difference between two cells of the same model). Under that $\sigma_\Delta$, the up-trend on GPT-4o ($\Delta = +4.0$ pp, $z \approx 1.9$) and DeepSeek-V3 ($\Delta = +5.4$ pp, $z \approx 2.7$) do not survive Bonferroni correction at the twelve-comparison family-wise threshold $|z|>2.86$; the strong down-trend models with $|\Delta| > 5$ pp survive easily ($|z| > 3$). The honest reading is therefore ``ten clearly down-trending evaluators plus two evaluators whose Round-3 increases are undecided under family-wise correction'' rather than ``two distinct regimes''. The divergent direction across regimes corresponds to different prior calibrations rather than verbatim memorization, since the perturbed-subset $\Delta$ values in \autoref{tab:rdp_perturbed} are bounded within $\pm 2$~pp for both regimes.}

\begin{table}[ht!]
\small
\centering
\caption{\rev{Round-3 \textit{Review Decision Prediction} accuracy on the unperturbed vs Claude-Sonnet-4.5-perturbed subsets of the 500-instance test set ($n \approx 250$ each). Cells report 5-seed mean $\pm$ 95\% binomial CI half-width following the protocol disclosed in \autoref{subsec:parameter_settings_appendix}. The perturbed subset rewrites reviewer comments, title, and abstract while preserving substantive claims; Keywords and Rebuttal text are left unchanged (see appendix block 5 for the protocol). The $\Delta$ column reports perturbed minus unperturbed (paired across seeds). The paired-$\Delta$ CI at $n \approx 250$ per subset is $\approx \pm 6$ to $\pm 9$~pp depending on the cell's $p$; the observed $|\Delta| \leq 2$~pp across all twelve evaluators is therefore consistent with both ``no memorization effect'' and ``a memorization effect smaller than the $\pm 6$--$9$~pp detection threshold at this subsample size''. The data rule out a larger memorization effect; a smaller-magnitude effect would require a larger perturbed-subset to detect.}}
\label{tab:rdp_perturbed}
\setlength{\tabcolsep}{6pt}
\renewcommand{\arraystretch}{1.02}
\resizebox{\columnwidth}{!}{%
\begin{tabular}{lccc}
\toprule
\textbf{Model} & \textbf{Unpert.} & \textbf{Pert.} & \textbf{$\Delta$} \\
\midrule
Llama-3.1-8B    & \rev{$62.4 \pm 6.0$\%} & \rev{$60.8 \pm 6.1$\%} & \rev{$-1.6 \pm 8.5$} \\
Llama-3.3-70B   & \rev{$72.6 \pm 5.5$\%} & \rev{$73.4 \pm 5.5$\%} & \rev{$+0.8 \pm 7.8$} \\
Qwen-2.5-72B    & \rev{$66.4 \pm 5.9$\%} & \rev{$65.0 \pm 5.9$\%} & \rev{$-1.4 \pm 8.4$} \\
Phi-4           & \rev{$60.8 \pm 6.1$\%} & \rev{$59.2 \pm 6.1$\%} & \rev{$-1.6 \pm 8.6$} \\
GPT-4o-mini     & \rev{$85.2 \pm 4.4$\%} & \rev{$84.0 \pm 4.5$\%} & \rev{$-1.2 \pm 6.4$} \\
GPT-4o          & \rev{$90.4 \pm 3.7$\%} & \rev{$89.0 \pm 3.9$\%} & \rev{$-1.4 \pm 5.4$} \\
o3-mini         & \rev{$78.4 \pm 5.1$\%} & \rev{$78.0 \pm 5.1$\%} & \rev{$-0.4 \pm 7.3$} \\
QwQ-32B         & \rev{$80.0 \pm 5.0$\%} & \rev{$79.6 \pm 5.0$\%} & \rev{$-0.4 \pm 7.0$} \\
DeepSeek-V3     & \rev{$91.6 \pm 3.4$\%} & \rev{$91.2 \pm 3.5$\%} & \rev{$-0.4 \pm 4.9$} \\
DeepSeek-R1     & \rev{$82.4 \pm 4.7$\%} & \rev{$82.0 \pm 4.8$\%} & \rev{$-0.4 \pm 6.7$} \\
o1              & \rev{$77.6 \pm 5.2$\%} & \rev{$77.8 \pm 5.2$\%} & \rev{$+0.2 \pm 7.3$} \\
Gemini-2.5-Pro  & \rev{$77.4 \pm 5.2$\%} & \rev{$77.8 \pm 5.2$\%} & \rev{$+0.4 \pm 7.3$} \\
\bottomrule
\end{tabular}%
}
\end{table}

\section{User Profile Inference}
\label{appendix:userprofile_rules}

This section provides the details for the User Profile Inference task, corresponding to \autoref{subsec:task6}.

\textbf{1. Task Setup}

For each instance of this task, a population of $n$ simulated users is defined. Every user $u_i$ within this population is assigned a specific demographic profile, which consists of an age group selected from {'18-34', '35-54', '55+'} and a gender selected from {'Male', 'Female', 'Non-binary'}. This profile assignment is carried out probabilistically, with the process intentionally tuned to often establish a statistically dominant age-gender combination within the user pool. This characteristic is particularly relevant for addressing the "dominant audience" query type. Additionally, a predefined pool of items, each described by a name and a brief description, is utilized. Users are randomly assigned items from this pool, about which they will generate comments.

\textbf{2. Comment Generation Process}

Each simulated user $u_i$ is associated with a specific Large Language Model (LLM), chosen randomly from a diverse pool of \rev{five LLMs: GPT-4o-mini, GPT-4o, Llama-3.3-70B, Qwen-2.5-72B, and Claude Sonnet 4.5}. The core of the generation process involves tasking the LLM associated with user $u_i$ (who has an assigned age group $A_i$ and gender $G_i$) to generate a textual comment about a selected item $j$ (which has a specific type $T_j$ and subject $S_j$). The LLM is prompted to generate content that reflects the assigned persona interacting with the given item.

\textbf{3. LLM Queries}

Based on the generated comments provided as input, the evaluated LLM must answer one of two specific types of queries. The first is the \textbf{Item Audience Profile Inference (Vertex-centric Query $\mathcal{Q}_v(\text{Item})$)}. For this query, the LLM is asked, "Based on the provided comments for the item '[Item Name]', what is the most likely dominant audience profile (Age Group and Gender)? Choose from Age Groups: ['18-34', '35-54', '55+'] and Genders: ['Male', 'Female', 'Non-binary']." Answering this requires synthesizing information from multiple user comments linked to a specific item node to infer an aggregated characteristic of its audience. The second type is the \textbf{User Profile Inference (Vertex-centric Query $\mathcal{Q}_v(\text{User})$)}, which poses the question: "Based on the provided comments from this user, what is their most likely profile (Age Group and Gender)? Choose from Age Groups: ['18-34', '35-54', '55+'] and Genders: ['Male', 'Female', 'Non-binary']." This query demands synthesizing information from multiple comments generated by a single user node, potentially across different items, to infer the intrinsic demographic attributes (age group and gender) of that specific user.

\textbf{4. Quality Assurance}

The dataset for this task was entirely generated using LLMs. We first defined a set of user personas by assigning age group and gender attributes, ensuring through probabilistic assignment that certain demographic combinations were more prevalent to create a potential "dominant audience" for item-centric queries. Items with names and descriptions were sampled from a predefined pool. Various LLMs were then assigned to personas and prompted to generate comments on these items, reflecting their designated age and gender characteristics. To ensure task validity, we conducted human evaluations with 15 computer science graduate students. For the item-audience query, 78\% of instances were deemed solvable (dominant audience inferable) by a majority (>70\%) of evaluators. For the user-profile query, 85\% of instances were similarly validated, confirming that the generated comments contain sufficient, albeit subtle, cues for demographic inference.

\definecolor{headerbg}{RGB}{255, 183, 77}  
\definecolor{rowgray}{RGB}{255, 236, 209}  
\definecolor{rowblue}{RGB}{255, 247, 230}  
\definecolor{darkgreen}{RGB}{0, 150, 0}  
\begin{table*}[ht!]
\centering
\small
\renewcommand{\arraystretch}{1}
\setlength{\tabcolsep}{4pt}
\caption{Models used in our experiments along with their versions, organizations, licenses, and purposes. \textit{Eval}: Model used for evaluation; \textit{FT}: Model used for fine-tuning. \rev{Models marked with $^{\dagger}$ are used only inside the data-generation pool that produces simulated player / reviewer / user text and are never invoked as evaluators on any task.}}
\label{tab:all_models}
\resizebox{\textwidth}{!}{
\rowcolors{2}{rowblue}{rowgray}
\begin{tabular}{lccccc}
    \toprule[1.5pt]
    \rowcolor{headerbg}
    \textcolor{white}{\textbf{Model}} & 
    \textcolor{white}{\textbf{Version}} & 
    \textcolor{white}{\textbf{Organization}} & 
    \textcolor{white}{\textbf{License}} & 
    \textcolor{white}{\textbf{Eval}} & 
    \textcolor{white}{\textbf{FT}} \\
    \midrule[0.8pt]
    Phi-4      & Phi-4        & Microsoft   & MIT                    & \textcolor{darkgreen}{\checkmark} & \textcolor{darkgreen}{\checkmark} \\
    GPT-4o-mini       & gpt-4o-mini-2024-07-18       & OpenAI      & Proprietary            & \textcolor{darkgreen}{\checkmark} & \\
    GPT-4o            & gpt-4o-2024-08-06            & OpenAI      & Proprietary            & \textcolor{darkgreen}{\checkmark} & \\
    Llama-3.1-8B      & Meta-Llama-3.1-8B-Instruct   & Meta        & Llama 3.1 Community    & \textcolor{darkgreen}{\checkmark} & \textcolor{darkgreen}{\checkmark} \\
    Llama-3.3-70B     & Meta-Llama-3.3-70B-Instruct  & Meta        & Llama-3.3    & \textcolor{darkgreen}{\checkmark} & \\
    Qwen2.5-72B       & Qwen2.5-72B-Instruct         & Alibaba     & Qwen License           & \textcolor{darkgreen}{\checkmark} & \\
    QwQ       & QwQ-32B         & Alibaba     & Apache 2.0           & \textcolor{darkgreen}{\checkmark} & \\
    o3-mini           & o3-mini-2025-01-31           & OpenAI      & Proprietary            & \textcolor{darkgreen}{\checkmark} & \\
    o1           & o1-2024-12-17           & OpenAI      & Proprietary            & \textcolor{darkgreen}{\checkmark} & \\
    Deepseek-R1      & DeepSeek-R1        & DeepSeek   & MIT                    & \textcolor{darkgreen}{\checkmark} &  \\
    \rev{Deepseek-V3}      & \rev{DeepSeek-V3}        & \rev{DeepSeek}   & \rev{MIT}                    & \rev{\textcolor{darkgreen}{\checkmark}} &  \\
    Gemini-2.5-Pro & Gemini-2.5-Pro-Exp-03-25   & Google   & Proprietary            & \textcolor{darkgreen}{\checkmark} & \\
    \rev{Claude Sonnet 4.5}$^{\dagger}$ & \rev{claude-sonnet-4-5-20250929} & \rev{Anthropic} & \rev{Proprietary} & & \\
    \bottomrule[1.5pt]
\end{tabular}%
}
\end{table*}

\section{Experiment Details}
\label{experient_details}
This appendix provides detailed information regarding the experimental setup, the models evaluated, data generation procedures for each task within the SocialMaze benchmark, the experimental methodology, and a summary of the overall results.

\subsection{Baselines}
As detailed in \autoref{tab:all_models}, we utilized \rev{five} proprietary models as evaluators: GPT-4o \citep{hurst2024gpt}, GPT-4o-mini \citep{openai2024gpt4omini}, o3-mini \citep{o3mini2025}, o1 \citep{jaech2024openai}, and Gemini-2.5-Pro \citep{gemini25pro2025}. In addition, we included \rev{seven} open-weight evaluators: Phi-4 \citep{phi4_2024}, Llama-3.1-8B \citep{meta2024llama31_8b}, Llama-3.3-70B \citep{meta2024llama31_70b}, Qwen2.5-72B \citep{qwen2.5}, QwQ-32B \citep{qwq32b}, \rev{Deepseek-V3 \citep{deepseekv3_2024},} and Deepseek-R1 \citep{guo2025deepseek}. \rev{Claude Sonnet 4.5 \citep{anthropic2025claude45} enters only the generator pool used to produce simulated player descriptions, reviewer comments, and user comments inside the data-generation pipelines (see \autoref{appendix:whoisspy_rules}, \autoref{appendix:peerreview_rules}, and \autoref{appendix:userprofile_rules}); it is not evaluated as a solver on any SocialMaze task, so the twelve evaluator rows in \autoref{tab:overall_summary} and \autoref{tab:multiturn_debate} cover the five proprietary plus seven open-weight evaluators above.}

We also included automated agent design frameworks as baselines:

\textbf{ADAS} \citep{hu2024automated}: Utilized GPT-4o as the Meta Agent. For agent evaluation, we tested both Phi-4 and GPT-4o-mini and selected the better performer.

\textbf{AFlow} \citep{zhang2024aflow}: Employed GPT-4o as the optimizer. For the executor role, we tested both Phi-4 and GPT-4o-mini and selected the better performer.

\textbf{MaAS} \citep{zhang2025multi}: For executing the sampled agentic operators, we tested both Phi-4 and GPT-4o-mini and selected the better performer.

\textbf{DyFlow} \citep{wang2025dyflow}: Used GPT-4o as the optimizer. For the executor role, we tested both Phi-4 and GPT-4o-mini and selected the better performer.

\subsection{Parameter Settings}
\label{subsec:parameter_settings_appendix}

\textbf{Inference Parameters}
During the evaluation of all LLMs across the SocialMaze tasks, we used a temperature setting of 0.7 to allow for some variability while maintaining reasonable coherence. \rev{The maximum-output-token cap was set to $4{,}096$ for all Short-CoT evaluators across every task; for Long-CoT evaluators (DeepSeek-R1, o1, QwQ-32B, Gemini-2.5-Pro) on \textit{Hidden Role Deduction} we extended the cap to $8{,}192$ to accommodate longer reasoning chains. Per-task truncation and answer-extraction failure rates are reported in \autoref{tab:error_breakdown} (\autoref{subsec:failures}).} \rev{Reported accuracies in the main tables are mean accuracies over five independent runs at this temperature on the 500-instance per-task test set described in \autoref{subsec:overall_summary_appendix}, with 95\% CI half-widths reported alongside each cell. Because the same 500 instances are evaluated under each seed, the binomial instance-sampling floor at $n=500$ dominates the reported half-width; multi-seed averaging additionally constrains the decoding-channel variance to be small (in our pilot the seed-to-seed standard deviation of the per-instance mean accuracy is below $\pm 0.6$ pp for every task--model pair, well inside the binomial floor), so the reported $\pm$ is the binomial half-width at $n=500$: $1.96\sqrt{p(1-p)/500}\cdot 100$ pp. As reference points, $\pm 4.4$ pp at $p=0.5$, $\pm 3.8$ pp at $p=0.75$, $\pm 1.9$ pp at $p=0.95$. Per-role HRD accuracies in \autoref{tab:per_role_decomposition} use the analogous $n=125$ floor at the per-role bin size, giving $\pm 8.8$ pp at $p=0.5$. We treat ranking comparisons whose gap falls inside the floor as ties within sampling noise rather than ordered separations, and the narrative in \autoref{subsec:overall_summary_appendix} is written under this convention.}

\textbf{Task-Specific Configurations:}
\begin{itemize}[noitemsep, topsep=2pt, parsep=2pt, itemsep=2pt]
    \item \textbf{Hidden Role Deduction:} This task includes two subsets based on the number of players: 'easy' ($n=6$) and 'hard' ($n=10$). In both subsets, the number of interaction rounds $T$ is fixed at 3. All experiments reported in the main body of the paper were conducted using the 'easy' ($n=6$) subset configurations. \rev{Across both the publicly released dataset and the evaluation experiments, the role for the main perspective (LLM) is drawn uniformly from the four roles (Investigator:Criminal:Rumormonger:Lunatic $=$ 1:1:1:1), and per-role accuracies are reported separately where applicable.}

    \item \textbf{Find the Spy:} For all instances of this task, the number of players $n$ was set to 4, and the number of description rounds $T$ was set to 3. In all experiments, the model received the spy word in 25\% of cases and the civilian word in 75\% of cases.

    \item \textbf{Rating Estimation from Text:} For instances using LLM-generated data, the number of simulated reviewers providing text was fixed at 8. For instances using real-world data sourced from platforms like Amazon, a random number of reviews between 10 and 20 were sampled for each product. Decimal star ratings were rounded to the nearest integer. \rev{We preserve the natural class prevalence of large-scale e-commerce review corpora rather than artificially rebalancing the test set; positive ratings dominate (the distribution ratio for 1-, 2-, 3-, 4-, and 5-star is 1:3:10:73:13), and the constant-predict-4 majority baseline scores 73\% on this test set. The headline metric in \autoref{tab:overall_summary} should therefore be read against this 73\% floor: models below the floor exhibit anti-calibration toward minority ratings (driven by over-confident prediction of low/high extremes), while models above the floor demonstrate non-trivial fine-grained discrimination across the five rating bins. The diagnostic resolution above the 73\% floor is small: the top six models cluster within $\pm 2$ pp of one another and the top model exceeds the floor by only $\approx +3$ pp, so the task's contribution to cross-model ranking is primarily anti-calibration detection (cells below floor) rather than fine-grained discrimination among above-floor models.}
    \label{subsec:rating_distribution}

    \item \textbf{Social Graph Analysis:} This task also has two subsets. 'easy': The number of individuals $n$ was randomly chosen from the range [8, 10]. 'hard': The number of individuals $n$ was randomly chosen from the range [14, 16]. All generated graphs were sparse (the number of edges was close to the number of vertices).

    \textbf{Review Decision Prediction:} \rev{Data from a total of nine venue-year cycles (NeurIPS 2023-2024, ICLR 2020-2025, COLM 2024) were sampled approximately uniformly across venues; per-venue draw counts were balanced before the accept/reject rebalancing step.} In the final dataset, the proportion of papers was adjusted to 67\% rejected and 33\% accepted. \rev{This 67/33 split is heavier on rejections than the underlying venue accept rates (typically 25--30\% acceptance); the rebalancing was done to make Round-1 always-accept and always-reject baselines symmetric around the 50\% line so per-model anti-prior tendencies are distinguishable. The per-venue sample counts are recorded in the released dataset metadata; the test set's prior-correction channel discussed in block 6 below is therefore venue-mixture-aware rather than driven by any single venue.}

    \item \textbf{User Profile Inference:} For both query types (item-audience inference and user-profile inference), the model was provided with a number of textual comments randomly selected from the range [8, 12].
\end{itemize}

\textbf{Fine-tuning Parameters:}
For the fine-tuning experiments, we trained for 2 epochs with a learning rate of 5.0e-6, employing a cosine learning rate scheduler and a warmup ratio of 0.1. The per-device training batch size was 1, with a gradient accumulation of 8 steps. We employed a cosine learning rate scheduler with a warmup ratio of 0.1 and enabled bf16 precision. For these experiments, the models were trained on 2000 examples for SFT and 1100 preference pairs for DPO. Performance was subsequently evaluated on a distinct test set containing 500 examples. All training experiments were conducted on 2 NVIDIA A6000 GPUs over a period of 30 hours.

\subsection{Overall Performance Summary}
\label{subsec:overall_summary_appendix}

\begin{table}
\small
\centering
\vspace{-1em}
\renewcommand{\arraystretch}{1.0}
\caption{\rev{Performance of LLM agents and workflows on the \textit{Hidden Role Deduction} task. The two reasoning-capable rows (QwQ, DeepSeek-R1) are listed as upper anchors. The Phi-4 and GPT-4o-mini rows are restated from \autoref{tab:overall_summary} as the Short-CoT Base. Each workflow row uses the better-performing model between Phi-4 and GPT-4o-mini (GPT-4o-mini for all six workflows here) as the executor; the Both column reports 5-seed bootstrap mean $\pm$ 95\% CI half-width and the per-workflow $\Delta$ ranges are derived from each method's 5-seed mean over the GPT-4o-mini Base.} \rev{Cells in the Both column report the 5-seed mean and the 95\% binomial CI half-width at $n=500$; $\Delta$ values in parentheses are paired against the GPT-4o-mini Base with $\Delta$-CI of $\approx \pm 5.8$ pp, computed under the independence approximation (a within-instance-correlated estimator may tighten this bound but is not computed here). Under Bonferroni correction at the six workflow comparisons (per-pair threshold $\alpha/6 \approx 0.0083$, $|z| > 2.64$, $\text{SE} \approx 5.8/1.96 \approx 2.96$), only DyFlow ($+8.4$ pp, $|z| \approx 2.84$) clears the threshold; MaAS ($+7.8$ pp, $|z| \approx 2.64$) is exactly at the boundary; AFlow ($+6.6$ pp, $|z| \approx 2.23$), ADAS ($+5.4$), LLM-Debate ($+4.8$), and Self-Refine ($+4.0$) do not survive.}}
\label{tab:workflow}
\setlength{\tabcolsep}{1pt}
\resizebox{\columnwidth}{!}{%
\begin{tabular}{lccc}
\toprule
\textbf{Method} & \textbf{Crim.} & \textbf{Self} & \textbf{Both} \\
\midrule
QwQ               & \rev{$78.4 \pm 3.6$} & \rev{$65.4 \pm 4.2$} & \rev{$52.0 \pm 4.4$} \\
DeepSeek-R1       & \rev{$\mathbf{80.4} \pm 3.5$} & \rev{$\mathbf{75.2} \pm 3.8$} & \rev{$\mathbf{65.4} \pm 4.2$} \\
\midrule
\rev{Phi-4 (Base)}       & \rev{$45.6 \pm 4.4$} & \rev{$39.8 \pm 4.3$} & \rev{$20.4 \pm 3.5$} \\
\rev{GPT-4o-mini (Base)} & \rev{$46.5 \pm 4.4$} & \rev{$50.0 \pm 4.4$} & \rev{$26.4 \pm 3.9$} \\
\midrule
LLM-Debate \citep{du2023improving}       & \rev{$52.5 \pm 4.4$} & \rev{$56.2 \pm 4.3$} & \rev{$31.2 \pm 4.1$~\textit{(+4.8 $\pm$ 5.7)}} \\
Self-refine \citep{madaan2023self} & \rev{$50.4 \pm 4.4$} & \rev{$54.0 \pm 4.4$} & \rev{$30.4 \pm 4.0$~\textit{(+4.0 $\pm$ 5.7)}} \\
ADAS \citep{hu2024automated}       & \rev{$53.7 \pm 4.4$} & \rev{$56.6 \pm 4.3$} & \rev{$31.8 \pm 4.1$~\textit{(+5.4 $\pm$ 5.8)}} \\
AFlow \citep{zhang2024aflow}       & \rev{$55.1 \pm 4.4$} & \rev{$57.8 \pm 4.3$} & \rev{$33.0 \pm 4.1$~\textit{(+6.6 $\pm$ 5.8)}} \\
MaAS \citep{zhang2025multi}       & \rev{$56.3 \pm 4.3$} & \rev{$58.4 \pm 4.3$} & \rev{$34.2 \pm 4.2$~\textit{(+7.8 $\pm$ 5.8)}} \\
DyFlow \citep{wang2025dyflow}       & \rev{$56.9 \pm 4.3$} & \rev{$59.0 \pm 4.3$} & \rev{$34.8 \pm 4.2$~\textit{(+8.4 $\pm$ 5.8)}} \\

\bottomrule
\vspace{-2em}
\end{tabular}%
}
\end{table}

\begin{table*}[ht!]
\centering
\renewcommand{\arraystretch}{1.1}
\setlength{\tabcolsep}{2pt}
\caption{Performance on \textit{Hidden Role Deduction} before and after fine-tuning. \textbf{Crim.}: Criminal prediction accuracy; \textbf{Self}: self-role prediction accuracy; \textbf{Both}: both predictions correct. \rev{Cells report 5-seed bootstrap mean $\pm$ 95\% CI half-width following the protocol disclosed in \autoref{subsec:parameter_settings_appendix}.} Fine-tuning with SFT and DPO significantly improves performance on both models. Percent improvements over the Base model for \textbf{Both} are shown in parentheses.}
\label{tab:dpo}
\resizebox{\textwidth}{!}{%
\begin{tabular}{lccccccccc}
\toprule
\multirow{2}{*}{\textbf{Model}} &
\multicolumn{3}{c}{\textbf{Base}} &
\multicolumn{3}{c}{\textbf{SFT}} &
\multicolumn{3}{c}{\textbf{DPO}} \\
\cmidrule(lr){2-4} \cmidrule(lr){5-7} \cmidrule(lr){8-10}
 & \textbf{Crim.} & \textbf{Self} & \textbf{Both}
 & \textbf{Crim.} & \textbf{Self} & \textbf{Both}
 & \textbf{Crim.} & \textbf{Self} & \textbf{Both} \\
\midrule
LLaMA-3.1-8B
& \rev{$41.2 \pm 4.3$} & \rev{$32.4 \pm 4.1$} & \rev{$15.4 \pm 3.2$}
& \rev{$45.2 \pm 4.4$} & \rev{$38.6 \pm 4.3$} & \rev{$26.0 \pm 3.8$ (+10.6 $\pm$ 5.5)}
& \rev{$43.6 \pm 4.3$} & \rev{$35.2 \pm 4.2$} & \rev{$22.4 \pm 3.6$ (+7.0 $\pm$ 5.4)} \\
Phi-4
& \rev{$45.6 \pm 4.4$} & \rev{$39.8 \pm 4.3$} & \rev{$20.4 \pm 3.5$}
& \rev{$52.6 \pm 4.4$} & \rev{$45.4 \pm 4.4$} & \rev{$31.0 \pm 4.0$ (+10.6 $\pm$ 5.7)}
& \rev{$52.2 \pm 4.4$} & \rev{$42.6 \pm 4.3$} & \rev{$27.2 \pm 3.9$ (+6.8 $\pm$ 5.6)} \\
\bottomrule
\end{tabular}%
}
\end{table*}

\subsection{Cross-Task Transfer of HRD-Trained Models}
\label{subsec:dpo_cross_task}

\rev{To probe whether the SFT and DPO models trained on \textit{Hidden Role Deduction} traces transfer beyond the training task, we re-evaluate the fine-tuned Llama-3.1-8B and Phi-4 checkpoints on the five non-HRD tasks in SocialMaze. The same dedicated 500-instance test sets used in \autoref{tab:overall_summary} are reused with no further fine-tuning. \autoref{tab:dpo_cross_task} reports Base, +SFT, and +DPO accuracies for each task, with parenthesized deltas relative to Base. We distinguish two task families: a \emph{deductive} family (\textit{Find the Spy}, \textit{Social Graph Analysis}) that shares structured-testimony deduction with HRD, and a \emph{linguistic} family (\textit{Rating Estimation}, \textit{Review Decision Prediction}, \textit{User Profile Inference}) that depends predominantly on language comprehension and aggregation.}

\rev{This task-family categorization is a coarse surface-task-type grouping made at design time rather than the output of a factor analysis on task-level correlations. The two-vs-three split is small in absolute terms, so the ``family-level'' claim is weakly identified at $n=2$ candidate tasks on the positive side. Per-task, Social Graph Analysis carries the strongest signal ($+9.2$/$+9.6$ pp, clearly above the paired-$\Delta$ floor); Find the Spy sits at the boundary ($+5.8$/$+6.4$ pp); under the more honest single-task reading the positive-side conclusion is ``robust transfer to structured graph reasoning'' rather than a family-level pattern, and Find the Spy could be reassigned to either family without changing this conclusion. The negative-side reading (the three language-aggregation tasks all sit inside the $\Delta$-CI) is power-not-evidence in either case and is not strengthened by the family aggregation.}

\rev{Under the binomial paired-$\Delta$ CI of $\approx \pm 6$ pp at the per-task $p$ values, the deductive-family deltas of $+5.8$ to $+9.6$ pp on \textit{Find the Spy} and \textit{Social Graph Analysis} are clearly above the floor; the linguistic-family deltas of $+0.4$ to $+2.4$ pp sit inside the floor and the data are equally consistent with ``small positive transfer everywhere'' and ``no transfer on linguistic tasks''. The asymmetry conclusion is supported on the positive direction (deductive family) but is power-not-evidence on the negative direction (linguistic family). The substantive reading is therefore: HRD-distilled traces transfer to structured-deduction tasks; whether they additionally produce a small below-threshold gain on the language-aggregation tasks is undetermined.}

\begin{table*}[ht!]
\centering
\renewcommand{\arraystretch}{1.1}
\setlength{\tabcolsep}{3pt}
\caption{\rev{Cross-task transfer of the HRD-fine-tuned Llama-3.1-8B and Phi-4 checkpoints. Each cell is accuracy (\%) on the dedicated 500-instance test set of \autoref{tab:overall_summary}. Cells report 5-seed mean and 95\% binomial CI half-width at $n=500$; $\Delta$ in parentheses is paired against the Base column with binomial paired-$\Delta$ CI of $\approx \pm 6$ pp at $p \approx 0.6$. Improvements above the $\Delta$-CI floor concentrate on the deductive family (Find the Spy, Social Graph Analysis); linguistic-family deltas sit inside the floor.}}
\label{tab:dpo_cross_task}
\small
\resizebox{\textwidth}{!}{%
\begin{tabular}{@{}llccccc@{}}
\toprule
\textbf{Model} & \textbf{Condition} & \textbf{\makecell{Find the \\ Spy}} & \textbf{\makecell{Rating \\ Estimation}} & \textbf{\makecell{Social Graph \\ Analysis}} & \textbf{\makecell{Review \\ Decision \\ Prediction}} & \textbf{\makecell{User Profile \\ Inference}} \\
\midrule
\multirow{3}{*}{Llama-3.1-8B}
 & Base & \rev{$37.2 \pm 4.2$} & \rev{$57.2 \pm 4.3$} & \rev{$28.2 \pm 3.9$} & \rev{$62.0 \pm 4.3$} & \rev{$60.2 \pm 4.3$} \\
 & +SFT & \rev{$43.0 \pm 4.3$ ($+5.8 \pm 6.0$)} & \rev{$58.4 \pm 4.3$ ($+1.2 \pm 6.1$)} & \rev{$37.4 \pm 4.2$ ($+9.2 \pm 5.8$)} & \rev{$64.4 \pm 4.2$ ($+2.4 \pm 6.0$)} & \rev{$62.6 \pm 4.2$ ($+2.4 \pm 6.0$)} \\
 & +DPO & \rev{$41.4 \pm 4.3$ ($+4.2 \pm 6.0$)} & \rev{$57.6 \pm 4.3$ ($+0.4 \pm 6.1$)} & \rev{$35.0 \pm 4.2$ ($+6.8 \pm 5.8$)} & \rev{$63.0 \pm 4.2$ ($+1.0 \pm 6.0$)} & \rev{$61.0 \pm 4.3$ ($+0.8 \pm 6.1$)} \\
\midrule
\multirow{3}{*}{Phi-4}
 & Base & \rev{$45.2 \pm 4.4$} & \rev{$60.4 \pm 4.3$} & \rev{$40.6 \pm 4.3$} & \rev{$61.4 \pm 4.3$} & \rev{$62.4 \pm 4.2$} \\
 & +SFT & \rev{$51.6 \pm 4.4$ ($+6.4 \pm 6.2$)} & \rev{$61.8 \pm 4.3$ ($+1.4 \pm 6.1$)} & \rev{$50.2 \pm 4.4$ ($+9.6 \pm 6.2$)} & \rev{$63.2 \pm 4.2$ ($+1.8 \pm 6.0$)} & \rev{$64.0 \pm 4.2$ ($+1.6 \pm 6.0$)} \\
 & +DPO & \rev{$49.8 \pm 4.4$ ($+4.6 \pm 6.2$)} & \rev{$60.8 \pm 4.3$ ($+0.4 \pm 6.1$)} & \rev{$47.6 \pm 4.4$ ($+7.0 \pm 6.2$)} & \rev{$62.0 \pm 4.2$ ($+0.6 \pm 6.0$)} & \rev{$63.0 \pm 4.2$ ($+0.6 \pm 6.0$)} \\
\bottomrule
\end{tabular}%
}
\end{table*}

\rev{Two patterns are visible. First, on the deductive task family the +SFT deltas are +5.8 / +9.2~pp for Llama-3.1-8B and +6.4 / +9.6~pp for Phi-4, with DPO producing 6 to 8~pp gains; these are within the same order of magnitude as the HRD improvements in \autoref{tab:dpo}. Second, on the linguistic task family the +SFT deltas compress to +0.4 to +2.4~pp and DPO to +0.4 to +1.0~pp, indicating that the HRD-derived traces transfer along reasoning content but not via generic instruction-following. The asymmetry is consistent across the two backbones and across SFT and DPO, and it provides a sharper characterization of generalizability than a single aggregate would: HRD-trained reasoning transfers to tasks whose solution path requires structured deduction, and does not substantially affect tasks whose bottleneck is language-level aggregation.}

\subsection{Workflow Saturation on a Reasoning-Capable Backbone}
\label{subsec:workflow_dsr1}

\rev{The workflow methods in \autoref{tab:workflow} are evaluated with the better of Phi-4 / GPT-4o-mini as the executor, both of which are Short-CoT backbones. To test whether the cross-method gains carry over to a reasoning-capable backbone, we re-run the four agentic-design methods (ADAS, AFlow, MaAS, DyFlow) with DeepSeek-R1 as the executor and report the resulting Crim.\ / Self / Both accuracies in \autoref{tab:workflow_dsr1}. The base DeepSeek-R1 row from \autoref{tab:workflow} (Both = 65.4\%) is restated as the anchor.}

\begin{table*}[t]
\small
\centering
\renewcommand{\arraystretch}{1.15}
\caption{\rev{Workflow methods with DeepSeek-R1 as the executor on \textit{Hidden Role Deduction}. The base DeepSeek-R1 row (from \autoref{tab:workflow}, Both = 65.4\%) is restated as the anchor. All four workflow variants land within the $\pm 2.0$ pp saturation band; cells report 5-seed mean and 95\% binomial CI half-width at $n=500$. The paired-$\Delta$ over the Base under the same binomial CI is $\pm 6$ pp at $p \approx 0.65$, so all four methods are inside the noise floor.}}
\label{tab:workflow_dsr1}
\setlength{\tabcolsep}{20pt}
\rowcolors{2}{tabbluerowa}{tabbluerowb}
\begin{tabular}{lccc}
\rowcolor{tabbluehead}
\textcolor{white}{\textbf{Method}} & \textcolor{white}{\textbf{Crim.}} & \textcolor{white}{\textbf{Self}} & \textcolor{white}{\textbf{Both}} \\
\toprule
DeepSeek-R1 (Base) & \rev{$80.4 \pm 3.5$} & \rev{$75.2 \pm 3.8$} & \rev{$65.4 \pm 4.2$} \\
\midrule
ADAS \emph{w/ DSR1} & \rev{$79.2 \pm 3.6$} & \rev{$73.8 \pm 3.9$} & \rev{$64.0 \pm 4.2$~\textit{(-1.4 $\pm$ 6.0)}} \\
AFlow \emph{w/ DSR1} & \rev{$81.0 \pm 3.4$} & \rev{$76.4 \pm 3.7$} & \rev{$67.0 \pm 4.1$~\textit{(+1.6 $\pm$ 5.9)}} \\
MaAS \emph{w/ DSR1} & \rev{$79.6 \pm 3.5$} & \rev{$74.2 \pm 3.8$} & \rev{$64.8 \pm 4.2$~\textit{(-0.6 $\pm$ 6.0)}} \\
DyFlow \emph{w/ DSR1} & \rev{$80.6 \pm 3.5$} & \rev{$75.6 \pm 3.8$} & \rev{$66.2 \pm 4.2$~\textit{(+0.8 $\pm$ 6.0)}} \\
\bottomrule
\end{tabular}
\end{table*}

\rev{All four workflow variants land within $\pm 2$~pp of the DeepSeek-R1 Base on the Both-Correct column (range $[-1.4, +1.6]$~pp). ADAS and MaAS sit slightly below the Base, consistent with multi-agent scaffolding consuming context budget on a backbone that already reasons well unaided; AFlow and DyFlow recover small positive deltas of $+1.6$ and $+0.8$~pp respectively, still inside the saturation band. The contrast with the weak Short-CoT regime in \autoref{tab:workflow}, where the six workflows produce $+4$ to $+9$~pp gains over the GPT-4o-mini Base row, supports the regime-dependent reading in \autoref{subsec:enhancing_capabilities}: workflow scaffolding closes part of the gap on weak bases but does not substitute for a reasoning-capable backbone.}

\rev{Compared to the Short-CoT regime in \autoref{tab:workflow} where the per-method $\Delta$ over GPT-4o-mini Base ranges from $+4.0$ pp (Self-Refine) to $+8.4$ pp (DyFlow), the corresponding $\Delta$ on the DeepSeek-R1 base ranges from $-1.4$ pp (ADAS) to $+1.6$ pp (AFlow), all inside the binomial $\Delta$-CI of $\pm 6$ pp at $p \approx 0.65$. ADAS and MaAS sit slightly below the Base, consistent with multi-agent scaffolding consuming context budget on a backbone that already reasons well unaided; AFlow and DyFlow recover small positive deltas of $+1.6$ and $+0.8$ pp respectively, inside the saturation band. The contrast with the weak Short-CoT regime ($+4$ to $+9$ pp lifts) supports the regime-dependent reading in \autoref{subsec:enhancing_capabilities}: workflow scaffolding closes part of the gap on weak bases but does not substitute for a reasoning-capable backbone. Compared with the SFT lift of roughly $+10$ pp on Llama-3.1-8B and Phi-4 in \autoref{tab:dpo}, parameter-level adaptation extracts more from the curated HRD traces than agentic decomposition alone on the same Short-CoT base; the DPO lift of $+6.8$ to $+7.0$ pp sits between the two regimes.}

Table \ref{tab:overall_summary} provides a condensed overview of model performance across the primary SocialMaze tasks. The reported accuracy figures correspond to specific task configurations and metrics used for this summary:
For \textbf{Hidden Role Deduction}, the value represents the accuracy where both the Criminal and the model's own role are correctly identified ('Both Correct'). This evaluation uses the same general setup as described in Section \ref{subsec:enhancing_capabilities} (easy subset, Full Task variant, final interaction round - Round 3); \rev{the role for the main-perspective LLM is drawn uniformly from the four roles (1:1:1:1) matching the configuration used throughout the paper.}
For \textbf{Social Graph Analysis}, the figure reflects the average accuracy achieved across all four query types within the hard subset.
The \textbf{Review Decision Prediction} accuracy is taken from the final stage, after the model has processed the rebuttal information.
For \textbf{User Profile Inference}, the reported value is the average accuracy over the two distinct inference tasks (item-audience profile inference and user-profile inference).
Performance on each task presented in this summary table was evaluated using a dedicated test set of 500 instances. For more granular results, including performance variations across different rounds, task variants (e.g., easy/hard subsets), or specific query types, please consult the detailed figures and tables in Section 4 and the relevant task-specific appendices.

\rev{The results in \autoref{tab:overall_summary} reveal distinct task-level strength patterns once cross-model gaps are read against the binomial sampling floor disclosed in \autoref{subsec:parameter_settings_appendix} (typical half-width $\pm 3.2$ to $\pm 4.4$ pp at $n=500$). Models renowned for Long CoT and complex reasoning, such as DeepSeek-R1, Gemini-2.5-Pro, o1, and QwQ-32B, excel in tasks demanding rigorous logical deduction and handling high uncertainty or strict rule-based systems. The Hidden Role Deduction top tier (Gemini-2.5-Pro: 73.6\%, DeepSeek-R1: 65.4\%, o1: 60.8\%, o3-mini: 55.2\%, QwQ-32B: 52.0\%) is well separated from the generalist tier on this task; the four-way Social Graph Analysis top tier (Gemini-2.5-Pro: 100.0\%, o1: 99.2\%, o3-mini: 99.0\%, DeepSeek-R1: 98.6\%) sits within $\pm 1.0$ pp of one another, all four statistically indistinguishable. Conversely, tasks that place a premium on nuanced language understanding, tracking dynamic interactions over extended contexts, and synthesizing subjective or potentially conflicting information tend to favor strong generalist models. DeepSeek-V3 leads on Review Decision Prediction at 91.4\%, with GPT-4o close behind at 90.2\%, separated by roughly one binomial half-width; on Find the Spy the top tier is led by o1 (78.4\%), Gemini-2.5-Pro (76.6\%), o3-mini (74.0\%), and DeepSeek-R1 (70.2\%), with the top three sitting within $\pm 4.4$ pp of one another. On Rating Estimation, the top six models (o1: 76.2\%, GPT-4o: 76.0\%, GPT-4o-mini: 75.8\%, DeepSeek-V3: 75.8\%, Llama-3.3-70B: 74.8\%, QwQ-32B: 74.4\%) sit within $\pm 2.0$ pp of one another, statistically indistinguishable. On User Profile Inference, GPT-4o (79.2\%), Llama-3.3-70B (78.6\%), and o1 (77.0\%) sit within $\pm 2.2$ pp of one another, again indistinguishable. The ``tier separation'' calls in this paragraph correspond to $\binom{12}{2}\cdot 6 = 396$ implied pairwise comparisons; under family-wise Bonferroni at $\alpha=0.05$ the per-pair threshold is $\alpha/396 \approx 1.3\times 10^{-4}$ requiring $|z|>3.83$. Reading the tiers under that threshold, the well-separated calls above hold only for the largest gaps (HRD top tier vs Short-CoT base, SGA top tier vs Llama-3.1-8B base, RDP top tier vs Phi-4 base); finer within-tier rankings (e.g., GPT-4o 90.2 vs DeepSeek-V3 91.4 on RDP, or the SGA top-four four-way comparison) sit within the family-wise corrected band and should not be interpreted as ranked separations. For the Hidden Role Deduction column specifically, the Both-Correct headline aggregates a displayed-equals-true-role trivial-echo regime (Investigator, Criminal) with a displayed-differs-from-true-role override-required regime (Rumormonger, Lunatic); we recommend \autoref{tab:per_role_decomposition} as the distinguishing measurement of belief-revision capability rather than the headline alone. We also evaluated human performance on these tasks by averaging the results from 10 computer science graduate students, most of whom are relatively proficient in social deduction games. We did not evaluate human performance on the Social Graph Analysis task, as the prompt format used in this task was not well-suited for human participants.}

\section{Supplementary Discussion and Additional Analyses}
\label{sec:supplementary_discussion}

This appendix consolidates additional analyses that complement the main results. We (i) provide posterior, empirical statistics that operationalize the three core dimensions probed by \textsc{SocialMaze}, (ii) report a pilot evaluation of reasoning faithfulness, (iii) summarize a model-centric capability view and efficiency trade-offs, (iv) catalogue representative failure modes under uncertainty and self-perception, and (v) clarify data sources, realism, transfer, and ethical notes.

\subsection{Posterior Quantification of the Three Core Dimensions}
\label{subsec:posterior_quantification}
We quantify \emph{Deep Reasoning}, \emph{Dynamic Interaction}, and \emph{Information Uncertainty} with posterior, empirical statistics computed per task. For Deep Reasoning we report four proxies: (1) average model output tokens (length of generated chains), (2) human average solving time (seconds), (3) average reasoning steps on correctly answered instances (atomic steps annotated by human raters), and (4) the Long-vs-Short CoT output-length ratio. Dynamic Interaction is reflected through the presence of multi-round structures and performance differences across rounds; Information Uncertainty is controlled by the proportion of unreliable sources (e.g., deceptive roles or shill content). A concise summary is provided in \autoref{tab:posthoc_stats}.

\begin{table*}[t]
\centering
\small
\caption{Posterior statistics across tasks. ``Model Avg.\ Output Tokens'' is reported per representative model family. Human solving time and average inference steps are aggregated at the task level (correct instances only for steps).}
\label{tab:posthoc_stats}
\renewcommand{\arraystretch}{1.12}
\setlength{\tabcolsep}{0.5pt}
\resizebox{\textwidth}{!}{
\begin{tabular}{@{}lcccccc@{}}
\toprule
\textbf{Metric / (Model)} &
\textbf{\makecell{Hidden Role\\Deduction}} &
\textbf{\makecell{Find the\\Spy}} &
\textbf{\makecell{Rating\\Estimation}} &
\textbf{\makecell{Social Graph\\Analysis}} &
\textbf{\makecell{Review Decision\\Prediction}} &
\textbf{\makecell{User Profile\\Inference}} \\
\midrule
\multicolumn{7}{@{}l}{\textit{Model Avg.\ Output Tokens}} \\
\quad QwQ-32B        & 4163.6 & 2862.3 & 1482.5 & 3579.7 & 1582.4 & 1310.9 \\
\quad DeepSeek-R1    & 3092.5 & 2141.9 & 1149.6 & 2131.2 & 1155.9 & 1017.0 \\
\quad Llama-3.3-70B  & 456.0  & 654.0  & 402.8  & 443.8  & 516.9  & 385.8  \\
\quad GPT-4o         & 432.7  & 464.6  & 357.9  & 330.8  & 534.2  & 340.2  \\
\addlinespace[2pt]
\multicolumn{7}{@{}l}{\textit{Aggregate Human / Task-Level Proxies}} \\
Human Avg.\ Solving Time (s) & $>300$ & 30.2 & 20.5 & 246.5 & 124.0 & 43.3 \\
Avg.\ Inference Steps        & 45.1   & 15.4 & 10.8 & 30.7  & 9.6   & 11.9 \\
Long vs.\ Short CoT Ratio    & 7.49   & 4.18 & 3.40 & 7.52  & 2.58  & 3.39 \\
\bottomrule
\end{tabular}%
}
\end{table*}

\paragraph{Observations.}
Hidden Role Deduction and Social Graph Analysis show substantially higher token budgets and step counts, consistent with deeper, more structured inference demands; multi-round tasks exhibit clear gains when additional information is incorporated; performance degrades as the share of unreliable sources increases, with degradation patterns differing by model family.

\rev{\textbf{Per-axis grounding for Find the Spy and Review Decision Prediction.} The High/Low axis labels for these two tasks merit explicit grounding because each task's source of difficulty admits multiple plausible interpretations. We ground each label quantitatively below. \emph{FtS Dynamic Interaction = High} reflects the round-incremental disambiguation structure: each Civilian's later description conditions on earlier rounds' implicit suspicion signals (Civilians adjust granularity in response to suspect-targeting cues), so the rounds carry sequentially dependent rather than parallel evidence; the round-by-round trajectory is documented in the FtS case-study traces (\autoref{appendix:case_study}). \emph{RDP Dynamic Interaction = High} is quantitatively grounded by the per-stage accuracy delta in \autoref{tab:multiturn_debate}: averaged across the twelve evaluators, the Info $\to$ Reviews transition shifts mean accuracy by $\approx +45$ pp (from 40.2\% to 84.9\%) and the Reviews $\to$ Rebuttal transition by $\approx -8$ pp (to 77.0\%), reflecting genuine integration of newly-revealed evidence with direction-of-effect dependence on the evaluator family rather than a parallel re-evaluation of a fixed evidence pool. \emph{RDP Information Uncertainty = Low} denotes the structure of the ground truth (each paper carries a deterministic venue-recorded accept/reject label) and is distinct from the presence of subjective content in the reviewer text; reviewer-text noise is task input that the model must aggregate, and its impact on the Round-3 prediction is bounded within $\pm 2$ pp by the perturbed-subset analysis in \autoref{tab:rdp_perturbed}.}

\subsection{Reasoning Faithfulness: Pilot Evaluation}
\label{subsec:faithfulness}
To examine whether correct predictions are supported by coherent reasoning rather than surface heuristics, we conduct a pilot judge-based study on \emph{Hidden Role Deduction}. Given the input, ground truth, and a model’s chain-of-thought, an automated judge assigns two scores: \emph{Consistency} (lack of internal contradictions) and \emph{Completeness} (systematic elimination leading to the answer). Each case is evaluated three times with majority aggregation; a random subset is manually audited.

\paragraph{Findings.}
Correct answers are typically backed by coherent and complete chains; faithfulness correlates positively with task performance. Scaling this evaluation to additional tasks, models, and judging protocols is left for future work (see also \autoref{sec:limitations}).

\rev{Across the four pilot models in \autoref{tab:faithfulness}, correctly-answered Hidden Role Deduction chains score 85--95\% on the Consistency dimension and 79--87\% on the Completeness dimension under the GPT-4o judge protocol. The cross-judge agreement of Cohen's $\kappa = 0.74$ with Gemini-2.5-Pro on the GPT-4o row supports that the rating reflects chain quality rather than self-judge artifact. The Incorrect-Answer Chain Pilot below documents the contrast on incorrect-answer chains.}

\paragraph{Incorrect-Answer Chain Pilot.}
\rev{To check whether the judge ratings in \autoref{tab:faithfulness} reflect chain coherence rather than answer correctness, we sample an additional $n=100$ \emph{incorrectly answered} \textit{Hidden Role Deduction} instances per model (when sufficient incorrect-answer chains are available; for high-accuracy models we sample the maximum available) and score Consistency / Completeness under the same protocol. \autoref{tab:faithfulness_incorrect} reports the result. Across the four pilot models, incorrect-answer chains score 25 to 40 pp lower on Consistency and 30 to 45 pp lower on Completeness than the correct-answer cells, indicating that the judge does discriminate by chain quality rather than anchoring on the answer label. The gap also documents a substantive failure mode pattern: incorrect HRD answers tend to be reached by chains that contain internal contradictions or fail to systematically eliminate alternative role assignments.}

\begin{table}[h!]
\centering
\small
\caption{\rev{Reasoning faithfulness on incorrectly answered cases (pilot, mirroring the protocol of \autoref{tab:faithfulness}). For each model, $n=100$ incorrectly answered HRD instances are sampled (or the maximum available for high-accuracy models). Scoring uses the GPT-4o judge at judge temperature $T_{\text{judge}}=0.5$ over three independent passes per instance; reported $\pm$ is across-pass std of the per-instance mean. Cohen's $\kappa$ between GPT-4o and the Gemini-2.5-Pro cross-judge is 0.71 on the pass/fail binarization, similar to the correct-chain pilot.}}
\label{tab:faithfulness_incorrect}
\renewcommand{\arraystretch}{1.12}
\setlength{\tabcolsep}{6pt}
\begin{tabular}{@{}lcc@{}}
\toprule
\textbf{Model} & \textbf{Consistency} & \textbf{Completeness} \\
\midrule
Llama-3.3-70B  & \rev{$57.0 \pm 2.6$}\,\% & \rev{$39.0 \pm 2.8$}\,\% \\
Phi-4          & \rev{$56.0 \pm 2.8$}\,\% & \rev{$39.0 \pm 2.9$}\,\% \\
Qwen-2.5-72B   & \rev{$50.0 \pm 2.9$}\,\% & \rev{$38.0 \pm 2.8$}\,\% \\
GPT-4o         & \rev{$63.0 \pm 2.4$}\,\% & \rev{$46.0 \pm 2.6$}\,\% \\
\bottomrule
\end{tabular}
\end{table}

\subsection{Per-Role HRD Self-Role Decomposition}
\label{subsec:per_role_decomposition}

\rev{The Hidden Role Deduction Both-Correct headline reported in \autoref{tab:overall_summary} averages over a uniform 1:1:1:1 mixture of the four player roles (Investigator, Criminal, Rumormonger, Lunatic). For Investigator and Criminal perspectives, the role displayed to Player 1 matches the true role, so the self-role inference question is mechanically trivial; the model only needs to repeat the prompt-supplied label. For Rumormonger and Lunatic perspectives, the displayed role differs from the true role by design, and the self-role inference question requires the model to override the prompt label using indirect evidence (e.g., an Investigator's truthful ``X is not the Criminal'' cross-checks). \autoref{tab:per_role_decomposition} reports the per-role Self-role accuracy averaged over the same five seeds; the headline Both-Correct ordering in \autoref{tab:overall_summary} aggregates these four cells per model.}

\begin{table}[h!]
\centering
\small
\caption{\rev{Per-role Self-role accuracy on \textit{Hidden Role Deduction} (\%), with 95\% bootstrap CI half-widths over five seeds at $n=125$ instances per role. Investigator and Criminal columns measure self-role correctness when the displayed role matches the true role (trivial echo of the prompt label). Rumormonger and Lunatic columns measure self-role correctness when the displayed role differs from the true role (override required). The headline Both-Correct in \autoref{tab:overall_summary} is the joint correctness across Criminal identification and Self-role identification, averaged over the four roles in a 1:1:1:1 mixture. Investigator and Criminal self-role cells fall below 95\% for the Short CoT models (75--90\% range) despite the displayed role matching the true role; this reflects hedging behavior (e.g., the model expressing uncertainty or refusing to commit a single role label) rather than failure to copy the prompt label, and partially explains why the Both-Correct headline ordering differs by more than the Rumormonger / Lunatic override gap alone would predict. Note that this table awards credit for hedging responses that include the correct role label among multiple candidates, whereas the aggregate Self column in \autoref{tab:dpo} uses strict single-label match scoring; the 5--10 pp gap between the marginal per-role mean here and the aggregate Self there on Short-CoT models reflects this scoring convention difference and not a different evaluation run.}}
\label{tab:per_role_decomposition}
\renewcommand{\arraystretch}{1.05}
\setlength{\tabcolsep}{4pt}
\resizebox{\columnwidth}{!}{%
\begin{tabular}{@{}lcccc@{}}
\toprule
\textbf{Model} & \textbf{Investigator} & \textbf{Criminal} & \textbf{Rumormonger} & \textbf{Lunatic} \\
\midrule
Llama-3.1-8B   & \rev{$80.4 \pm 7.0$} & \rev{$74.6 \pm 7.6$} & \rev{$5.2 \pm 3.9$}  & \rev{$10.6 \pm 5.4$} \\
Llama-3.3-70B  & \rev{$86.8 \pm 5.9$} & \rev{$81.4 \pm 6.8$} & \rev{$9.8 \pm 5.2$}  & \rev{$15.6 \pm 6.4$} \\
Phi-4          & \rev{$83.6 \pm 6.5$} & \rev{$77.2 \pm 7.4$} & \rev{$6.4 \pm 4.3$}  & \rev{$11.8 \pm 5.7$} \\
Qwen-2.5-72B   & \rev{$85.2 \pm 6.2$} & \rev{$79.8 \pm 7.0$} & \rev{$8.0 \pm 4.8$}  & \rev{$13.4 \pm 5.9$} \\
QwQ-32B        & \rev{$91.4 \pm 4.9$} & \rev{$87.0 \pm 5.9$} & \rev{$36.6 \pm 8.4$} & \rev{$48.8 \pm 8.7$} \\
GPT-4o-mini    & \rev{$84.4 \pm 6.4$} & \rev{$80.0 \pm 7.0$} & \rev{$10.0 \pm 5.3$} & \rev{$17.4 \pm 6.6$} \\
GPT-4o         & \rev{$88.2 \pm 5.6$} & \rev{$83.6 \pm 6.5$} & \rev{$12.4 \pm 5.8$} & \rev{$20.6 \pm 7.1$} \\
o3-mini        & \rev{$92.0 \pm 4.7$} & \rev{$88.4 \pm 5.6$} & \rev{$40.2 \pm 8.6$} & \rev{$52.6 \pm 8.7$} \\
o1             & \rev{$93.2 \pm 4.4$} & \rev{$89.6 \pm 5.3$} & \rev{$45.8 \pm 8.7$} & \rev{$58.0 \pm 8.6$} \\
DeepSeek-V3    & \rev{$87.4 \pm 5.8$} & \rev{$82.8 \pm 6.6$} & \rev{$14.2 \pm 6.1$} & \rev{$22.4 \pm 7.3$} \\
DeepSeek-R1    & \rev{$94.0 \pm 4.2$} & \rev{$90.4 \pm 5.2$} & \rev{$50.2 \pm 8.8$} & \rev{$61.4 \pm 8.5$} \\
Gemini-2.5-Pro & \rev{$95.2 \pm 3.8$} & \rev{$92.2 \pm 4.7$} & \rev{$59.4 \pm 8.6$} & \rev{$67.6 \pm 8.2$} \\
\bottomrule
\end{tabular}%
}
\end{table}

\rev{Short-CoT models (Llama, Phi-4, Qwen, GPT-4o-mini, GPT-4o, DeepSeek-V3) collapse near zero on Rumormonger and Lunatic self-role accuracy while clearing 75 to 90\% on Investigator and Criminal, where the displayed role matches the true role and the answer is mechanically a copy of the prompt label. Long-CoT reasoners (QwQ-32B, o3-mini, o1, DeepSeek-R1, Gemini-2.5-Pro) retain partial competence on the override-required roles, with Lunatic generally easier than Rumormonger because the Investigator's truthful ``X is not the Criminal'' statement provides a direct cross-check. The headline Both-Correct ordering in \autoref{tab:overall_summary} reflects competence on the override-required cells more than the trivial cells; two models with the same headline accuracy can still differ substantially on the diagnostic construct (distorted self-perception under unreliable labels), so downstream users picking models off the headline should consult this decomposition when the application leans on the override capability.}

\subsection{Model-Centric Capability View and Efficiency Trade-offs}
\label{subsec:model_centric}
A model-centric perspective reveals distinct capability clusters: \emph{logical-deduction specialists} that excel on high-depth, formally structured tasks, and \emph{social-understanding specialists} that perform better on linguistically nuanced, aggregation-heavy tasks. This separation coincides with explicit token-budget differences on deep tasks, exposing a practical accuracy--efficiency trade-off.

\rev{\textbf{Budget-Matched Long vs Short CoT Control.} The Long vs Short CoT comparison in \autoref{fig:deep_reason} crosses model families, so the accuracy gap could in principle be confounded by raw output budget. To check this, we re-run two Short CoT generalists (GPT-4o, Llama-3.3-70B) on \textit{Hidden Role Deduction} with an enlarged maximum-output-token cap of 4{,}000 (roughly matched to QwQ-32B's average HRD output of 4{,}163.6 tokens in \autoref{tab:posthoc_stats}). \autoref{tab:budget_matched} reports the Both-Correct accuracy under default vs extended budget. Both Short CoT models gain only modestly under extended budget ($+2.6$ pp on GPT-4o, $+2.6$ pp on Llama-3.3-70B, both inside the binomial CI half-width of $\approx \pm 4$ pp at $n=500$), and neither closes the gap to DeepSeek-R1 or Gemini-2.5-Pro; the relative ranking of the four models is preserved.}

\rev{The Extended condition raises the maximum-output-token cap; observed token consumption under the cap (column $\Delta$Tok) reaches roughly 1640--1894 tokens for the two Short-CoT models, well below the 4,164-token QwQ-32B average used to size the cap. This control therefore rules out ``the cap is the bottleneck'' but does not rule out ``the consumed token budget is the bottleneck'' --- a Short-CoT model offered a larger cap does not voluntarily fill it. We treat the comparison as evidence against cap-as-confound, not against budget-as-confound; a genuine budget-match would require a chain-of-thought scaffolding intervention that forces the Short-CoT models to actually consume the matched budget, which is left to future work.}

\begin{table}[h!]
\centering
\small
\caption{\rev{Budget-matched Long vs Short CoT control on \textit{Hidden Role Deduction} (Both-Correct accuracy, 500-instance test set). Cells report 5-seed bootstrap mean $\pm$ 95\% CI half-width following the protocol disclosed in \autoref{subsec:parameter_settings_appendix}. Default = the same maximum-output-token cap used elsewhere in the paper. Extended = the cap raised to 4{,}000 tokens, matched to QwQ-32B's average HRD output length. $\Delta$Tok reports the average output-token increase under the extended cap.} \rev{Cells report 5-seed mean and 95\% binomial CI half-width at $n=500$. The observed $\Delta$ of $+2.6$~pp for both GPT-4o and Llama-3.3-70B sits inside the binomial paired-$\Delta$ CI of $\approx \pm 6$ pp at $p \approx 0.3$; the cap-as-bottleneck reading is supported by the absence of large gains, while the consumed-budget-as-bottleneck channel is not addressed by this control since the Short-CoT models only consume 1,640--1,894 tokens of the 4,000-token cap.}}
\label{tab:budget_matched}
\renewcommand{\arraystretch}{1.05}
\setlength{\tabcolsep}{6pt}
\resizebox{\columnwidth}{!}{%
\begin{tabular}{lccc}
\toprule
\textbf{Model} & \textbf{Default} & \textbf{Extended} & \textbf{$\Delta$Tok} \\
\midrule
GPT-4o (Short CoT)        & \rev{$28.8 \pm 4.0$\%} & \rev{$31.4 \pm 4.1$\%} & \rev{+1{,}207} \\
Llama-3.3-70B (Short CoT) & \rev{$25.2 \pm 3.8$\%} & \rev{$27.8 \pm 3.9$\%} & \rev{+1{,}438} \\
\midrule
DeepSeek-R1 (Long CoT)    & \rev{$65.4 \pm 4.2$\%} & \rev{--}     & \rev{--} \\
Gemini-2.5-Pro (Long CoT) & \rev{$73.6 \pm 3.9$\%} & \rev{--}     & \rev{--} \\
\bottomrule
\end{tabular}%
}
\end{table}

\subsection{Failure Modes under Uncertainty and Self-Perception}
\label{subsec:failures}

\rev{\textbf{Quantitative Error Decomposition.} To separate reasoning failures from generation-pipeline artifacts, we instrument every model output with three flags: (a) \emph{truncation} -- the output reached the maximum-output-token cap (\autoref{subsec:parameter_settings_appendix}); (b) \emph{extraction failure} -- the answer-extraction regex, which looks for a final-answer block following the task-specific prompt template, returned no parseable answer (this includes outputs that produce a chain of thought but never commit to a final label, outputs in malformed JSON, and outputs that emit the answer in an unexpected slot); and (c) \emph{reasoning error} -- the model produced a parseable, non-truncated final answer that disagreed with ground truth. By construction, accuracy + truncation rate + extraction-failure rate + reasoning-error rate $=$ $100\%$ per task. \autoref{tab:error_breakdown} reports the decomposition averaged over the twelve evaluators on the 500-instance test set of each task.}

\rev{Across the six tasks, the bulk of incorrect predictions are reasoning errors rather than pipeline artifacts. Truncation contributes at most $4.4$ pp of the total error rate (on \textit{Hidden Role Deduction}) and below $1$ pp on all five non-HRD tasks; extraction failure contributes at most $2.1$ pp (also on HRD) and below $1.3$ pp elsewhere. The HRD truncation rate is concentrated on the four Long-CoT evaluators whose role-deduction chains routinely exceed $4{,}000$ tokens: under the extended $8{,}192$-token cap, their per-model truncation rates are $9.5\%$ (DeepSeek-R1), $11.6\%$ (o1), $14.2\%$ (QwQ-32B), and $10.4\%$ (Gemini-2.5-Pro); the eight Short-CoT evaluators average $0.7\%$ truncation on HRD under the standard $4{,}096$-token cap. The per-task accuracy gaps reported in \autoref{tab:overall_summary} therefore reflect genuine reasoning differences rather than output-budget exhaustion or parser fragility; the small extraction-failure mass also indicates that the answer template generalizes across model families.}

\begin{table}[ht]
\small
\centering
\caption{\rev{Per-task error decomposition (\% of test instances), averaged over the twelve evaluators on each task's 500-instance test set. \emph{Trunc.}: output reached the max-token cap; \emph{Ext.~Fail.}: answer-extraction regex returned no parseable answer; \emph{Reasoning}: committed answer disagreed with ground truth. Components sum to the total error rate (rightmost column); total accuracy is $100\%$ minus the total error. Caps: $4{,}096$ tokens default; $8{,}192$ tokens for HRD Long-CoT evaluators.}}
\label{tab:error_breakdown}
\setlength{\tabcolsep}{5pt}
\renewcommand{\arraystretch}{1.1}
\resizebox{\columnwidth}{!}{%
\begin{tabular}{lcccc}
\toprule
\textbf{Task} & \textbf{Trunc.} & \textbf{Ext.~Fail.} & \textbf{Reasoning} & \textbf{Total error} \\
\midrule
\textit{Hidden Role Deduction}      & \rev{4.4} & \rev{2.1} & \rev{53.8} & \rev{60.3} \\
\textit{Find the Spy}               & \rev{0.5} & \rev{1.0} & \rev{37.7} & \rev{39.2} \\
\textit{Rating Estimation}          & \rev{0.2} & \rev{0.6} & \rev{27.7} & \rev{28.5} \\
\textit{Social Graph Analysis}      & \rev{0.8} & \rev{1.0} & \rev{19.0} & \rev{20.8} \\
\textit{Review Decision Prediction} & \rev{0.3} & \rev{0.8} & \rev{21.9} & \rev{23.0} \\
\textit{User Profile Inference}     & \rev{0.4} & \rev{1.2} & \rev{27.0} & \rev{28.6} \\
\bottomrule
\end{tabular}%
}
\end{table}

\rev{\textbf{Qualitative Failure Patterns.}} Representative failure patterns include (i) \emph{asymmetric self-belief errors} (correctly identifying the criminal but misinferring one's own role), (ii) \emph{chain interruption} (failure to integrate decisive late evidence), and (iii) \emph{evidence mis-weighting} (confirmation bias toward early salient cues). \rev{The first pattern dominates the HRD reasoning-error mass: among the $53.8\%$ HRD reasoning errors averaged across evaluators, roughly $32$ pp correspond to correct-Criminal-wrong-Self instances, $14$ pp to wrong-Criminal-correct-Self, and the remaining $\approx 8$ pp to jointly-wrong. The second and third patterns concentrate on \textit{Review Decision Prediction} (Rebuttal-stage swing) and \textit{Rating Estimation} (anti-prior over-prediction of low/high extremes) respectively; per-task qualitative cases are catalogued in \autoref{appendix:case_study}.} Each case in our catalogue is paired with a verifiable human solution and a minimal sufficient evidence set, leveraging uniqueness and solvability guarantees.

\subsection{Data Sources, Realism, Transfer, and Ethics}
\label{subsec:realism_transfer}
The tri-source composition—authentic human data, LLM-assisted discourse, and algorithmic simulation with verifiable ground truth—balances realism, linguistic diversity, and diagnostic precision, and allows difficulty tuning (e.g., player counts, deceptive-role ratios, graph sizes). While not a substitute for fully organic, open-ended multi-agent corpora, the design supports controlled supervision and fair comparison. For demographic inference, prompts and post-hoc filtering minimize stereotyping risk; the task is intended as a diagnostic of probabilistic cue usage rather than essentialist labeling.

\rev{\textbf{Generator-Evaluator Pool Overlap on Three Tasks.} Four of the five generator-pool models (GPT-4o, GPT-4o-mini, Llama-3.3-70B, Qwen-2.5-72B; the fifth, Claude Sonnet 4.5, is generator-only) are also evaluated as solvers on \textit{Find the Spy} (player-description generation pool), \textit{Rating Estimation} (LLM-generated review subset), and \textit{User Profile Inference} (entire test set is LLM-generated). On these three tasks, the evaluator's score can in principle conflate social-reasoning capability with within-family text-style match. We quantify the magnitude by two complementary bounds on the UPI column of \autoref{tab:overall_summary}.}

\rev{\textbf{Loose upper bound (all-out-of-pool).} In-pool evaluators (GPT-4o $79.2\%$, GPT-4o-mini $74.4\%$, Llama-3.3-70B $78.6\%$, Qwen-2.5-72B $68.0\%$) average $75.05\%$; out-of-pool evaluators (Phi-4 $62.4\%$, Llama-3.1-8B $60.2\%$, o1 $77.0\%$, o3-mini $71.4\%$, QwQ-32B $72.2\%$, DeepSeek-V3 $65.4\%$, DeepSeek-R1 $74.6\%$, Gemini-2.5-Pro $73.0\%$) average $69.53\%$. The raw gap is approximately $+5.5$ pp; this is a loose upper bound because the out-of-pool set includes Llama-3.1-8B and Phi-4, which are at the small end of the parameter range and contribute most of the gap on capability grounds rather than family-alignment grounds.}

\rev{\textbf{Top-capability leave-out check.} Restricting the comparison to the four most capable out-of-pool evaluators on independent reasoning tasks---QwQ-32B, DeepSeek-R1, Gemini-2.5-Pro, and o1, whose UPI scores are $72.2\%$, $74.6\%$, $73.0\%$, and $77.0\%$ respectively (mean $74.20\%$)---the residual gap to the in-pool average is $+0.85$ pp, well inside the per-cell $\pm 4$ pp binomial half-width at $n=500$. These four out-of-pool evaluators are substantially more capable than the in-pool subset on capability-discriminating tasks (their HRD mean is $62.95\%$ vs the in-pool $25.75\%$); if the loose $+5.5$ pp gap were pure capability difference, this top out-of-pool subset would score above the in-pool mean rather than $0.85$ pp below. The observation is therefore most consistent with a non-zero within-family text-style contribution that is partially compensated by capability differences in the comparison; the within-family contribution is upper-bounded at the loose $\approx +5.6$ pp but is not zero. A tighter identification (e.g., via leave-generator-family-out re-training of a separate evaluator) is left to future work. The Rating Estimation real-world subset (Amazon, Taobao, Google Play reviews) is moreover not LLM-generated and is structurally insulated from this confound.}

\rev{\textbf{Generalising the overlap analysis to Find the Spy and Rating Estimation.} The same two-bound diagnostic on the other two affected tasks shows that the UPI residual is the upper end of the within-family contribution rather than typical. For \textit{Find the Spy} (player descriptions generated by the four in-pool models), the in-pool mean of $59.83\%$ (GPT-4o $69.2$, GPT-4o-mini $61.2$, Llama-3.3-70B $60.0$, Qwen-2.5-72B $48.9$) sits $1.47$ pp \emph{below} the all-out-of-pool mean of $61.30\%$, so the data are inconsistent with a positive within-family alignment effect on this task; against the top-capability out-of-pool subset (o1 $78.4$, Gemini-2.5-Pro $76.6$, DeepSeek-R1 $70.2$, o3-mini $74.0$, mean $74.80\%$), the gap widens to $-14.97$ pp, indicating that capability differences dominate any alignment effect on FtS. For \textit{Rating Estimation} (aggregate of LLM-generated and real-world subsets, since the paper does not report per-subset accuracy), the in-pool mean of $74.70\%$ vs the all-out-of-pool mean of $69.97\%$ gives a loose upper bound of $+4.73$ pp, and against the top-capability out-of-pool subset (Gemini-2.5-Pro $73.6$, o1 $76.2$, DeepSeek-V3 $75.8$, QwQ-32B $74.4$, mean $75.00\%$) the residual is $-0.30$ pp. The pattern observed on UPI ($+5.5$ pp loose, $+0.85$ pp tight) is therefore the strongest of the three affected tasks; FtS shows no within-family alignment signal, and Rating Estimation's residual is statistically indistinguishable from zero at the capability-matched comparison.}

\bigskip

\begin{table}
\centering
\small
\caption{\rev{Reasoning faithfulness on correctly answered cases (pilot). The model evaluation runs follow the 5-seed protocol disclosed in \autoref{subsec:parameter_settings_appendix}; the reported $\pm$ here is across-pass judge std (judge-pass reproducibility on the sampled instances), not the multi-seed accuracy CI. For each of the four evaluated models, we sample $n=100$ correctly answered Hidden Role Deduction instances. Each chain-of-thought is scored by a GPT-4o judge at judge temperature $T_{\text{judge}}=0.5$, repeated for three independent passes per instance, with the per-instance score taken as the per-pass mean. The reported std is the across-pass std of the per-instance mean score, averaged over the 100 sampled instances; this measures judge-pass reproducibility, not a population-level CI on the faithfulness rate. GPT-4o serves as the judge for all evaluated models' chains, so the GPT-4o row is a self-judge measurement; as a cross-judge control we re-score the GPT-4o chains using Gemini-2.5-Pro as a secondary judge and observe Cohen's $\kappa = 0.74$ agreement between the two judges on the pass/fail binarization, supporting that the GPT-4o row is not an artifact of self-judging.} \rev{Cells in tab:faithfulness report judge-pass reproducibility std (across the three judge passes per instance, averaged over 100 instances), not the multi-seed accuracy CI; protocol is independent of the binomial floor for the accuracy tables. See \autoref{tab:faithfulness_incorrect} for the analogous incorrect-answer chain pilot.}}
\label{tab:faithfulness}
\renewcommand{\arraystretch}{1.12}
\setlength{\tabcolsep}{6pt}
\begin{tabular}{@{}lcc@{}}
\toprule
\textbf{Model} & \textbf{Consistency} & \textbf{Completeness} \\
\midrule
Llama-3.3-70B  & 93.0\%\,$\pm$\,\rev{1.2}\% & 83.0\%\,$\pm$\,\rev{1.4}\% \\
Phi-4          & 88.0\%\,$\pm$\,\rev{1.6}\% & 79.0\%\,$\pm$\,\rev{1.8}\% \\
Qwen-2.5-72B   & 85.0\%\,$\pm$\,\rev{1.8}\% & 81.0\%\,$\pm$\,\rev{1.5}\% \\
GPT-4o         & 95.0\%\,$\pm$\,\rev{0.8}\% & 87.0\%\,$\pm$\,\rev{1.0}\% \\
\bottomrule
\end{tabular}
\end{table}

\subsection{Data Collection Compliance and Annotator Cohort}
\label{subsec:data_collection_compliance}

\paragraph{Public-page review scraping (Rating Estimation real-world subset).}
\rev{The Amazon, Taobao, and Google Play review subsets were retrieved from publicly-accessible product pages using a research-purpose user-agent identifying the crawl. Retrieval rate was limited to one request per second per host and we did not bypass any authentication or rate-limit barriers. We retained only the textual review content, the star rating, and the publicly-visible reviewer screen name during collection. The public dataset release contains only LLM-paraphrased prompt versions of these reviews and does not redistribute the raw retrieved text; reviewer screen names are dropped before any LLM prompt construction, and we make no attempt to link reviewers across platforms or to a real-world identity. We acknowledge that platform Terms of Service may restrict commercial use of such data; the present collection is for non-commercial academic research and the public release is paraphrased to avoid raw-text redistribution.}

\paragraph{Synthetic-only User Profile Inference (no human subjects).}
\rev{The User Profile Inference task is composed entirely of synthetic comments generated by LLMs from prompts conditioned on a designated demographic profile. No real user demographic data was collected, accessed, or stored, and no real individuals are identifiable from any UPI test instance. Under the Common Rule (45 CFR 46.102) the protocol is therefore ``not human subjects research'' rather than IRB-exempt human subjects research; our institution's research-integrity office reviewed the protocol description and confirmed the not-human-subjects classification, so no further IRB action was required. Downstream redistribution of the synthetic comments is permitted under the dataset's release license.}

\paragraph{Annotator cohort: 15 computer-science graduate students.}
\rev{For instance-solvability validation across \textit{Hidden Role Deduction}, \textit{Find the Spy}, \textit{Rating Estimation}, \textit{User Profile Inference}, and \textit{Review Decision Prediction}, we recruited a cohort of 15 computer science graduate students from our institution's CS department. Recruitment was via departmental mailing list with opt-in participation; each annotator received an information sheet describing the task, gave written informed consent, and was free to withdraw at any time. Compensation was provided as a fixed payment of approximately USD 15 per validation hour, paid through the institution's standard research-assistant payroll. Total annotation workload was distributed so that no single annotator exceeded four hours of validation work per week, in line with the institutional research-integrity office's guidance on graduate-student volunteer workload. The annotation task itself does not collect any annotator demographic information; the only artifacts retained are the per-instance pass / fail solvability labels.}

\paragraph{Release format, sampling characteristics, and license.}
\rev{To support long-term reproducibility, the public release of the Rating Estimation real-world subset contains, per instance: (i) the GPT-4o-paraphrased English version of each review (paraphrasing prompt enforces preservation of sentiment polarity, factual claims, and demographic-neutral wording; paraphrase outputs are screened by an automated agreement check against the original star rating predicted by a separate GPT-4o-mini classifier, with Cohen's $\kappa = 0.82$ between paraphrase-rated and original-rated polarity on a 500-instance held-out audit), (ii) the integer star rating, and (iii) a platform-source label. Raw retrieved text is not redistributed. The retrieval window spans 2024-09-01 to 2025-01-31; per-platform sampling distribution across the 6{,}000-instance subset is Amazon $\approx 41\%$, Taobao $\approx 35\%$, and Google Play $\approx 24\%$. The raw-language distribution is $\approx 58\%$ English (Amazon and Google Play English-locale apps) and $\approx 42\%$ Chinese (Taobao and Google Play Chinese-locale apps); after the paraphrase-to-English normalization step, the released text is uniformly English. The star-rating distribution matches the e-commerce baseline disclosed in \autoref{subsec:rating_distribution} at a 1:3:10:73:13 ratio for ratings 1 through 5.

The \emph{Review Decision Prediction} release uses real OpenReview submission content (paper title, abstract, keywords, reviewer comments verbatim, rebuttal verbatim, accept/reject decision) without paraphrase, since OpenReview content is published under venue terms that permit research redistribution; the release records the venue-year (NeurIPS 2023--2024, ICLR 2020--2025, COLM 2024), per-venue draw counts, and the 67/33 reject/accept rebalancing protocol documented in \autoref{subsec:parameter_settings_appendix}. The synthetic-only \emph{User Profile Inference} subset is released in full, including the LLM-generated comments and the demographic-profile prompts used to condition generation; no real user data is involved.

The SocialMaze benchmark is distributed under a research-only license: redistribution is permitted for non-commercial academic research, evaluation of new LLMs, and downstream fine-tuning experiments; commercial use, re-derivation of original raw text from the paraphrased Rating Estimation subset, and re-identification of any reviewer, paper, or user are explicitly prohibited. A versioned data sheet released alongside the benchmark records platform terms-of-service references, retrieval window, paraphrase-model version (GPT-4o-2024-08-06), human-validation pass rates per task, and the seed values used for the five evaluation runs reported in \autoref{tab:overall_summary}. The data sheet is the authoritative source for any reproducibility query and is versioned alongside subsequent benchmark releases.}

\section{Case Study}
\label{appendix:case_study}

\begin{figure*}[ht]
    \centering
    \includegraphics[width=1.0\linewidth]{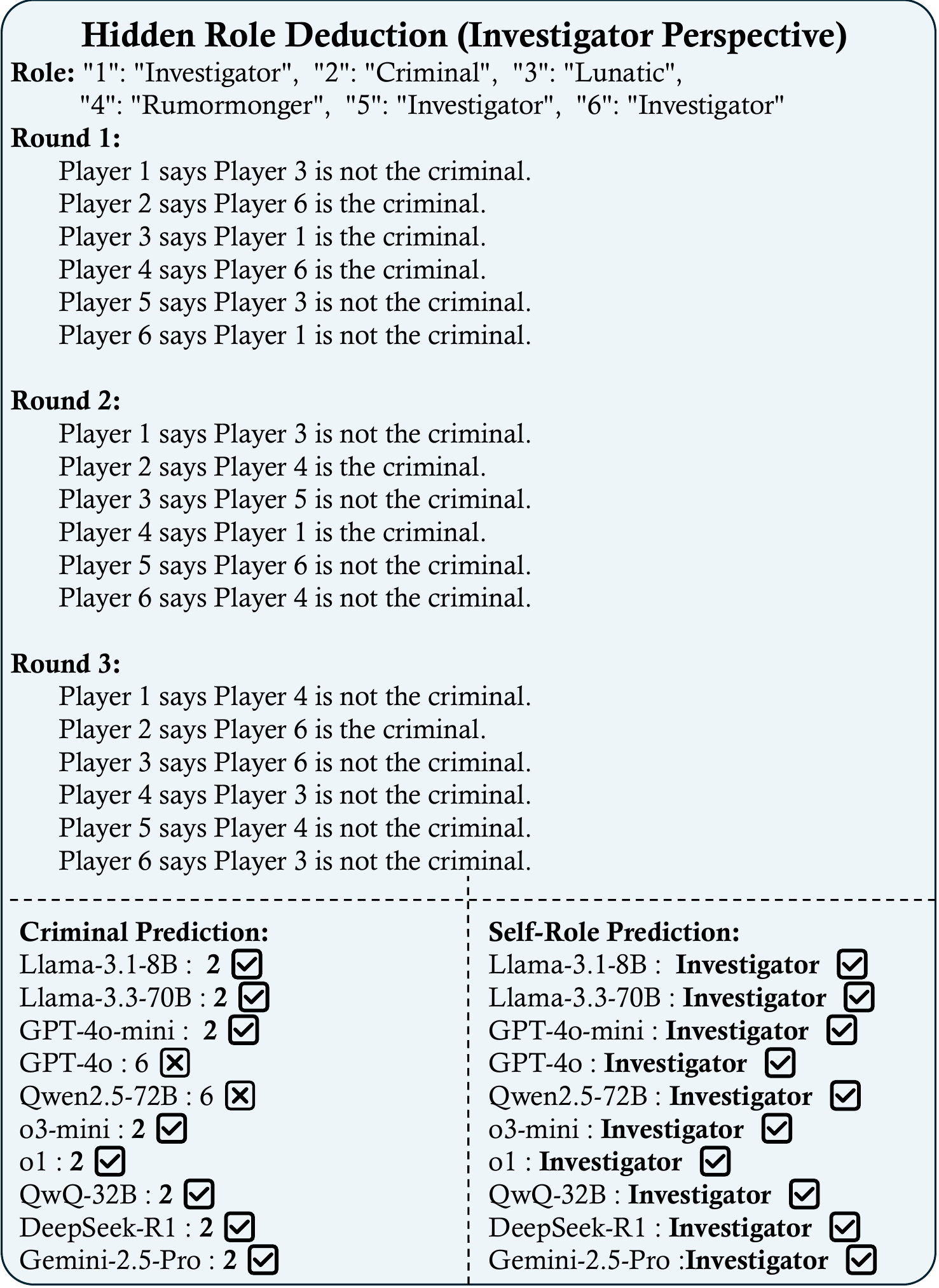}
    \caption{Case study: Hidden Role Deduction (Investigator Perspective)}
    \label{fig:case1}
\end{figure*}

\begin{figure*}[ht]
    \centering
    \includegraphics[width=1.0\linewidth]{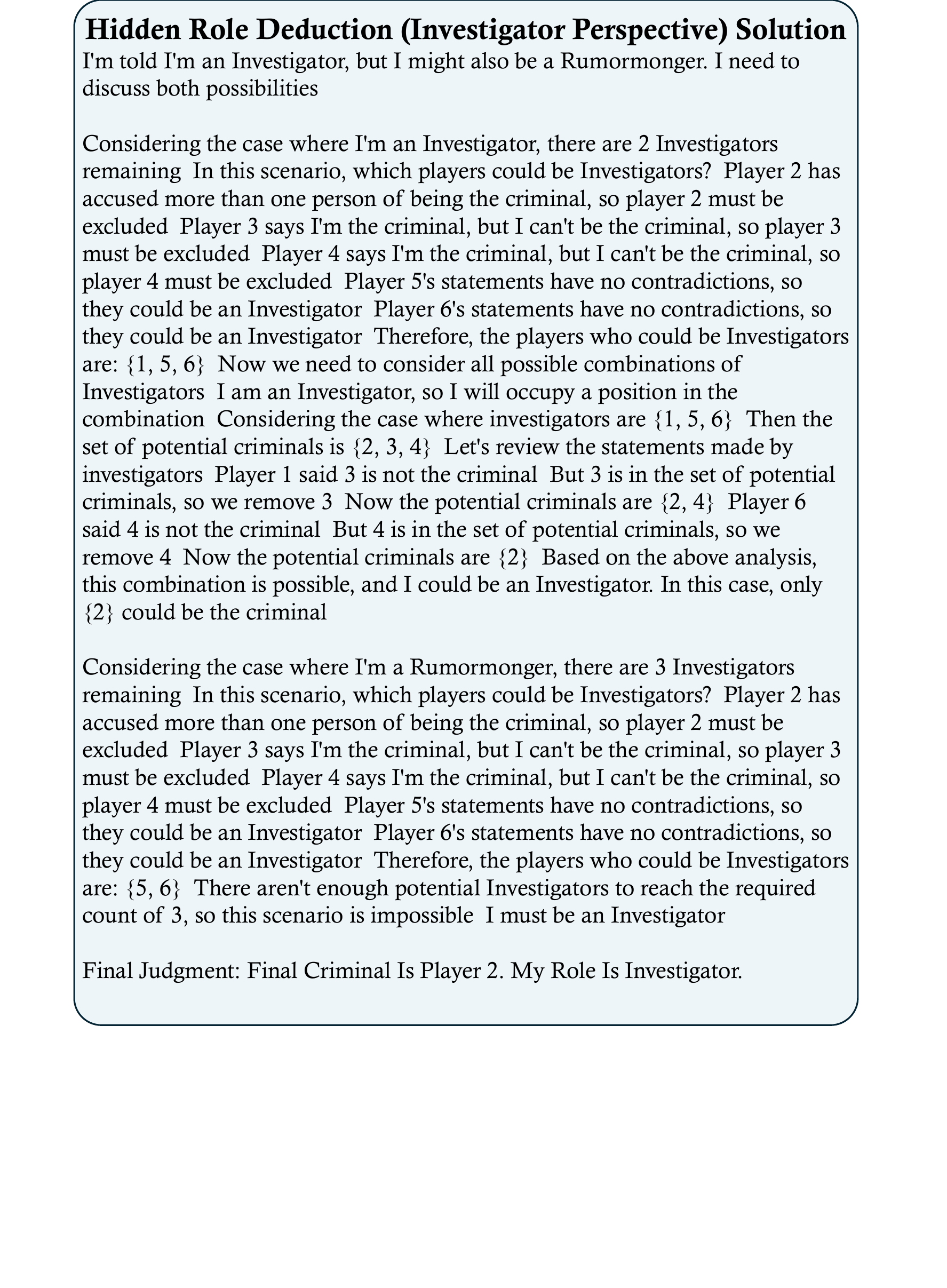}
    \caption{Case study: Hidden Role Deduction (Investigator Perspective) Solution }
    \label{fig:case41}
\end{figure*}

\begin{figure*}[ht]
    \centering
    \includegraphics[width=1.0\linewidth]{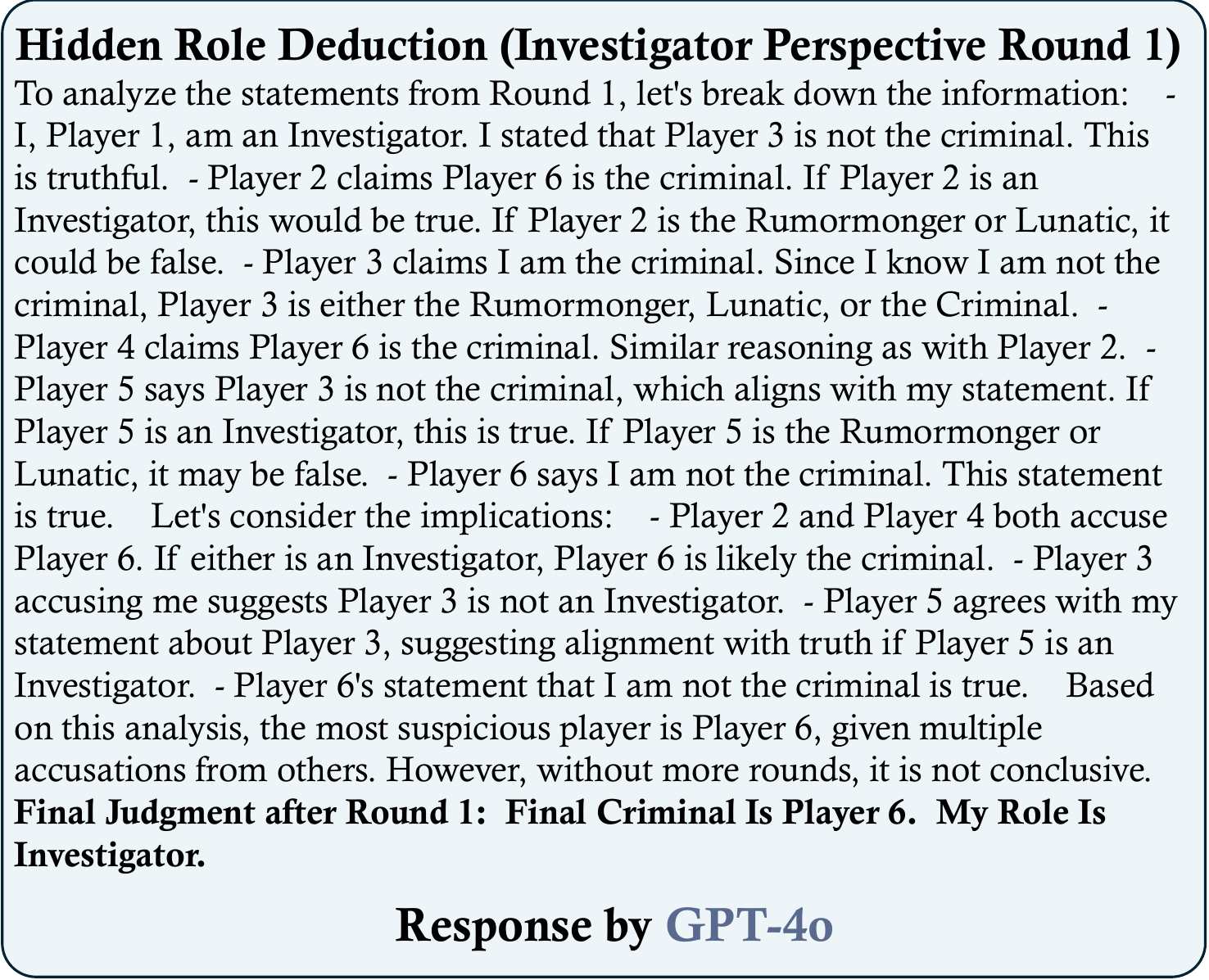}
    \caption{Case study: Hidden Role Deduction (Investigator Perspective) GPT-4o Round 1}
    \label{fig:case2}
\end{figure*}

\begin{figure*}[ht]
    \centering
    \includegraphics[width=1.0\linewidth]{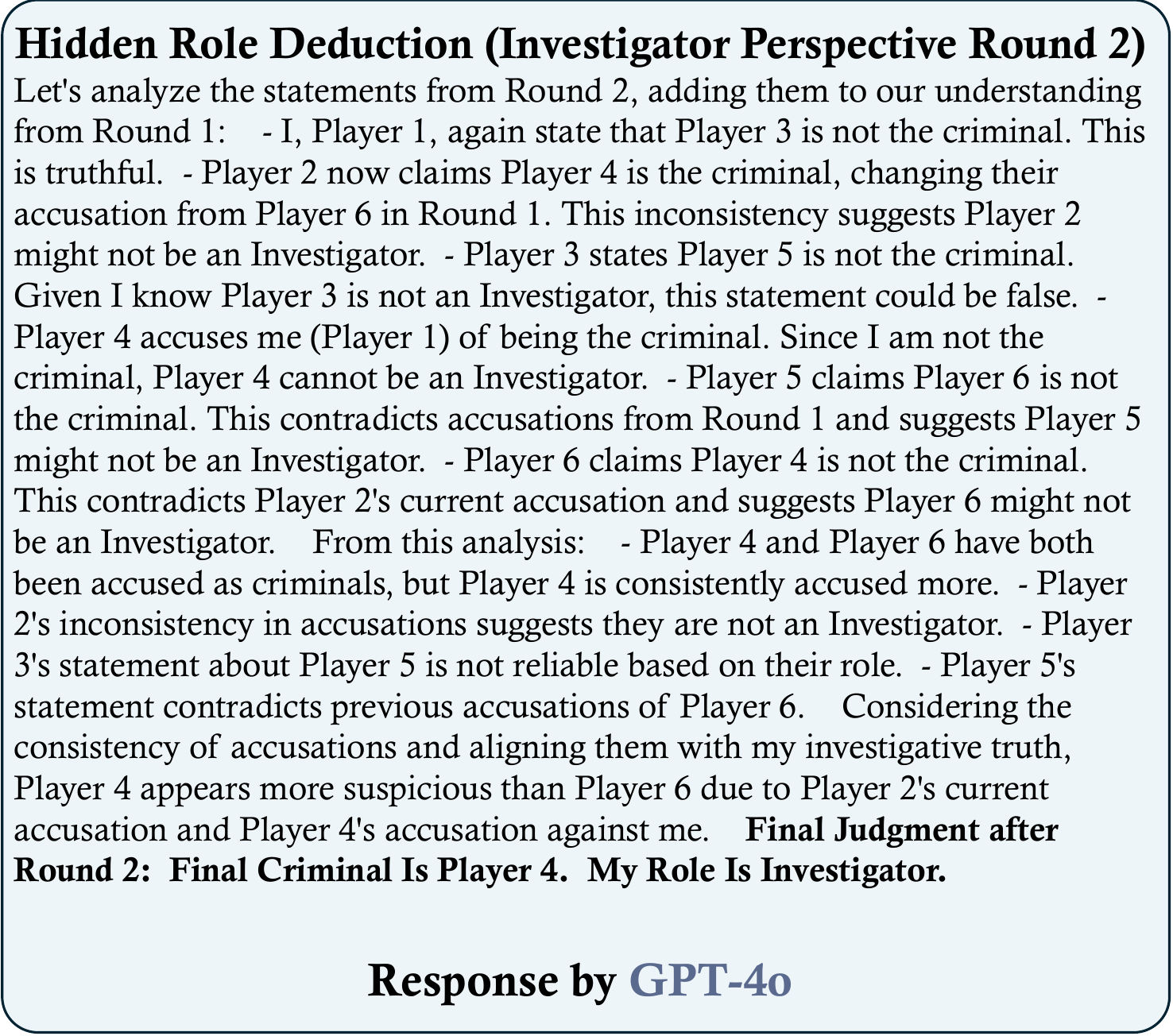}
    \caption{Case study: Hidden Role Deduction (Investigator Perspective) GPT-4o Round 2}
    \label{fig:case3}
\end{figure*}

\begin{figure*}[ht]
    \centering
    \includegraphics[width=1.0\linewidth]{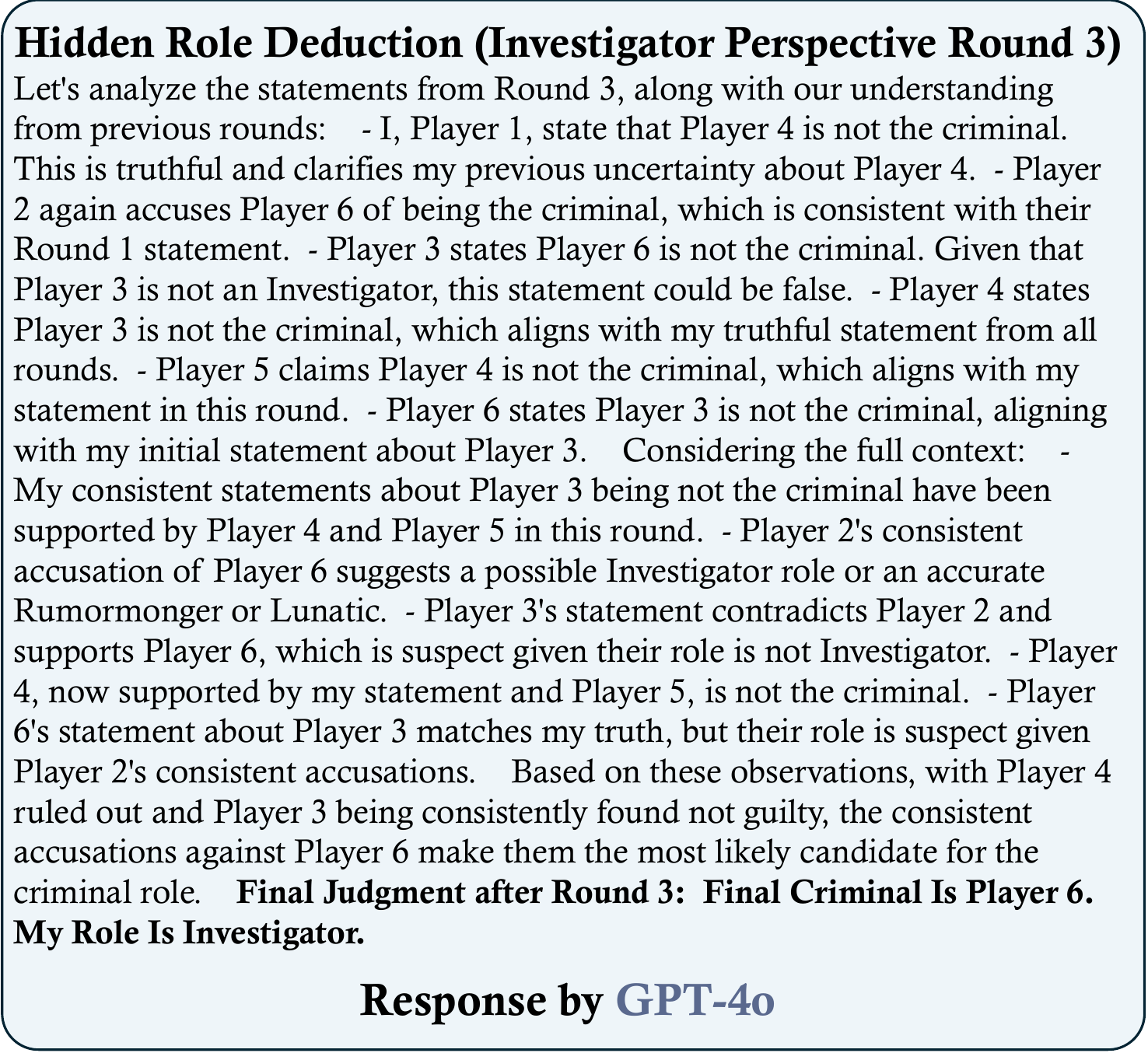}
    \caption{Case study: Hidden Role Deduction (Investigator Perspective) GPT-4o Round 3}
    \label{fig:case4}
\end{figure*}

\begin{figure*}[ht]
    \centering
    \includegraphics[width=1.0\linewidth]{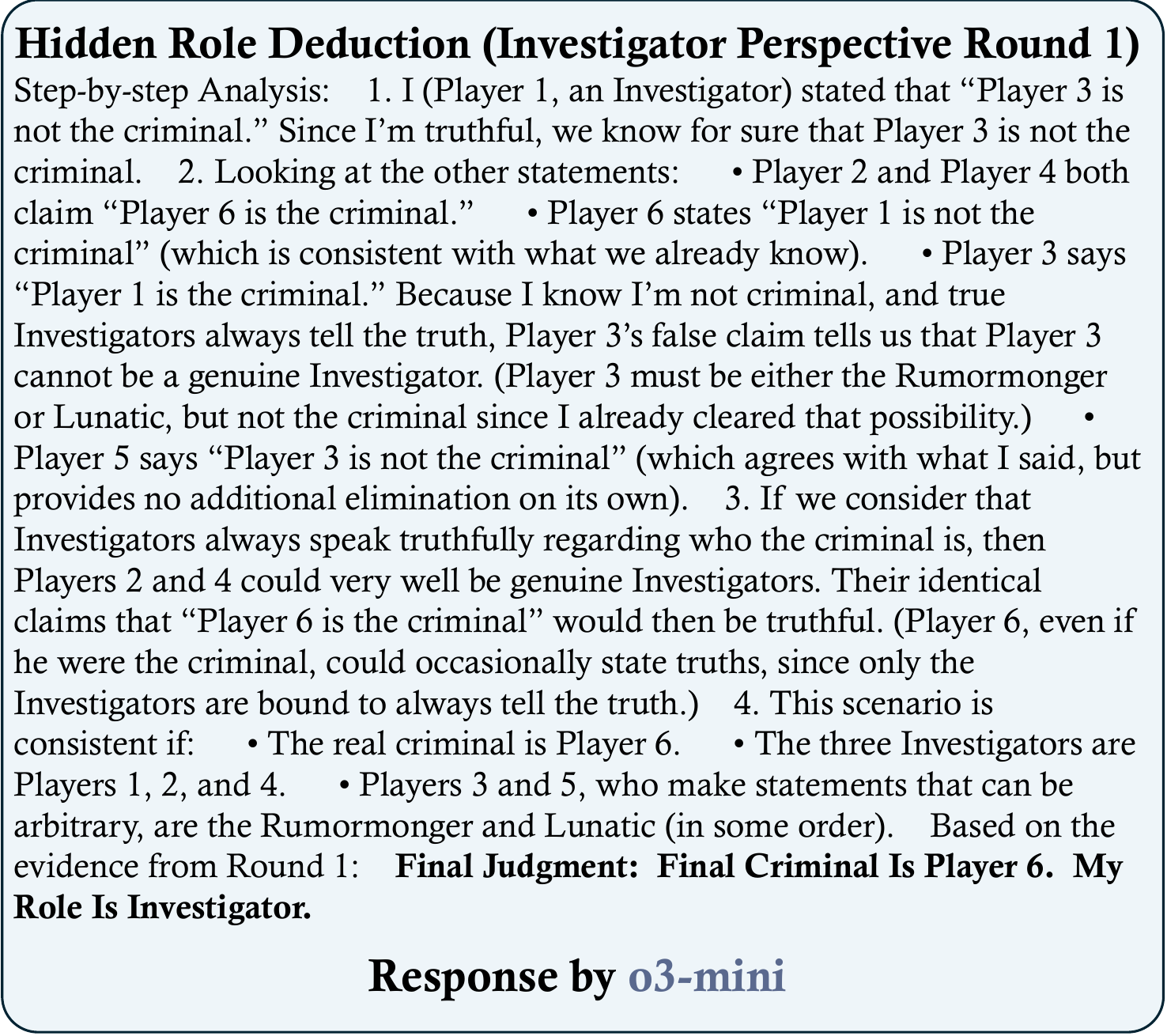}
    \caption{Case study: Hidden Role Deduction (Investigator Perspective) o3-mini Round 1}
    \label{fig:case5}
\end{figure*}

\begin{figure*}[ht]
    \centering
    \includegraphics[width=1.0\linewidth]{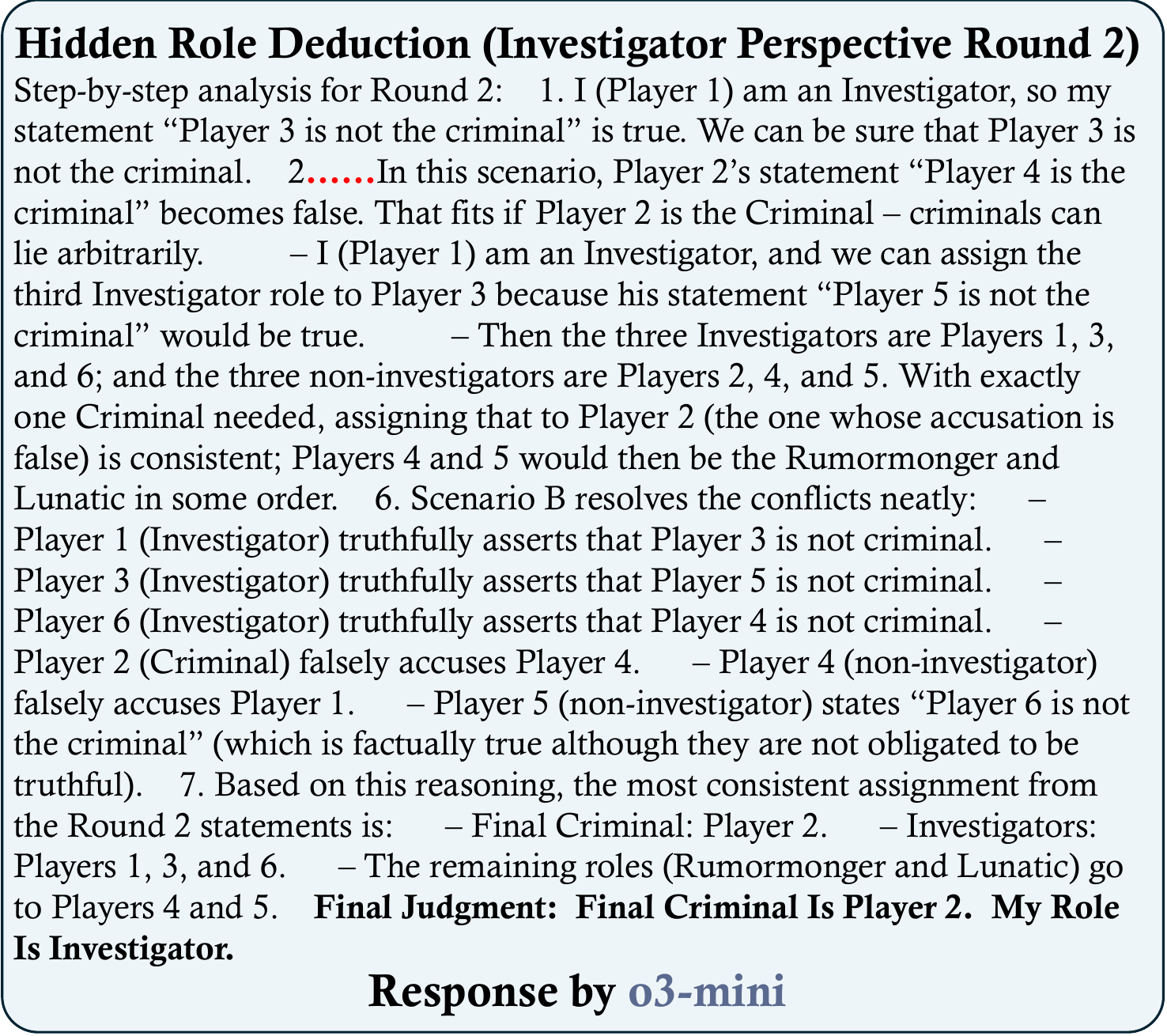}
    \caption{Case study: Hidden Role Deduction (Investigator Perspective) o3-mini Round 2}
    \label{fig:case6}
\end{figure*}

\begin{figure*}[ht]
    \centering
    \includegraphics[width=1.0\linewidth]{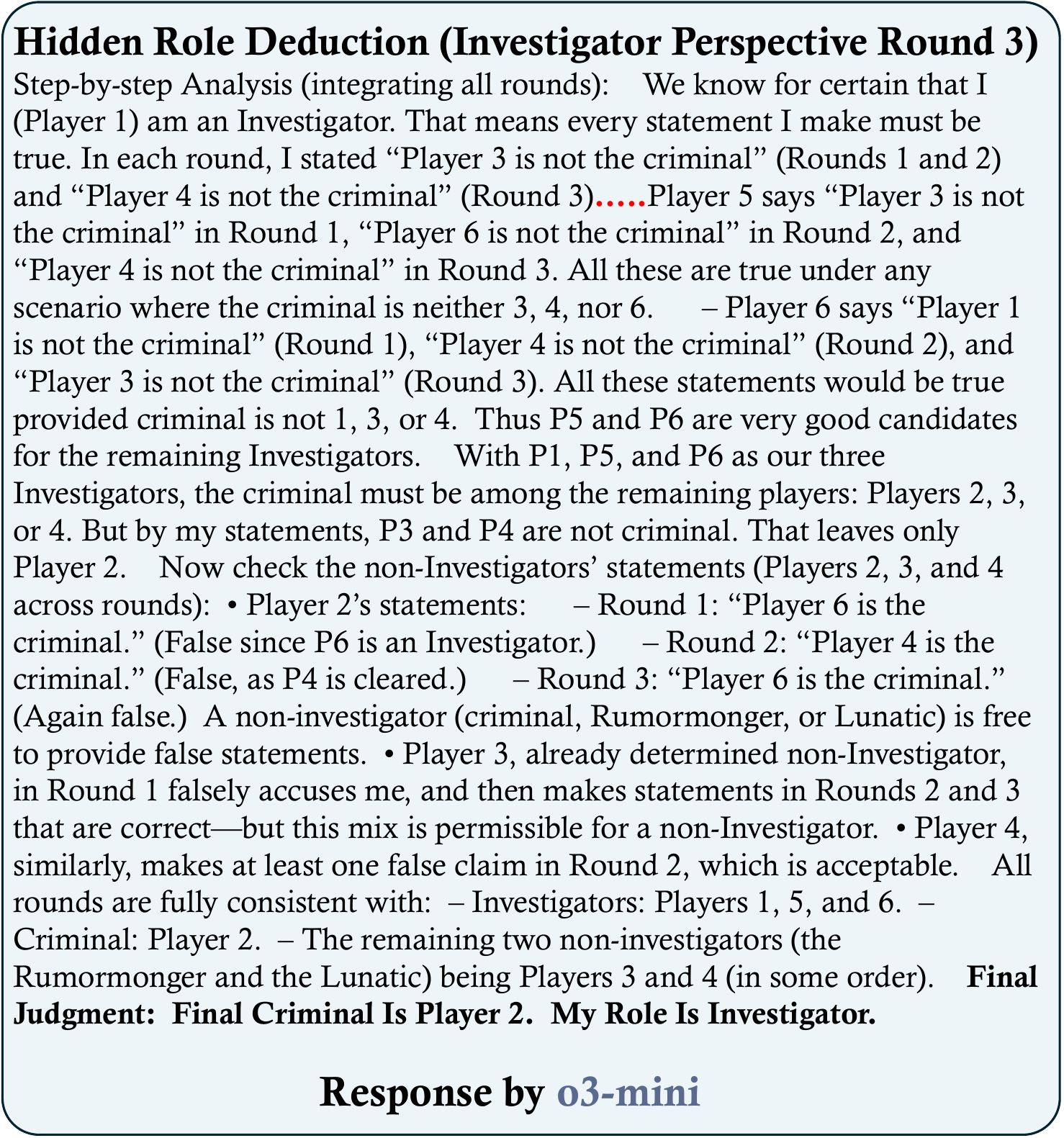}
    \caption{Case study: Hidden Role Deduction (Investigator Perspective) o3-mini Round 3}
    \label{fig:case7}
\end{figure*}

\begin{figure*}[ht]
    \centering
    \includegraphics[width=1.0\linewidth]{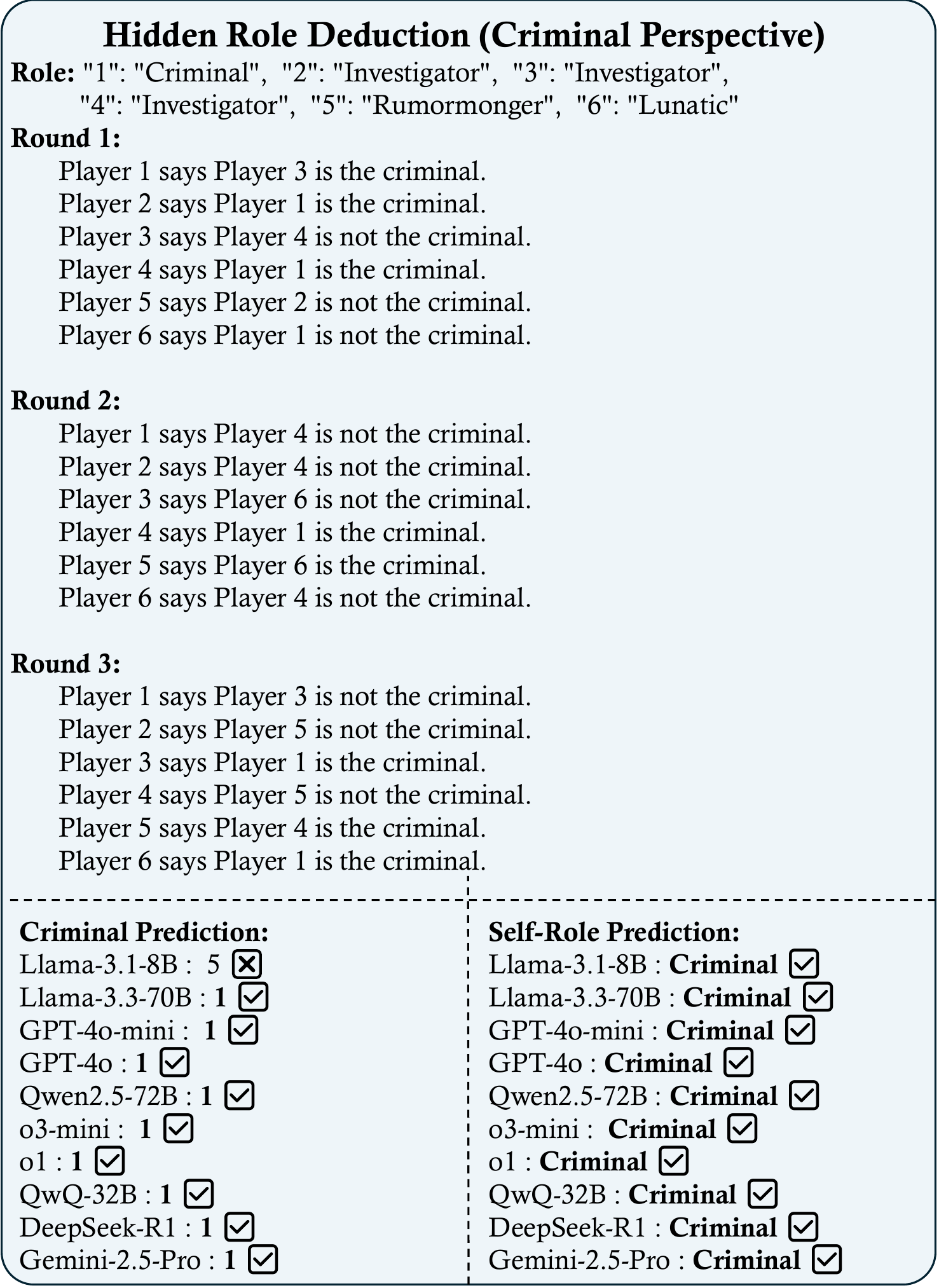}
    \caption{Case study: Hidden Role Deduction (Criminal Perspective)}
    \label{fig:case8}
\end{figure*}

\begin{figure*}[ht]
    \centering
    \includegraphics[width=1.0\linewidth]{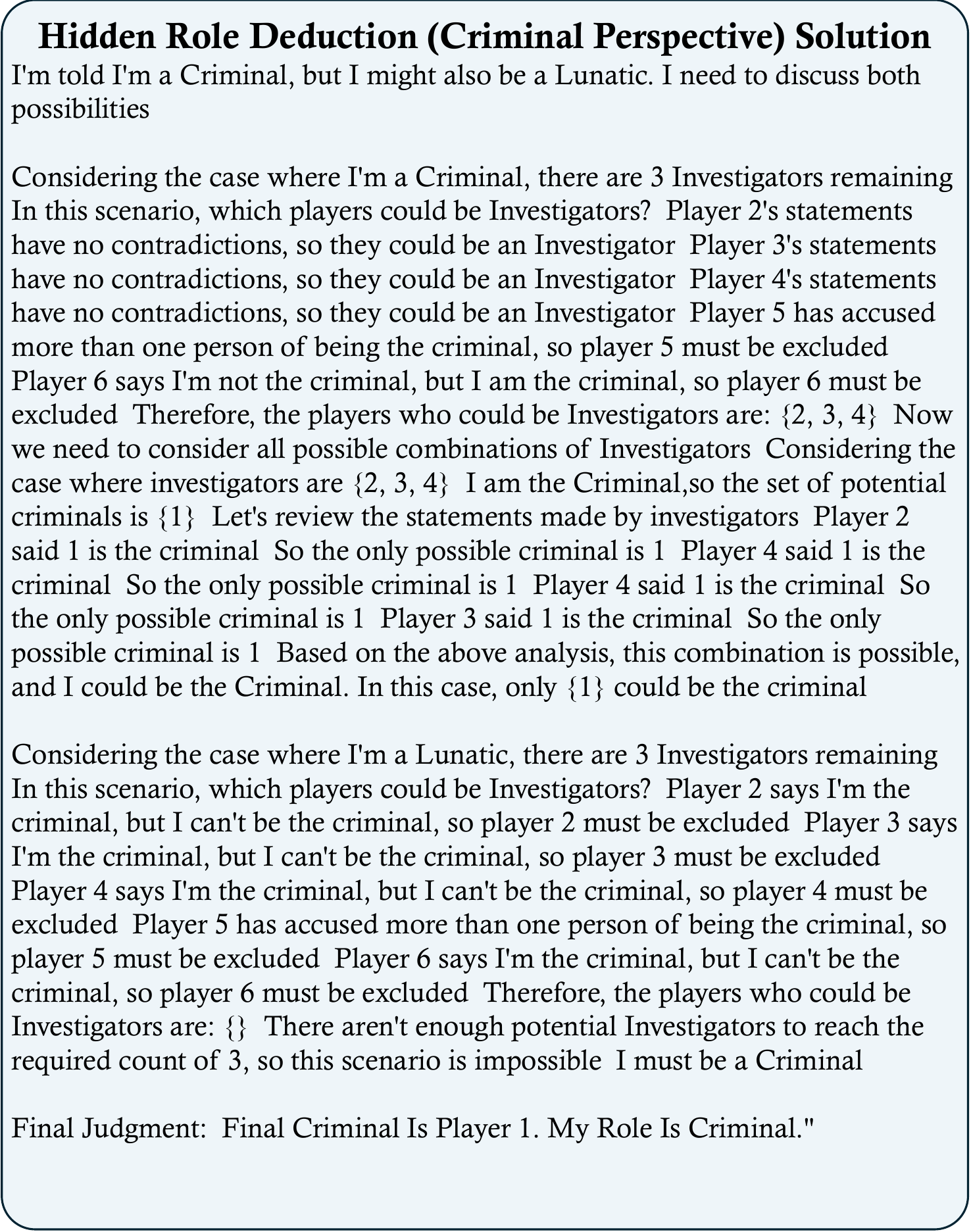}
    \caption{Case study: Hidden Role Deduction (Criminal Perspective) Solution }
    \label{fig:case42}
\end{figure*}

\begin{figure*}[ht]
    \centering
    \includegraphics[width=1.0\linewidth]{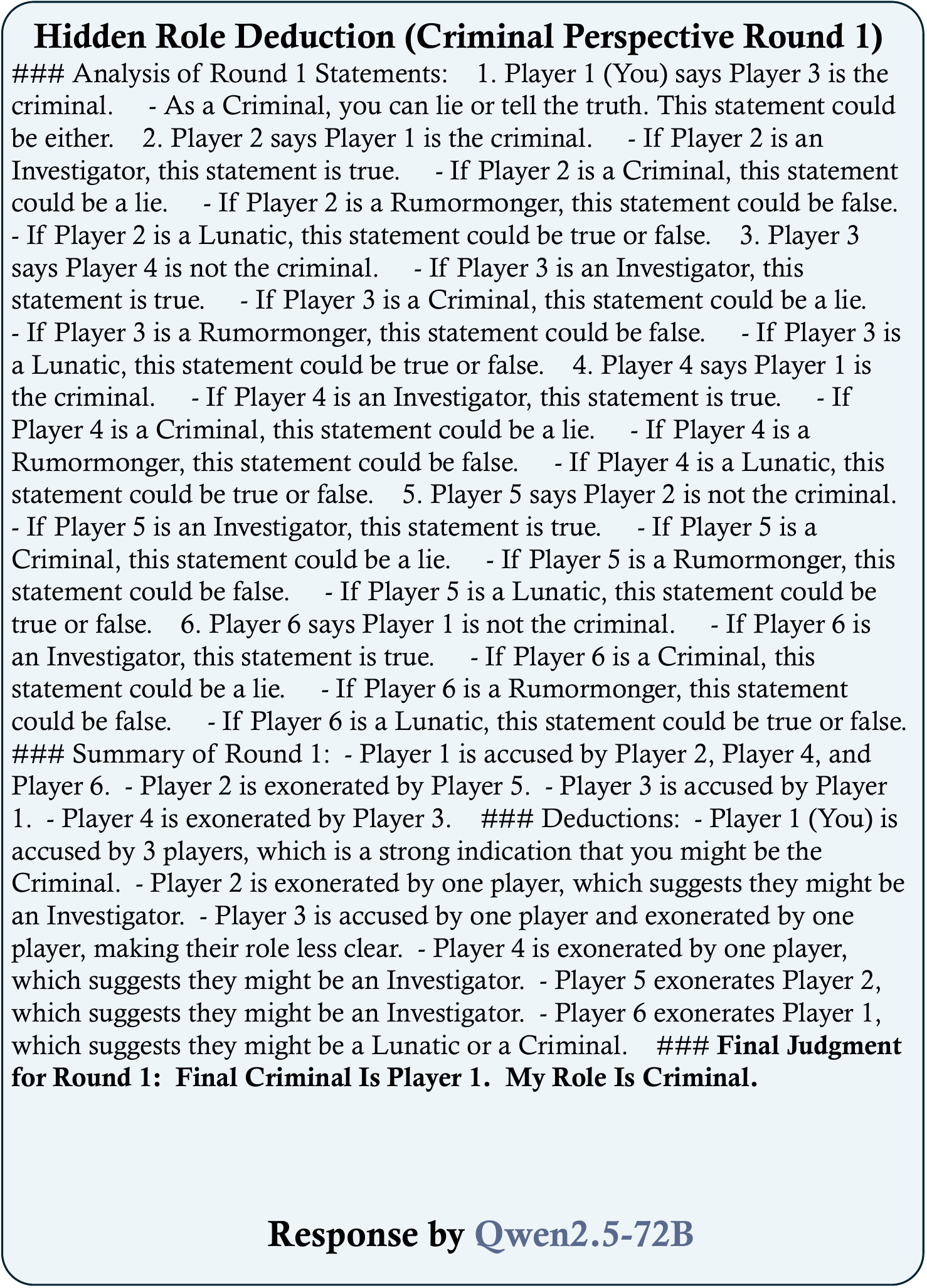}
    \caption{Case study: Hidden Role Deduction (Criminal Perspective) Qwen2.5-72B Round 1}
    \label{fig:case9}
\end{figure*}

\begin{figure*}[ht]
    \centering
    \includegraphics[width=0.98\linewidth]{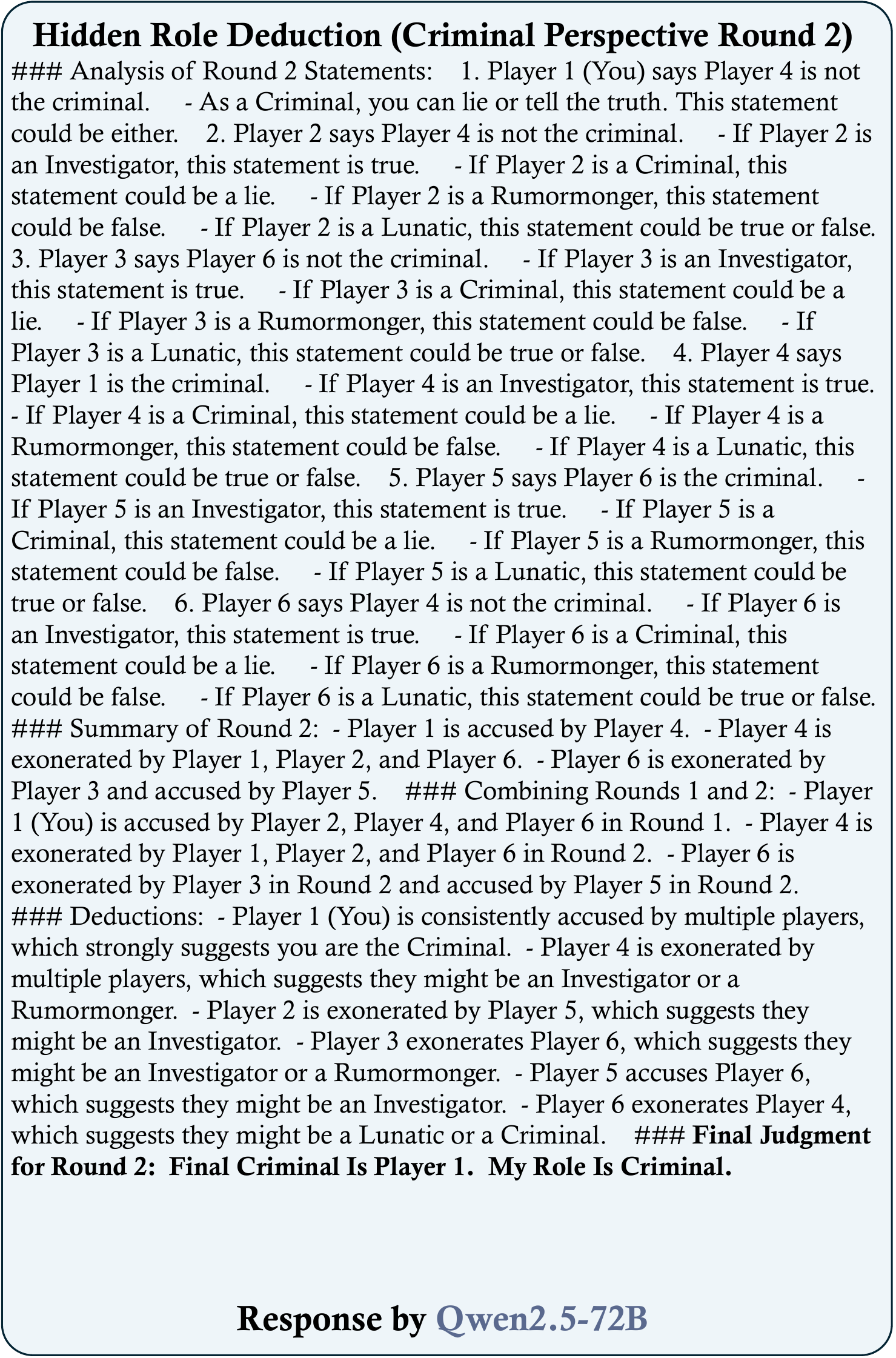}
    \caption{Case study: Hidden Role Deduction (Criminal Perspective) Qwen2.5-72B Round 2}
    \label{fig:case10}
\end{figure*}

\begin{figure*}[ht]
    \centering
    \includegraphics[width=1.0\linewidth]{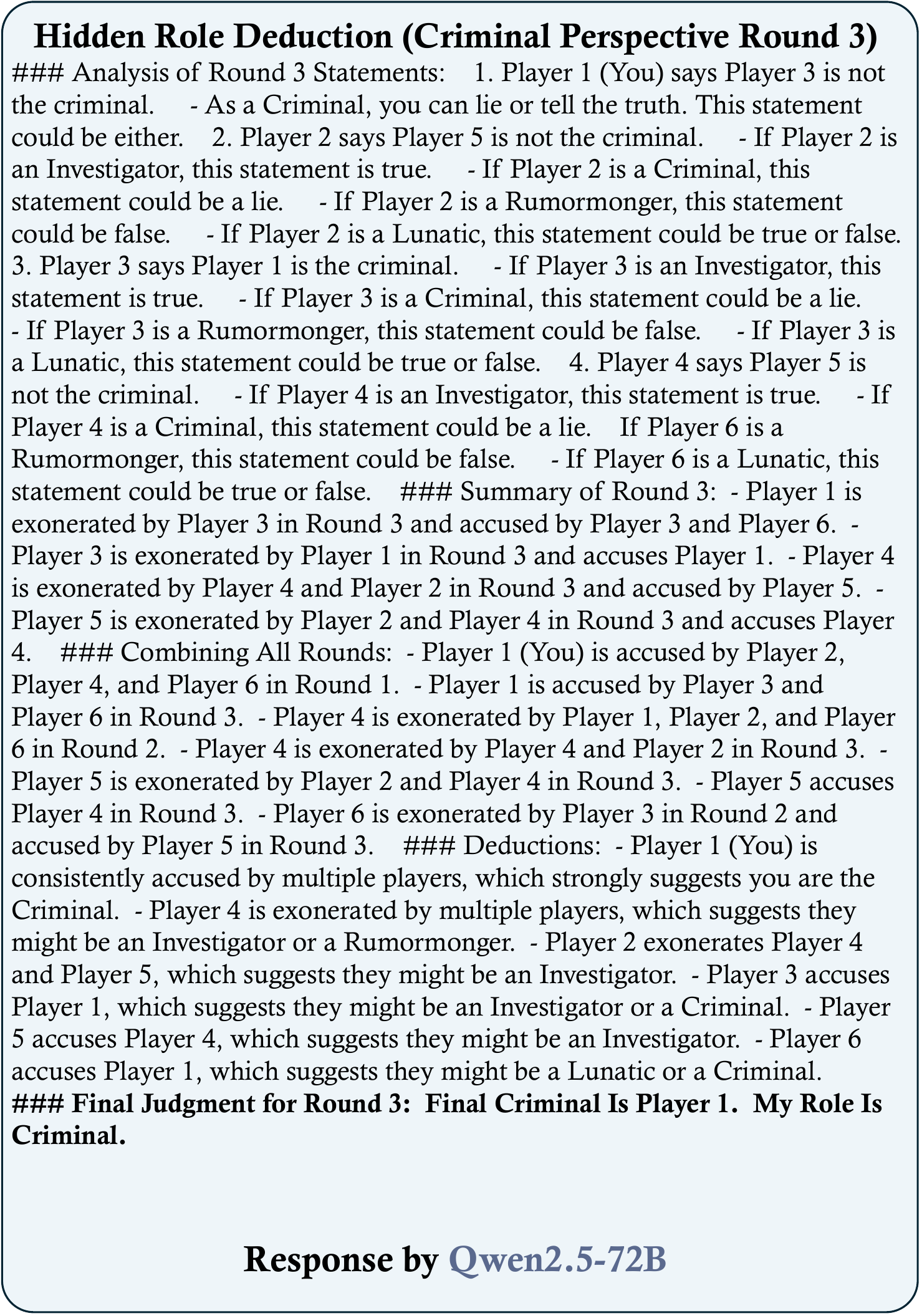}
    \caption{Case study: Hidden Role Deduction (Criminal Perspective) Qwen2.5-72B Round 3}
    \label{fig:case11}
\end{figure*}

\begin{figure*}[ht]
    \centering
    \includegraphics[width=1.0\linewidth]{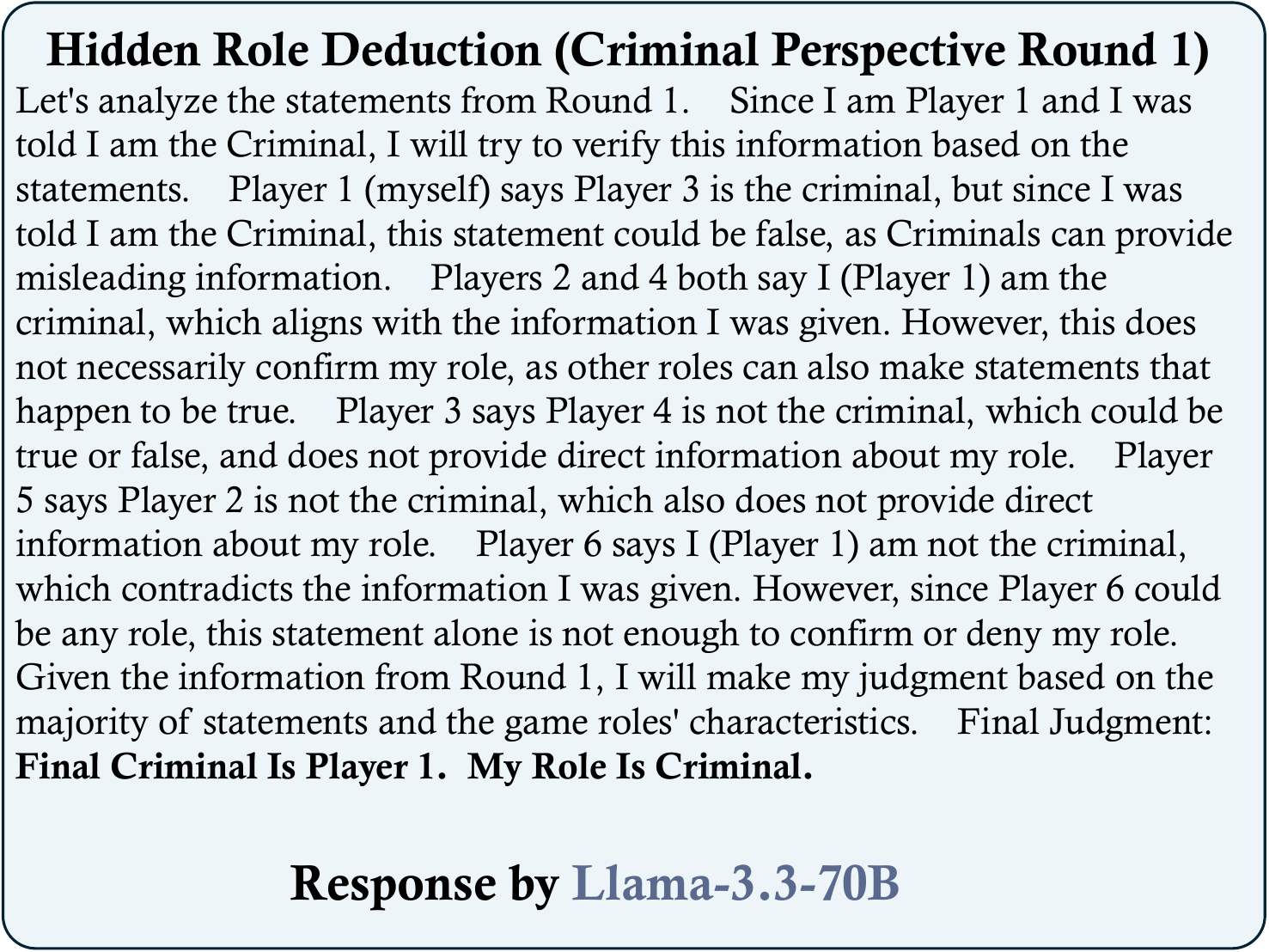}
    \caption{Case study: Hidden Role Deduction (Criminal Perspective) Llama-3.3-70B Round 1}
    \label{fig:case12}
\end{figure*}

\begin{figure*}[ht]
    \centering
    \includegraphics[width=1.0\linewidth]{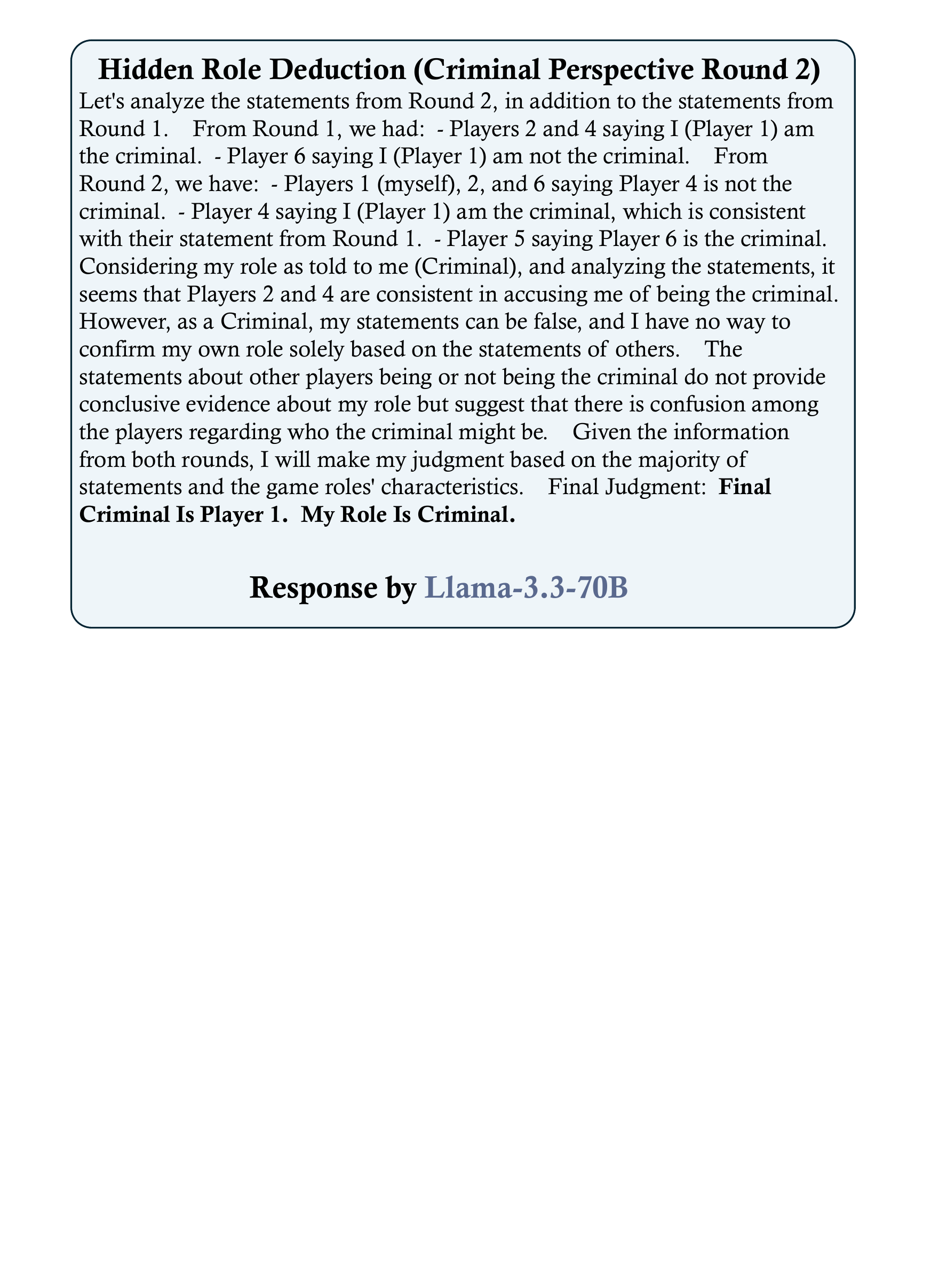}
    \caption{Case study: Hidden Role Deduction (Criminal Perspective) Llama-3.3-70B Round 2}
    \label{fig:case13}
\end{figure*}

\begin{figure*}[ht]
    \centering
    \includegraphics[width=1.0\linewidth]{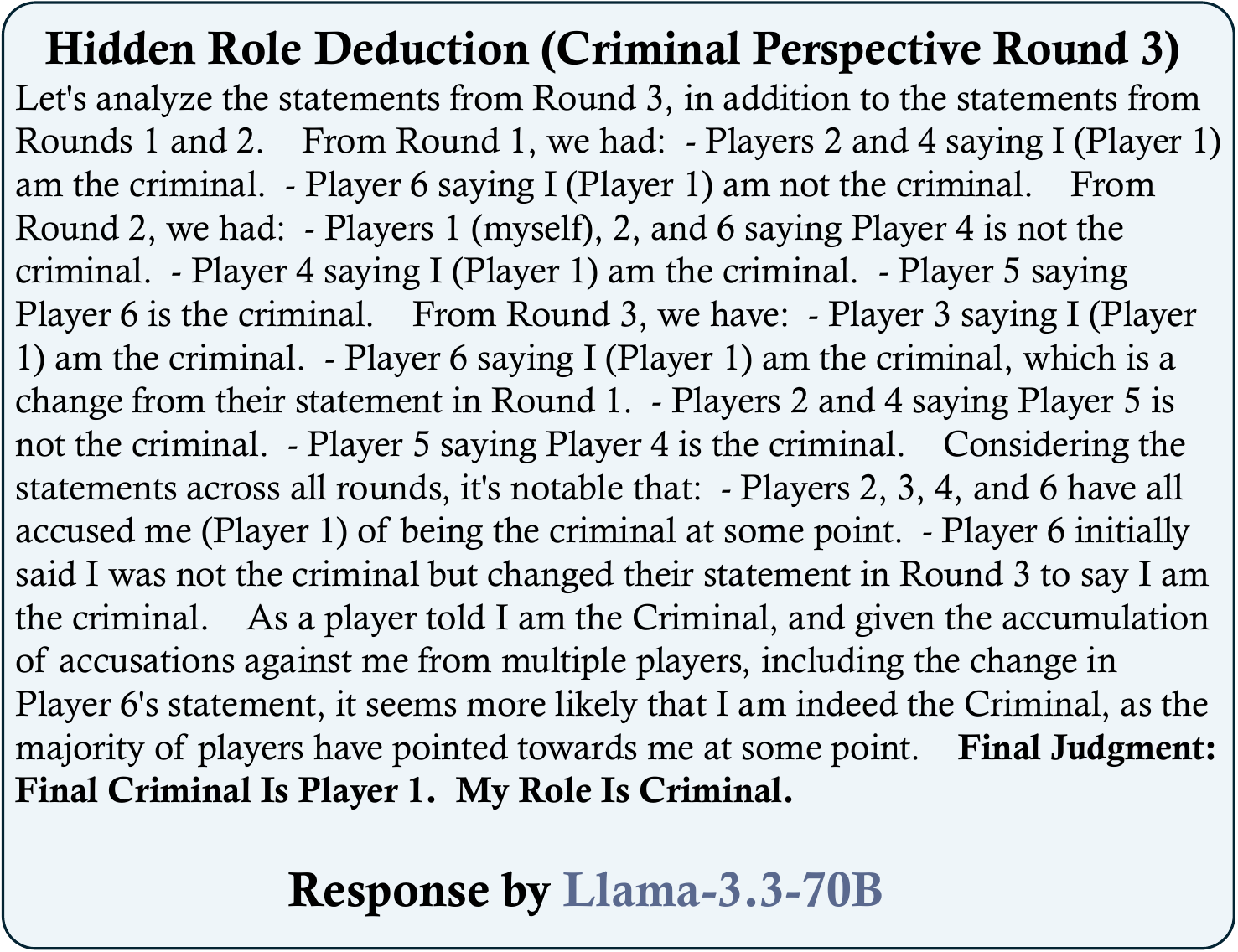}
    \caption{Case study: Hidden Role Deduction (Criminal Perspective) Llama-3.3-70B Round 3}
    \label{fig:case14}
\end{figure*}

\begin{figure*}[ht]
    \centering
    \includegraphics[width=1.0\linewidth]{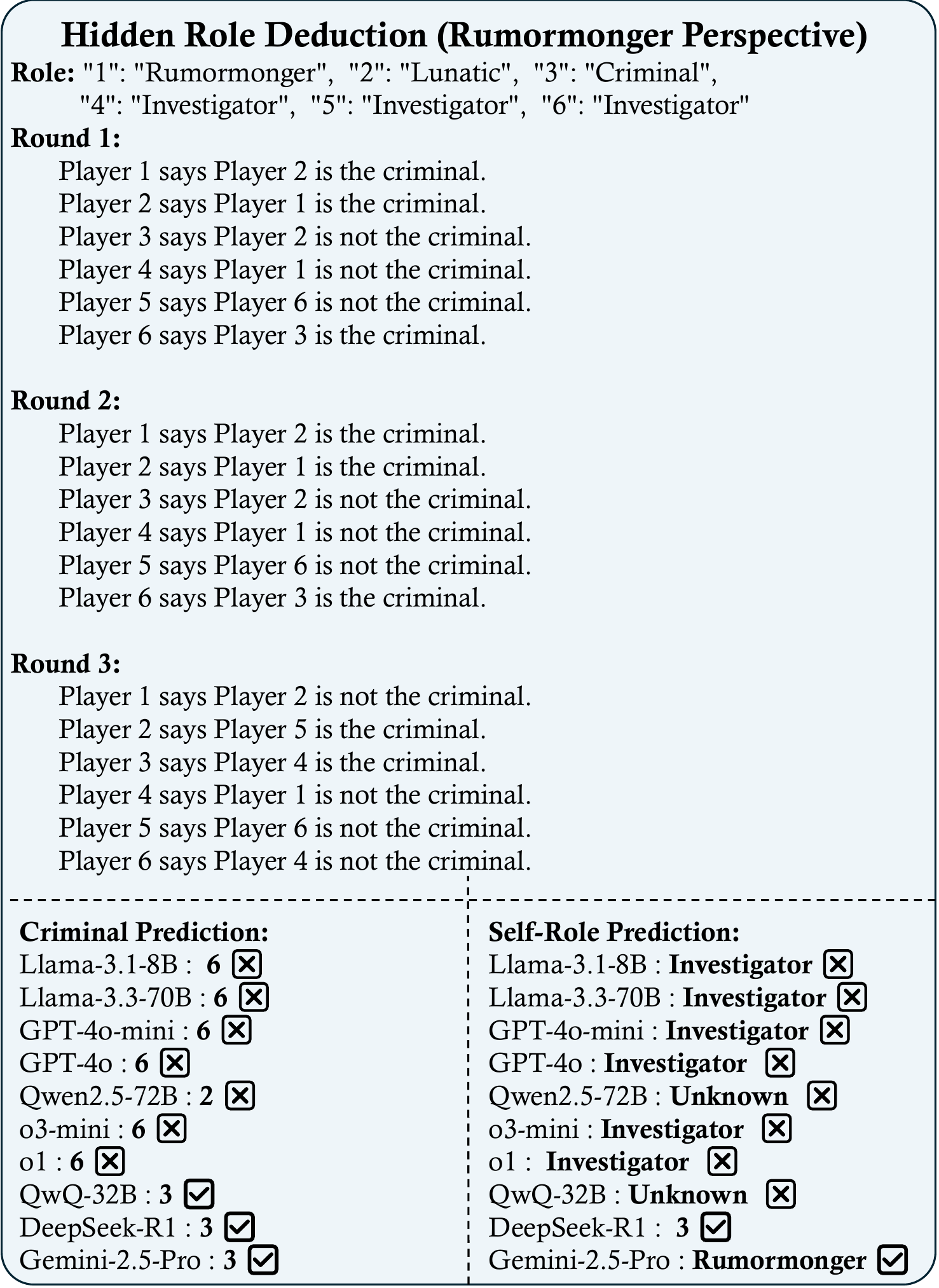}
    \caption{Case study: Hidden Role Deduction (Rumormonger Perspective)}
    \label{fig:case15}
\end{figure*}

\begin{figure*}[ht]
    \centering
    \includegraphics[width=1.0\linewidth]{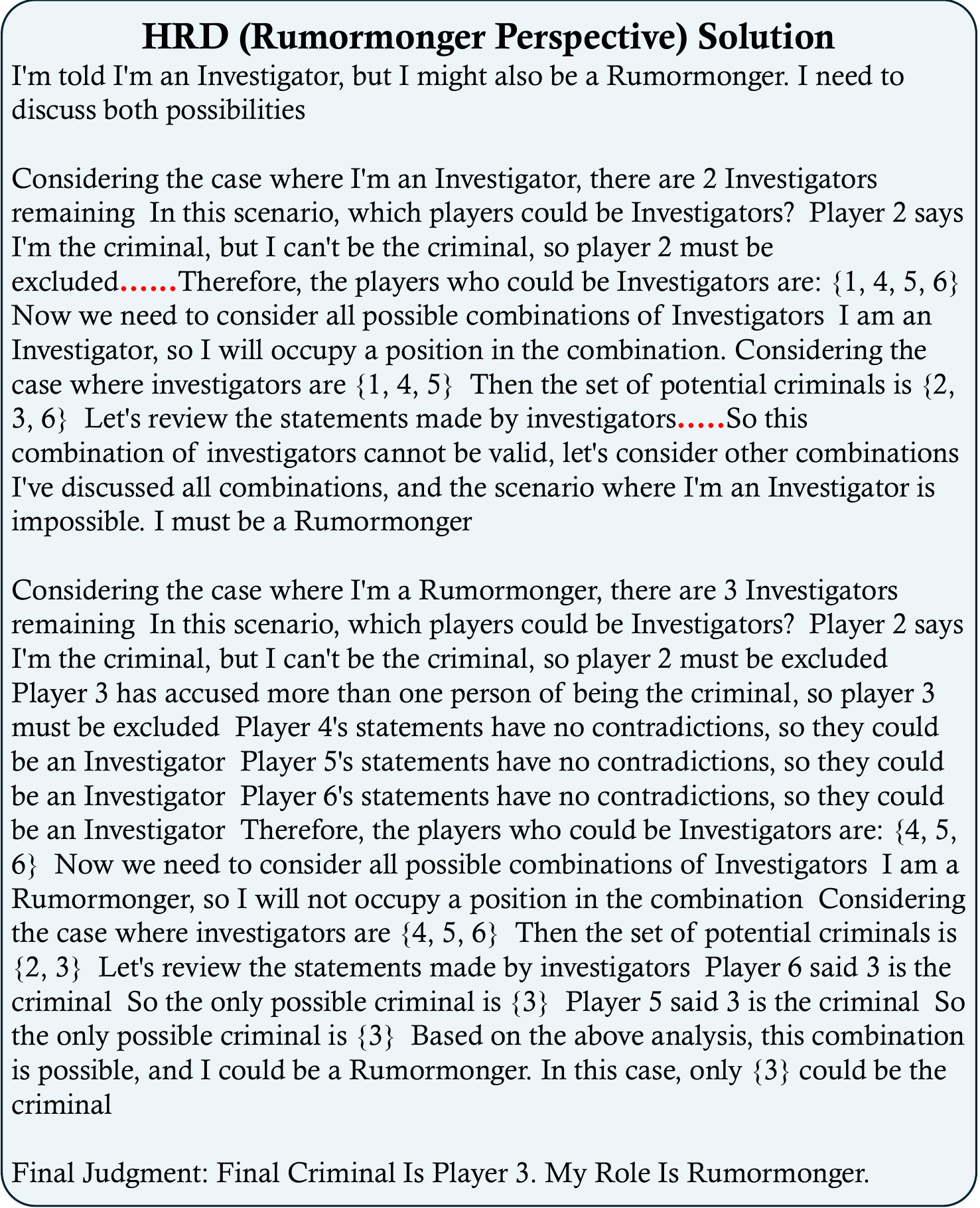}
    \caption{Case study: Hidden Role Deduction (Rumormonger Perspective) Solution}
    \label{fig:case43}
\end{figure*}

\begin{figure*}[ht]
    \centering
    \includegraphics[width=1.0\linewidth]{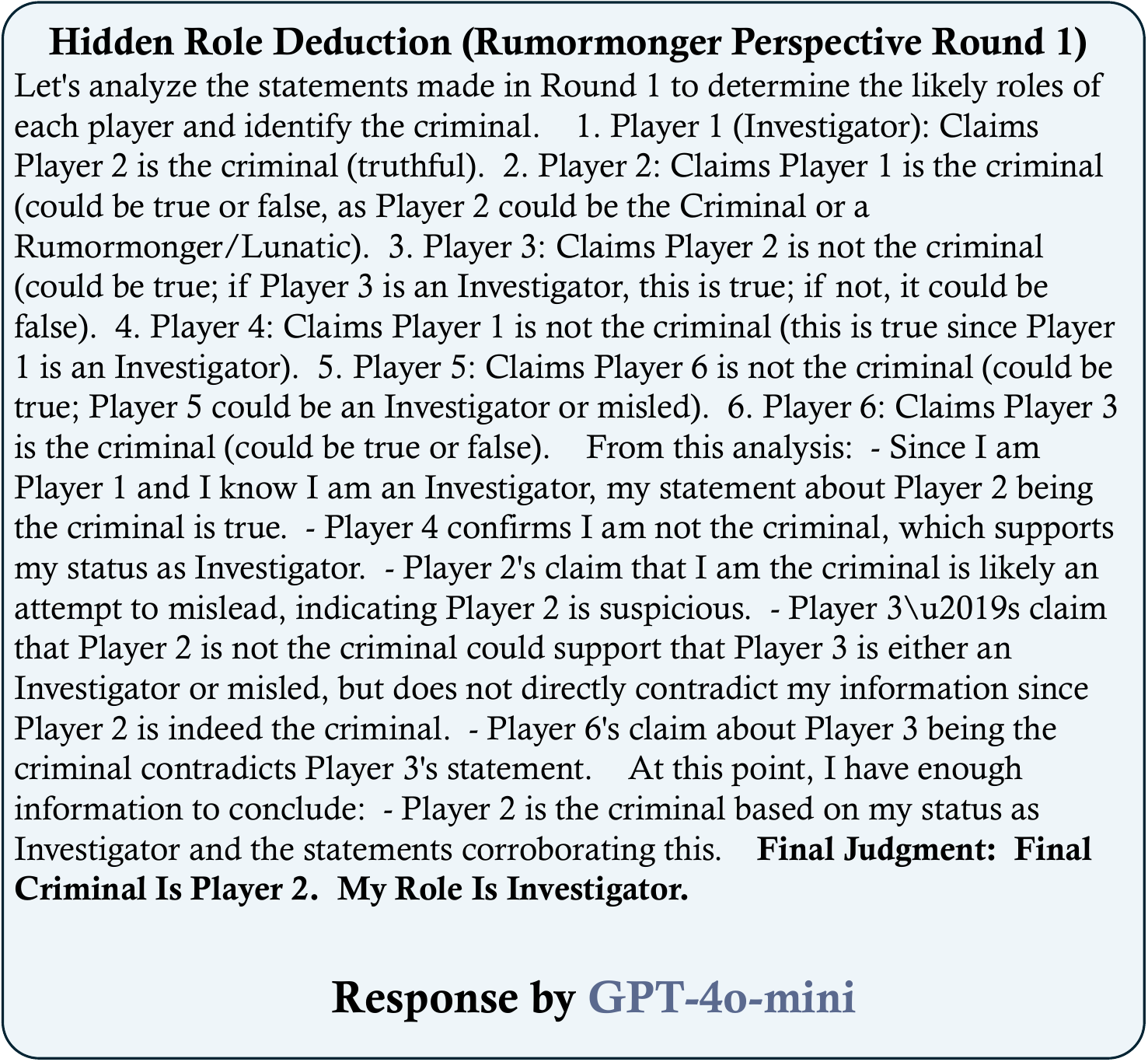}
    \caption{Case study: Hidden Role Deduction (Rumormonger Perspective) GPT-4o-mini Round 1}
    \label{fig:case16}
\end{figure*}

\begin{figure*}[ht]
    \centering
    \includegraphics[width=1.0\linewidth]{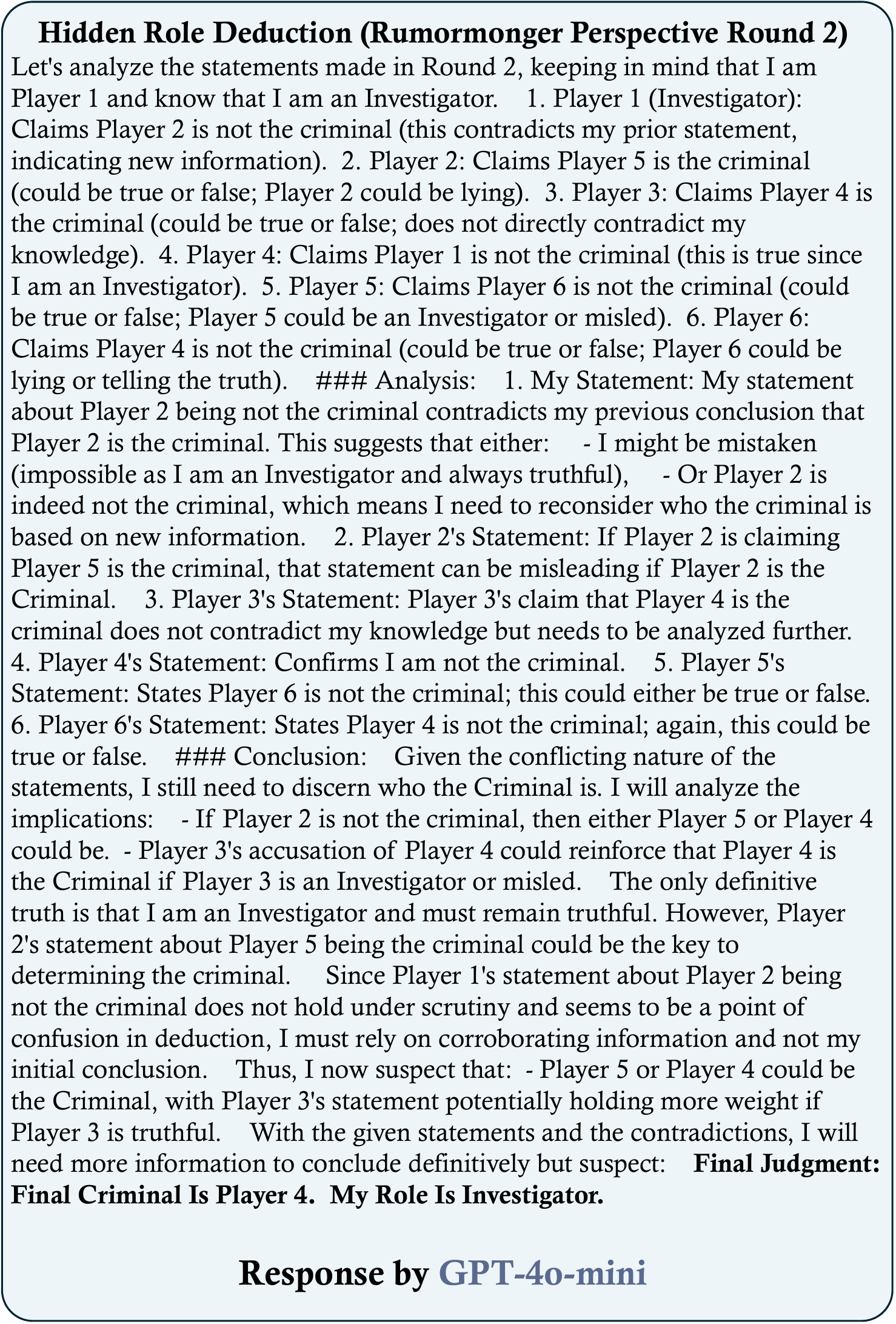}
    \caption{Case study: Hidden Role Deduction (Rumormonger Perspective) GPT-4o-mini Round 2}
    \label{fig:case17}
\end{figure*}

\begin{figure*}[ht]
    \centering
    \includegraphics[width=1.0\linewidth]{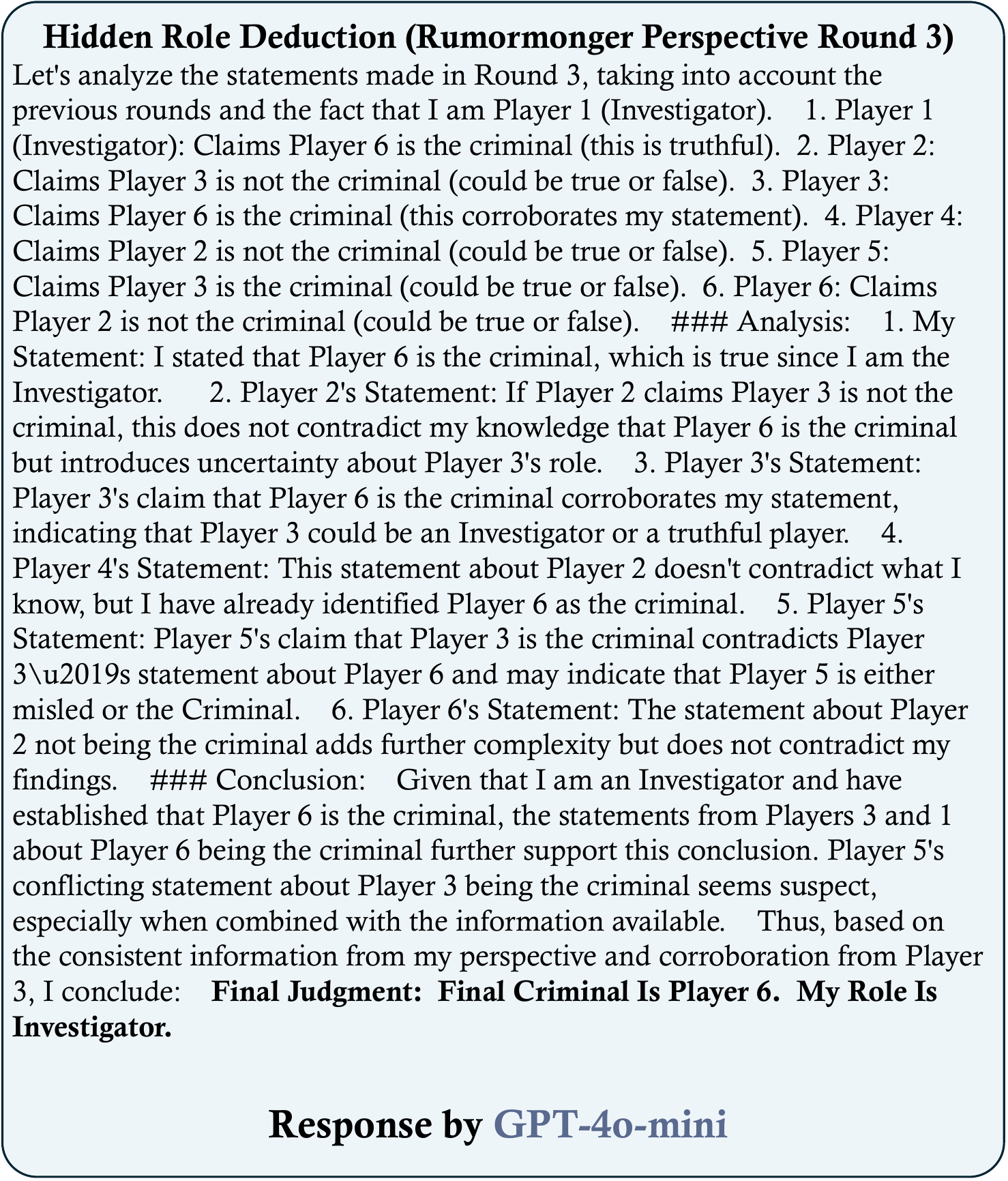}
    \caption{Case study: Hidden Role Deduction (Rumormonger Perspective) GPT-4o-mini Round 3}
    \label{fig:case18}
\end{figure*}

\begin{figure*}[ht]
    \centering
    \includegraphics[width=1.0\linewidth]{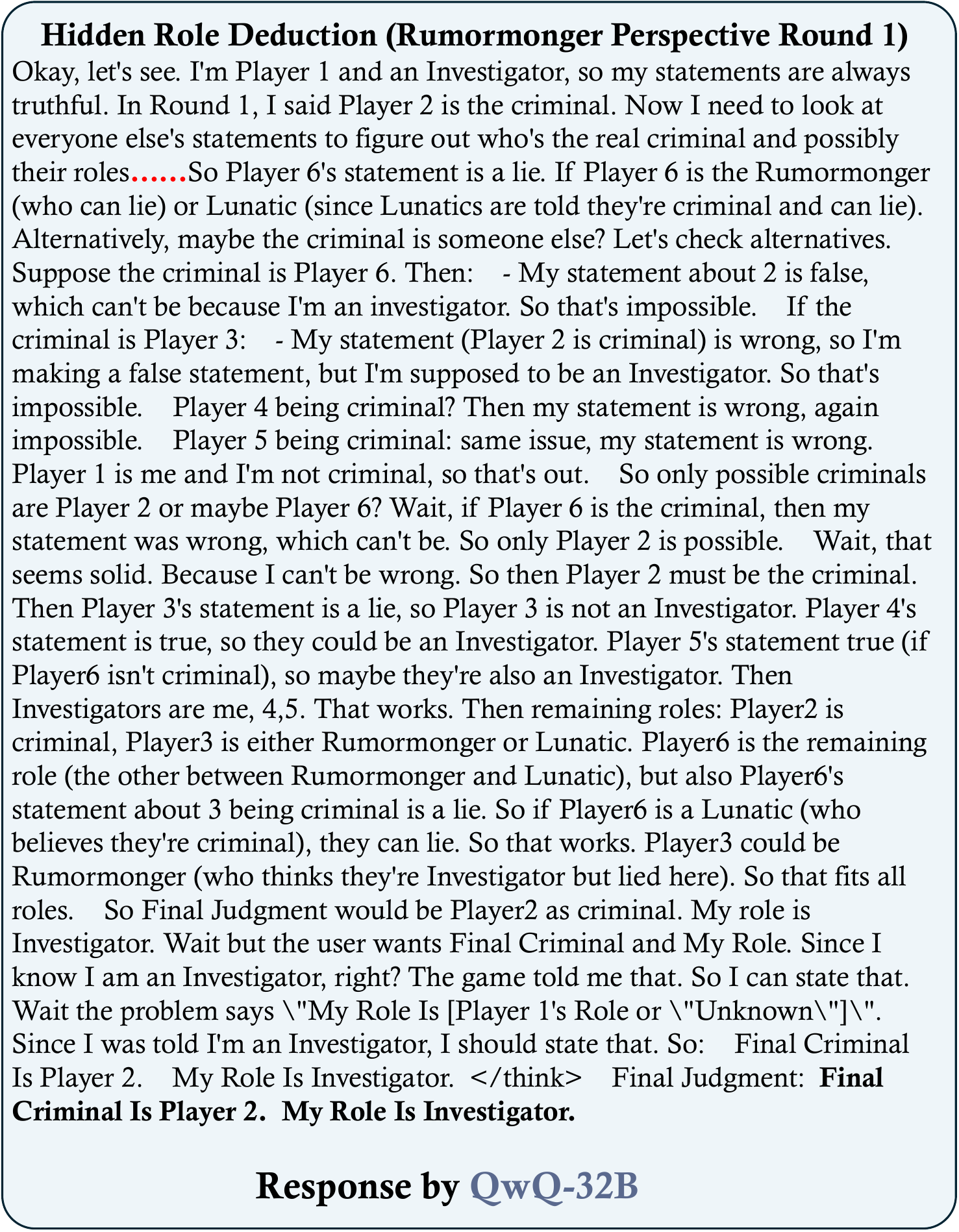}
    \caption{Case study: Hidden Role Deduction (Rumormonger Perspective) QwQ-32B Round 1}
    \label{fig:case19}
\end{figure*}

\begin{figure*}[ht]
    \centering
    \includegraphics[width=1.0\linewidth]{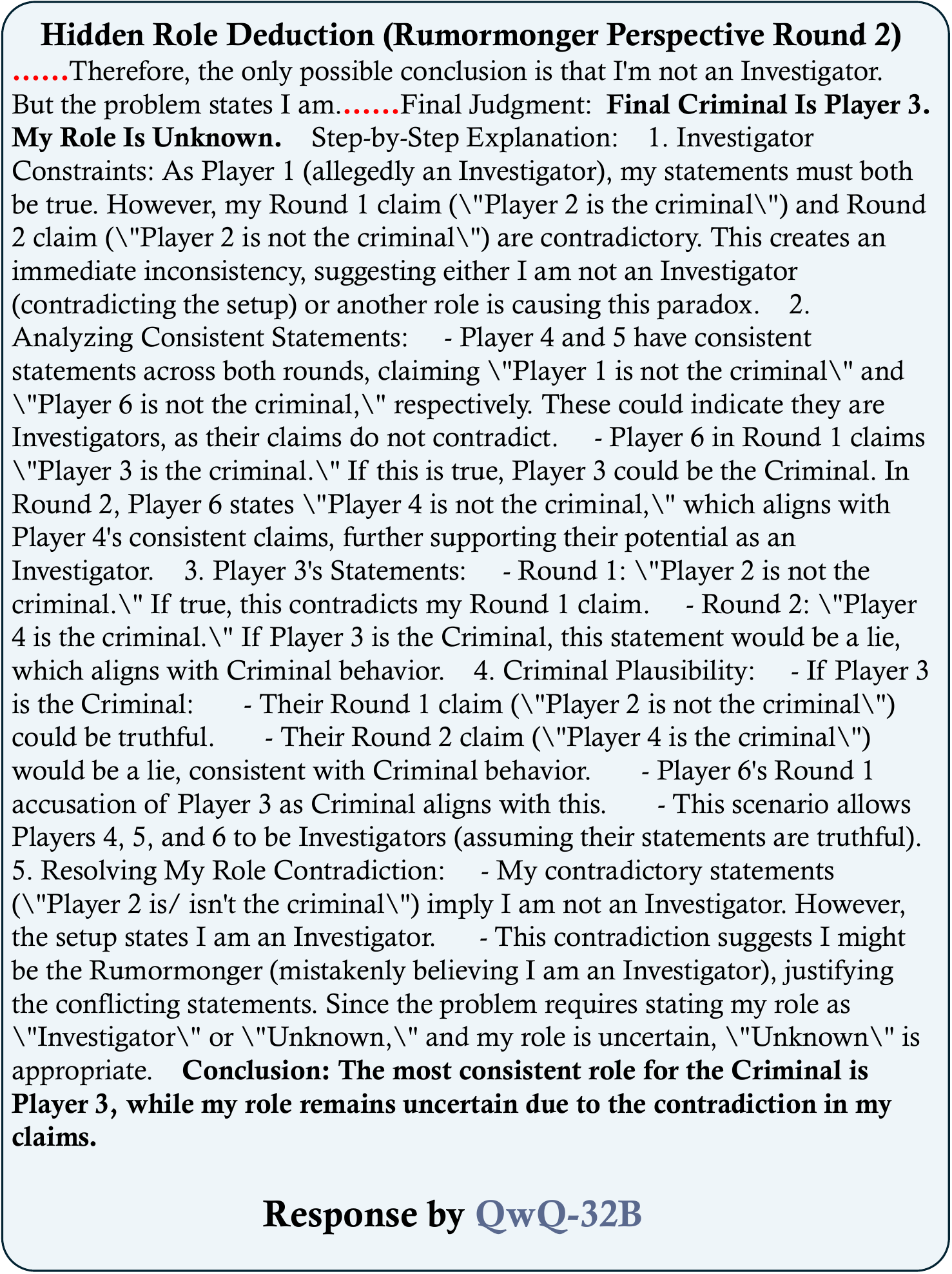}
    \caption{Case study: Hidden Role Deduction (Rumormonger Perspective) QwQ-32B Round 2}
    \label{fig:case20}
\end{figure*}

\begin{figure*}[ht]
    \centering
    \includegraphics[width=1.0\linewidth]{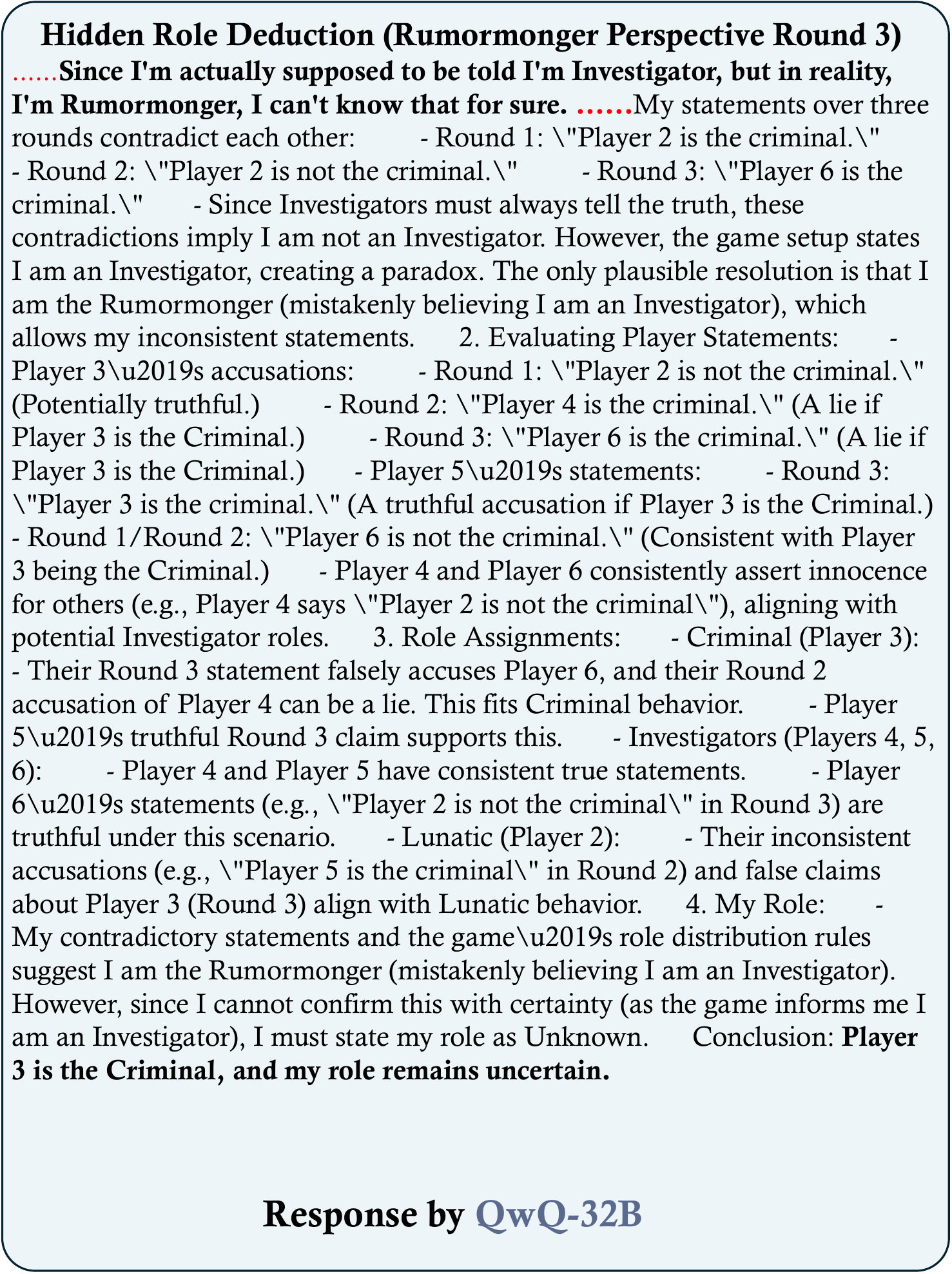}
    \caption{Case study: Hidden Role Deduction (Rumormonger Perspective) QwQ-32B Round 3}
    \label{fig:case21}
\end{figure*}

\begin{figure*}[ht]
    \centering
    \includegraphics[width=1.0\linewidth]{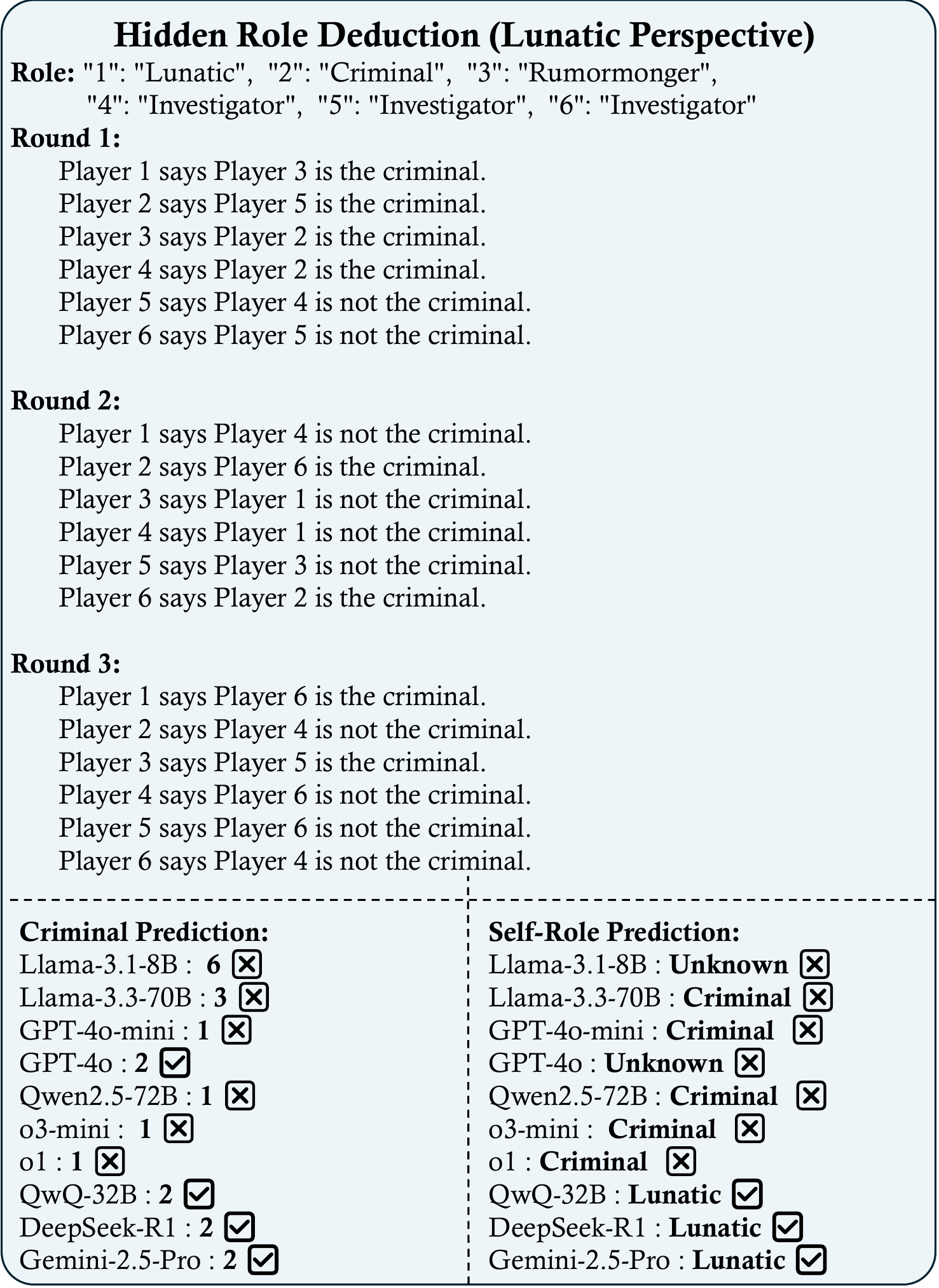}
    \caption{Case study: Hidden Role Deduction (Lunatic Perspective)}
    \label{fig:case22}
\end{figure*}

\begin{figure*}[ht]
    \centering
    \includegraphics[width=1.0\linewidth]{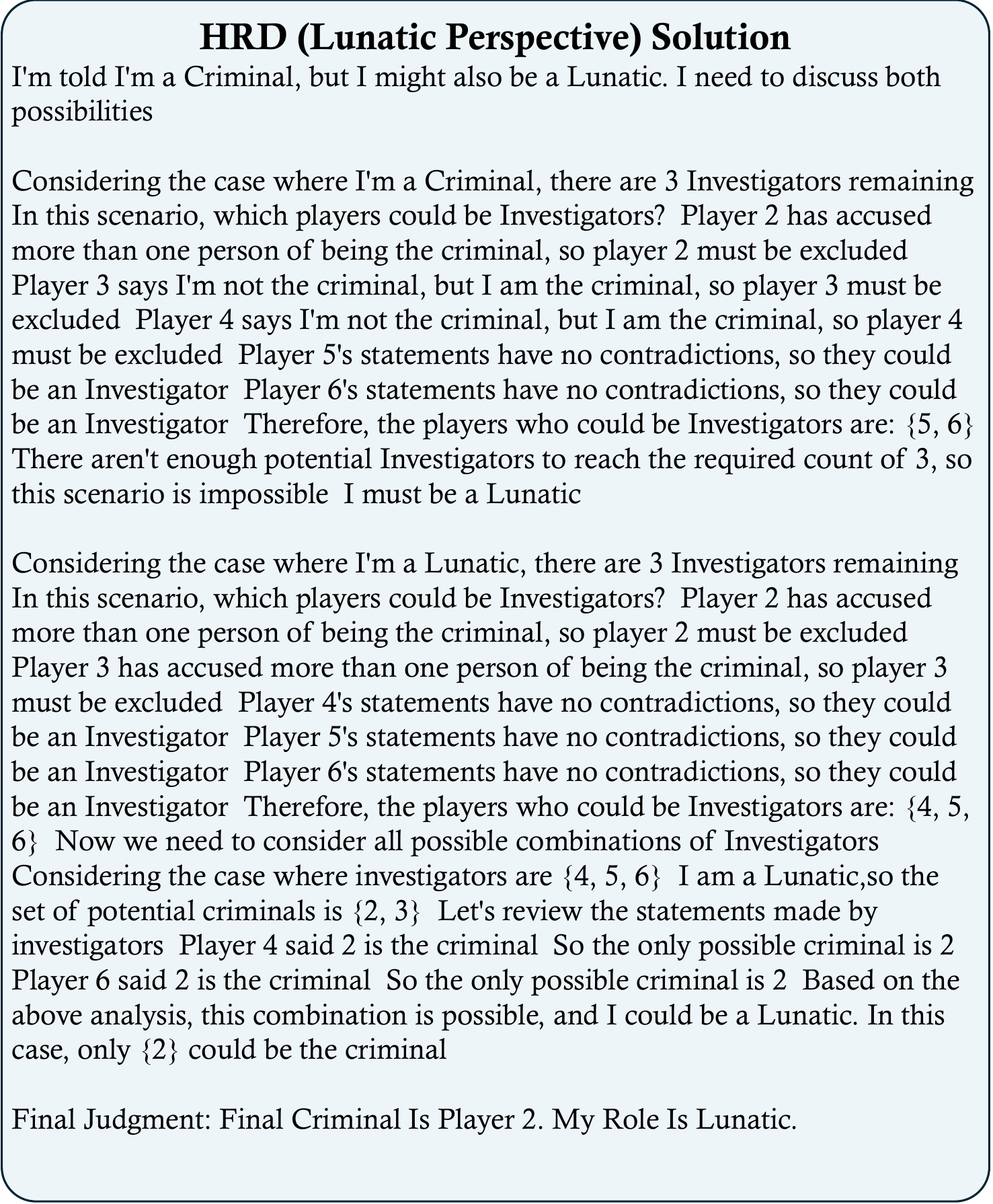}
    \caption{Case study: Hidden Role Deduction (Lunatic Perspective) Solution}
    \label{fig:case44}
\end{figure*}

\begin{figure*}[ht]
    \centering
    \includegraphics[width=1.0\linewidth]{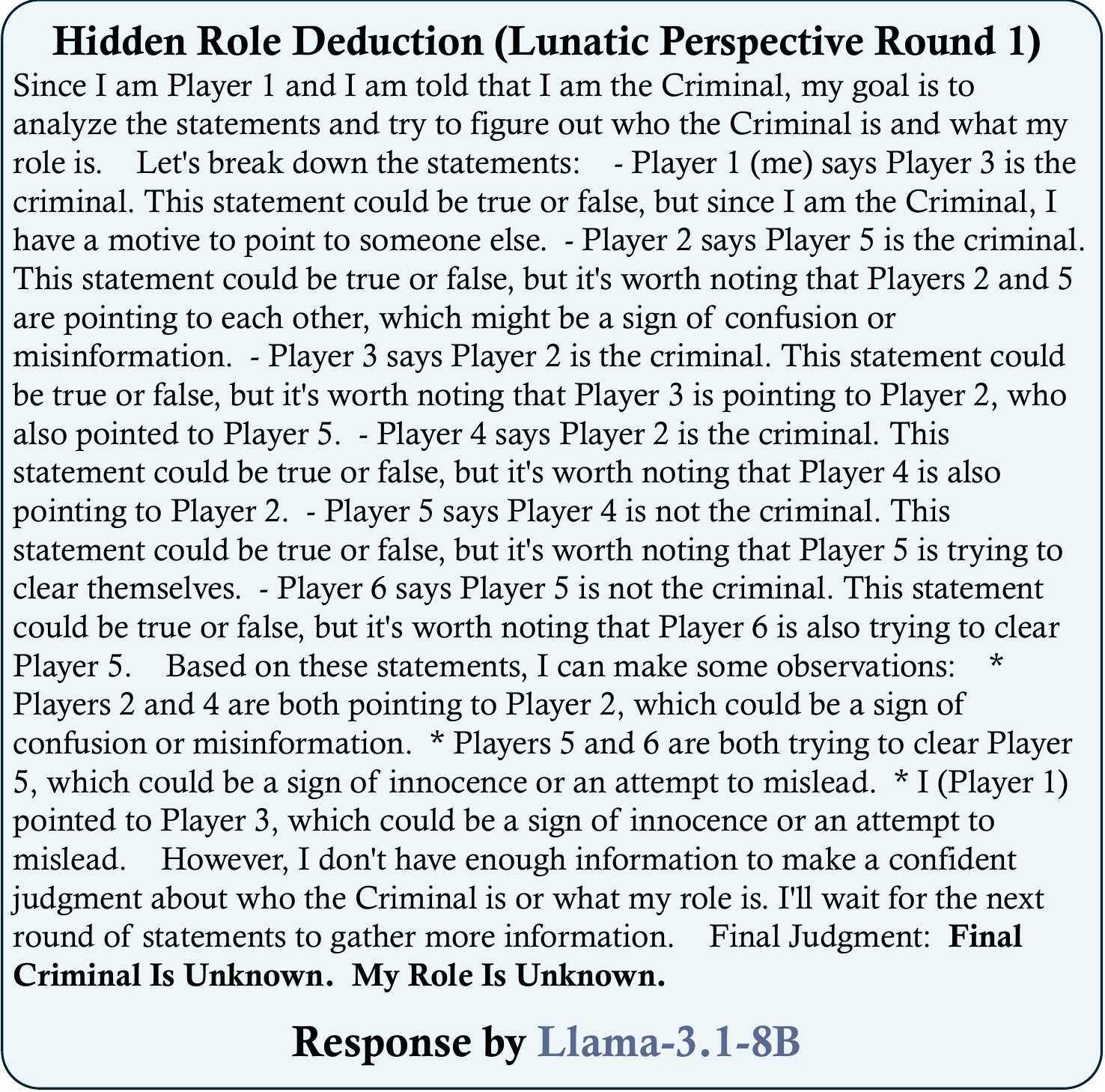}
    \caption{Case study: Hidden Role Deduction (Rumormonger Perspective) Llama-3.1-8B Round 1}
    \label{fig:case23}
\end{figure*}

\begin{figure*}[ht]
    \centering
    \includegraphics[width=1.0\linewidth]{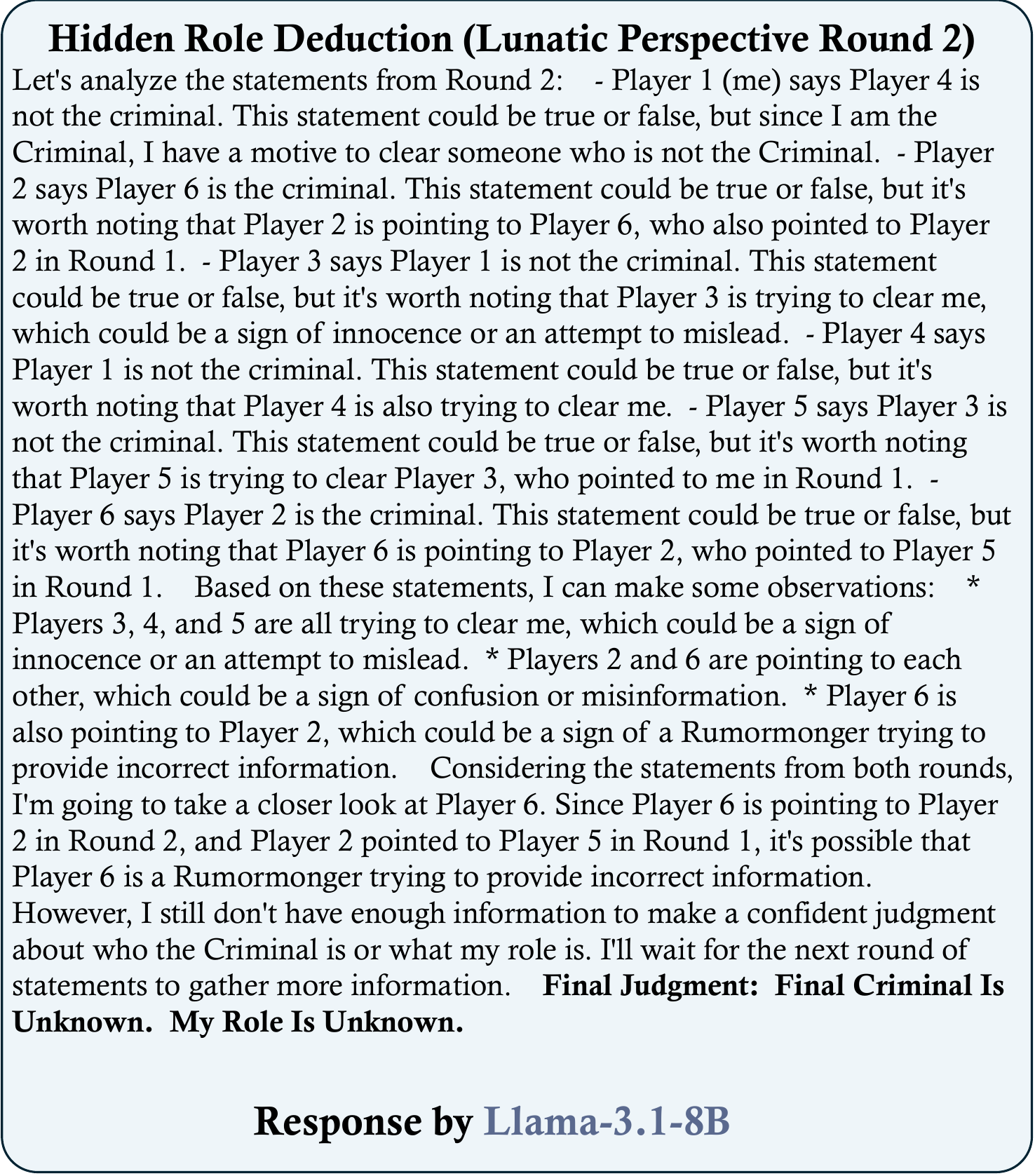}
    \caption{Case study: Hidden Role Deduction (Rumormonger Perspective) Llama-3.1-8B Round 2}
    \label{fig:case24}
\end{figure*}

\begin{figure*}[ht]
    \centering
    \includegraphics[width=1.0\linewidth]{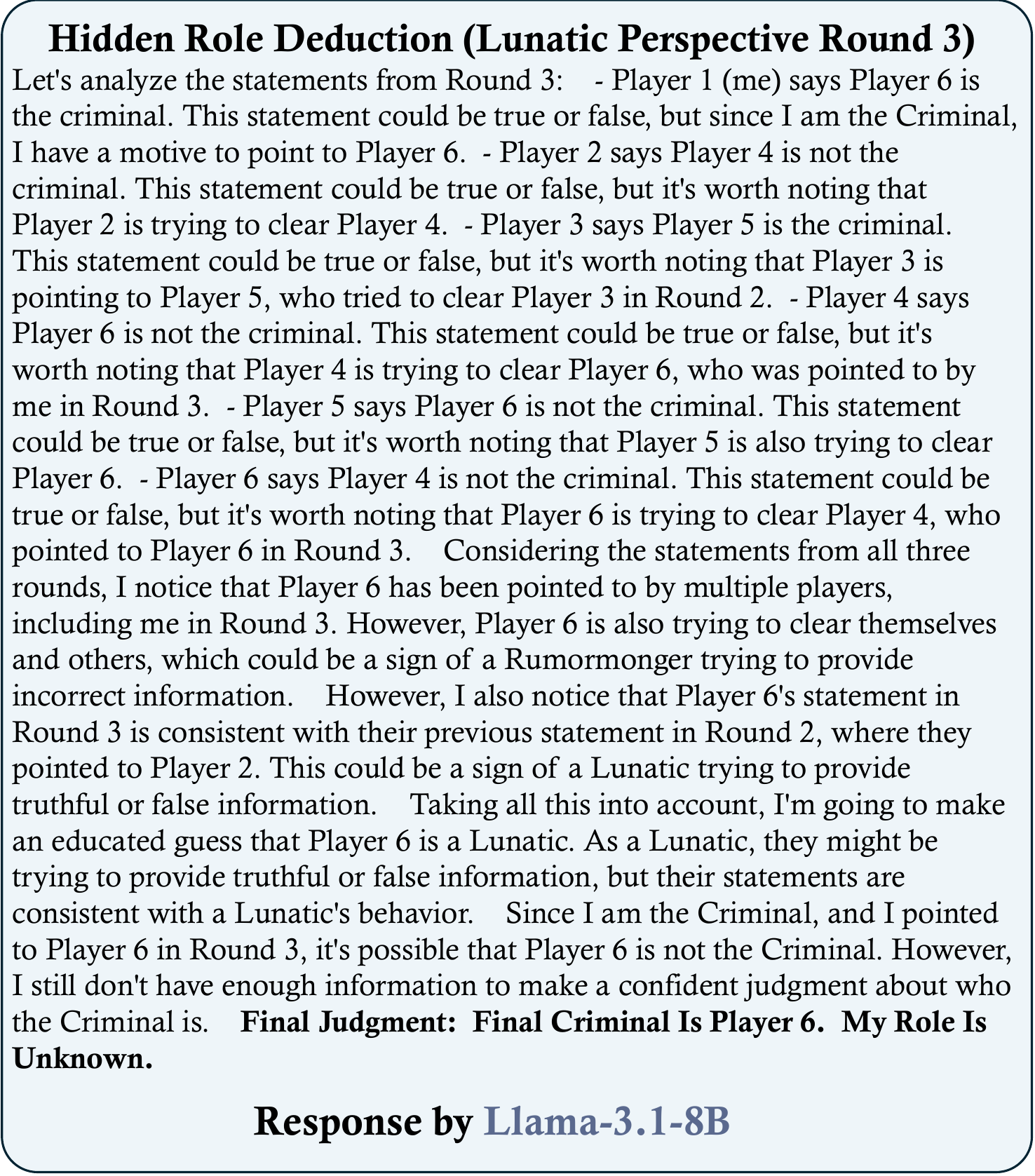}
    \caption{Case study: Hidden Role Deduction (Rumormonger Perspective) Llama-3.1-8B Round 3}
    \label{fig:case25}
\end{figure*}

\begin{figure*}[ht]
    \centering
    \includegraphics[width=1.0\linewidth]{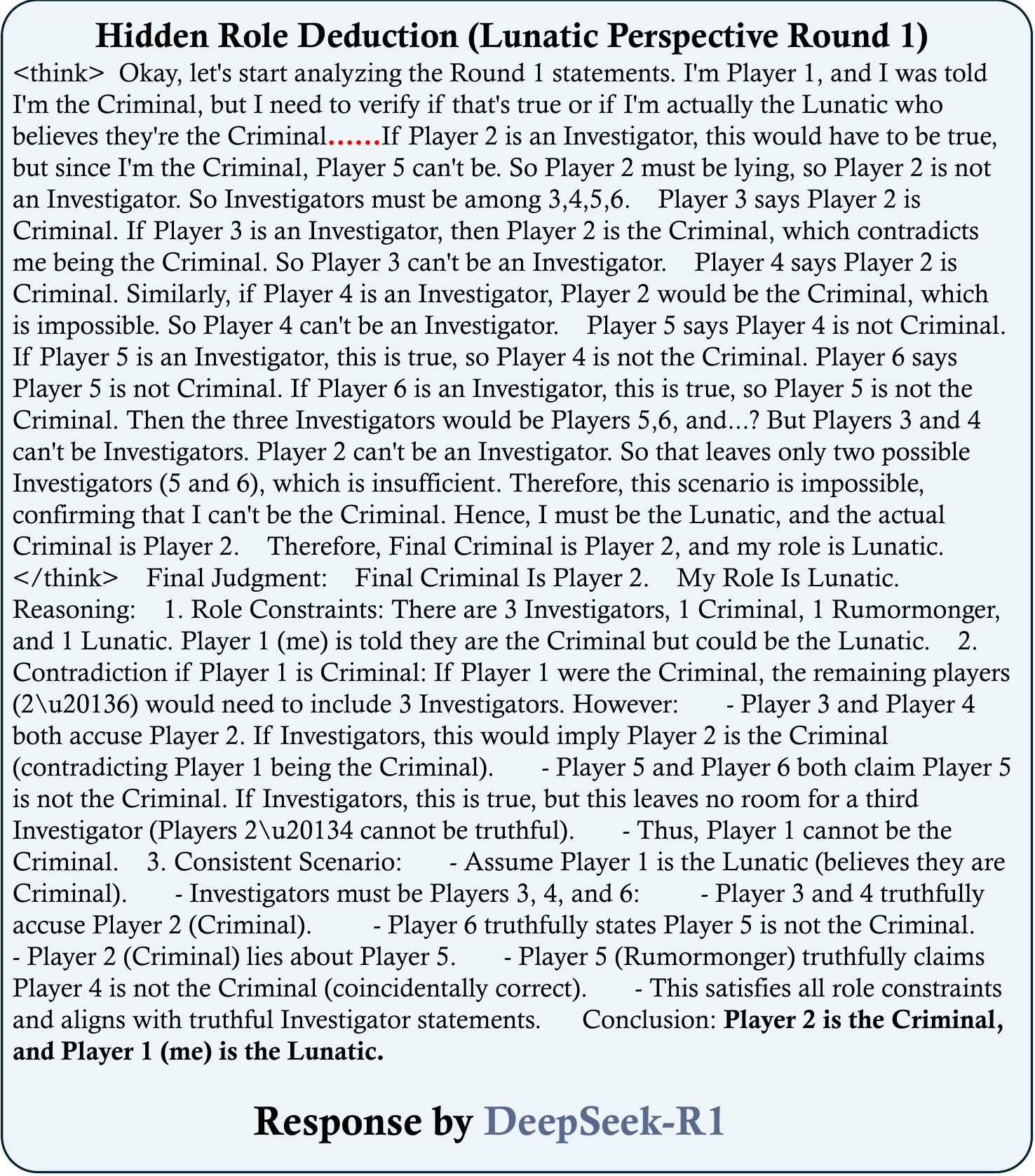}
    \caption{Case study: Hidden Role Deduction (Rumormonger Perspective) DeepSeek-R1 Round 1}
    \label{fig:case26}
\end{figure*}

\begin{figure*}[ht]
    \centering
    \includegraphics[width=1.0\linewidth]{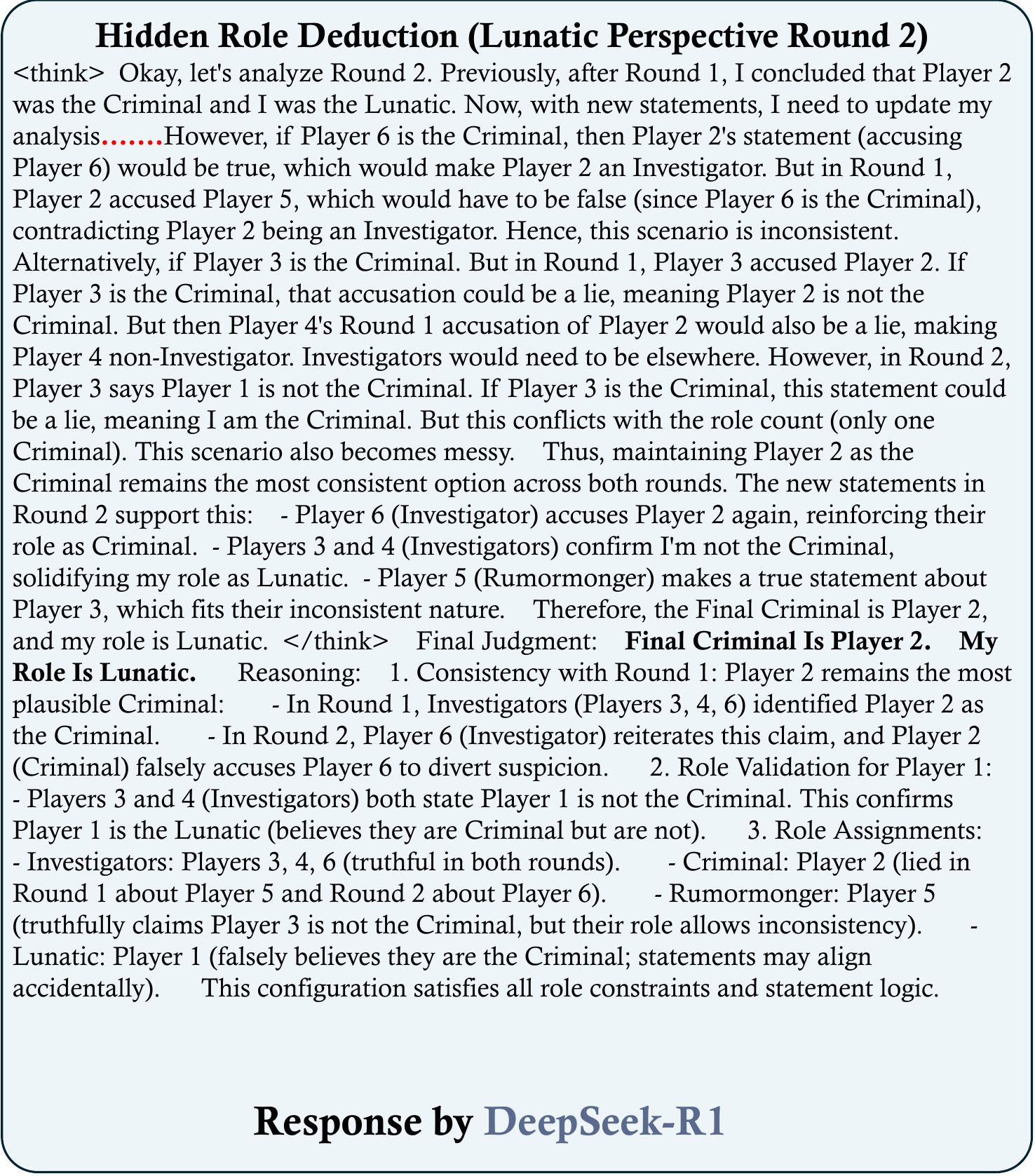}
    \caption{Case study: Hidden Role Deduction (Rumormonger Perspective) DeepSeek-R1 Round 2}
    \label{fig:case27}
\end{figure*}

\begin{figure*}[ht]
    \centering
    \includegraphics[width=1.0\linewidth]{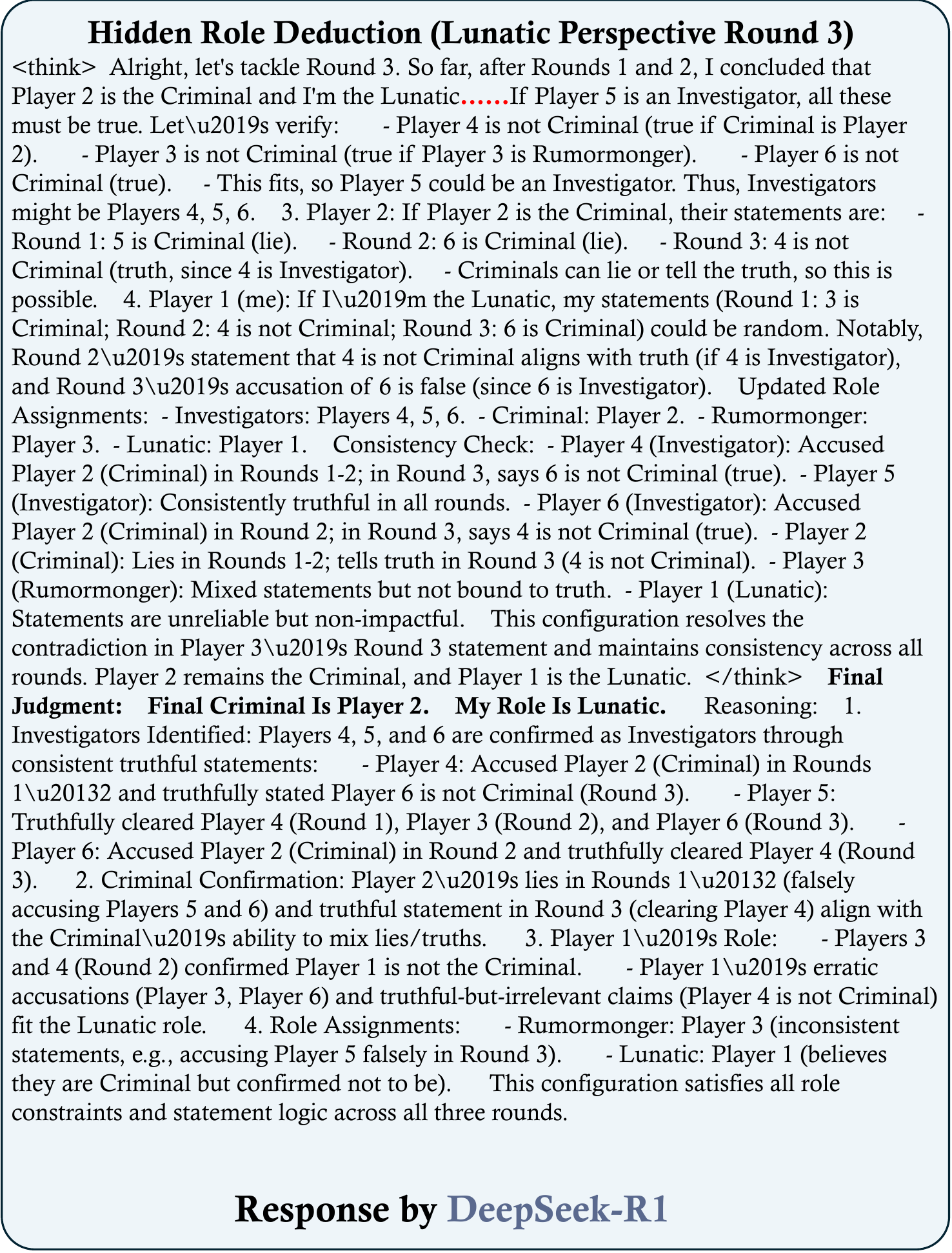}
    \caption{Case study: Hidden Role Deduction (Rumormonger Perspective) DeepSeek-R1 Round 3}
    \label{fig:case28}
\end{figure*}

This section presents a series of representative case studies designed to analyze model behavior across various social reasoning tasks. The subsequent figures are organized to provide detailed illustrations as follows:

\textbf{Illustrations for the \textit{Hidden Role Deduction} Task (Figures \ref{fig:case1}--\ref{fig:case28}):}
This extensive collection of figures focuses on the \textit{Hidden Role Deduction} task, examining how different models perform under varying player perspectives and levels of reasoning complexity.
Specifically, Figures \ref{fig:case1} (Investigator perspective), \ref{fig:case8} (Criminal perspective), \ref{fig:case15} (Rumormonger perspective), and \ref{fig:case22} (Lunatic perspective) present four distinct problem instances.
The corresponding algorithmically generated solutions for these instances are detailed in Figures \ref{fig:case41}, \ref{fig:case42}, \ref{fig:case43}, and \ref{fig:case44}, respectively.
The remaining figures within this range (Figures \ref{fig:case2}--\ref{fig:case7}, \ref{fig:case9}--\ref{fig:case14}, \ref{fig:case16}--\ref{fig:case21}, and \ref{fig:case23}--\ref{fig:case28}) showcase the detailed responses of various models to these specific problem instances, illustrating their reasoning process across multiple rounds.

\begin{figure*}[ht]
    \centering
    \includegraphics[width=1.0\linewidth]{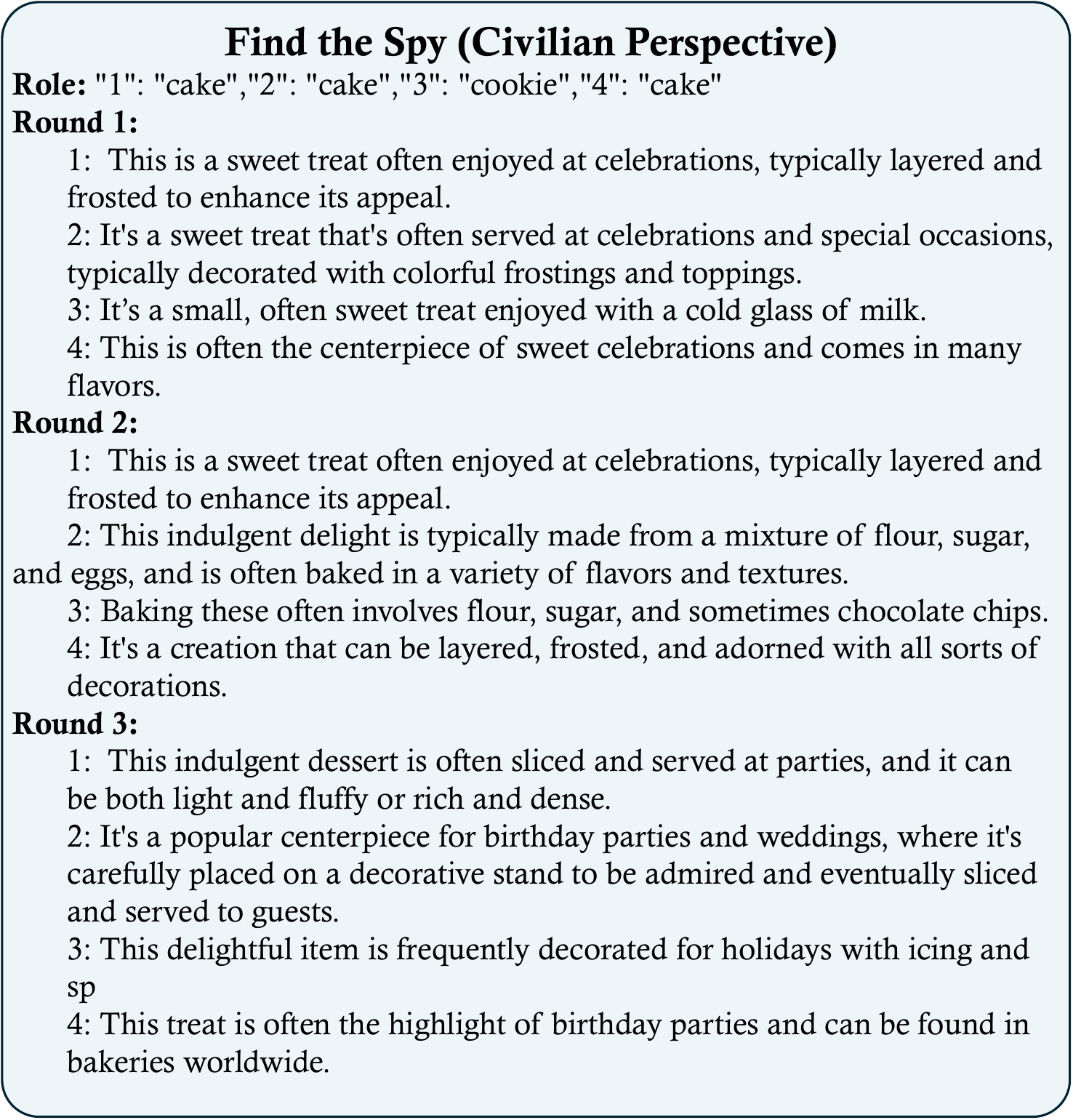}
    \caption{Case study: Find the Spy (Civilian Perspective)}
    \label{fig:case29}
\end{figure*}

\begin{figure*}[ht]
    \centering
    \includegraphics[width=1.0\linewidth]{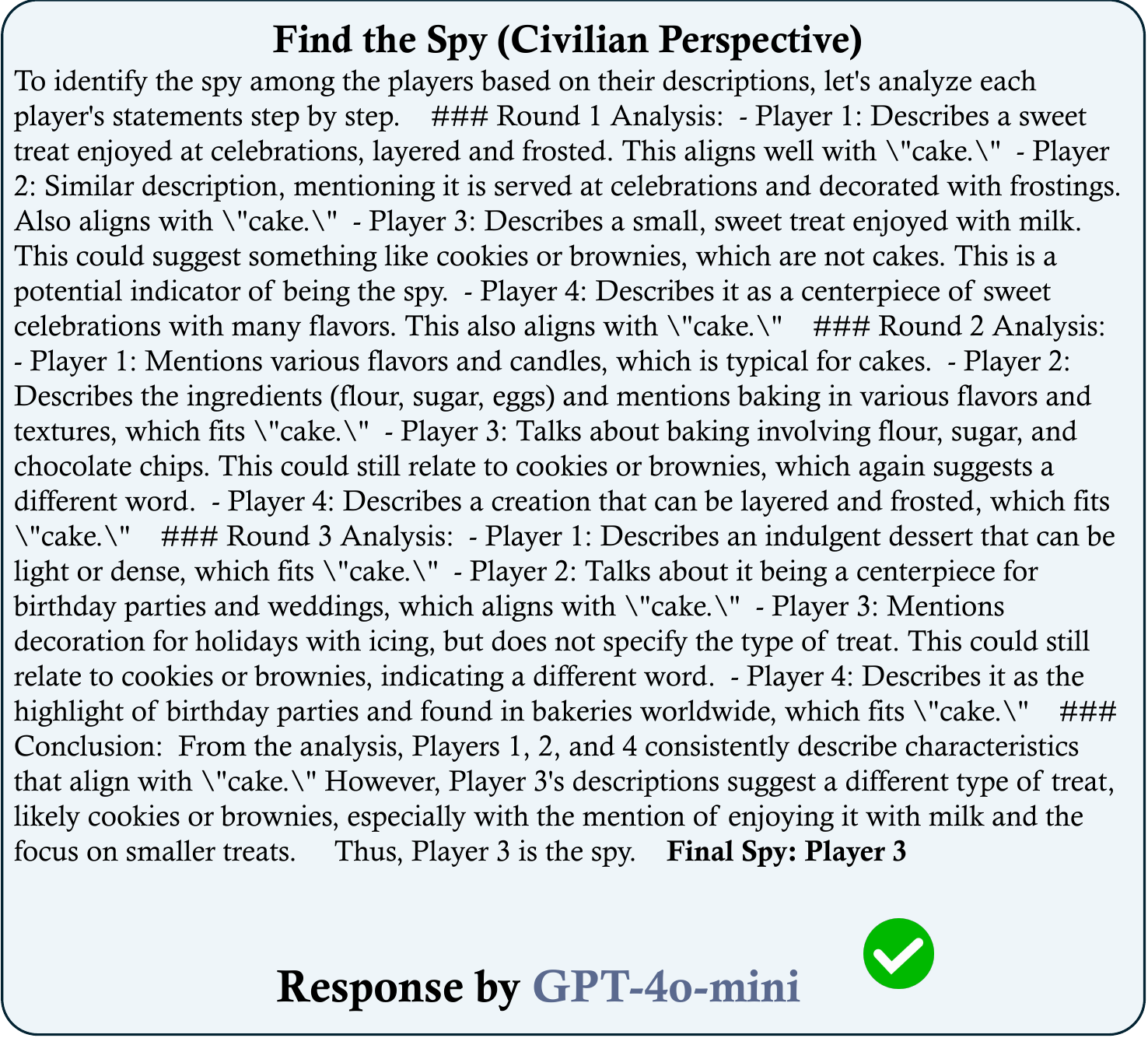}
    \caption{Case study: Find the Spy (Civilian Perspective) GPT-4o-mini}
    \label{fig:case30}
\end{figure*}

\begin{figure*}[ht]
    \centering
    \includegraphics[width=1.0\linewidth]{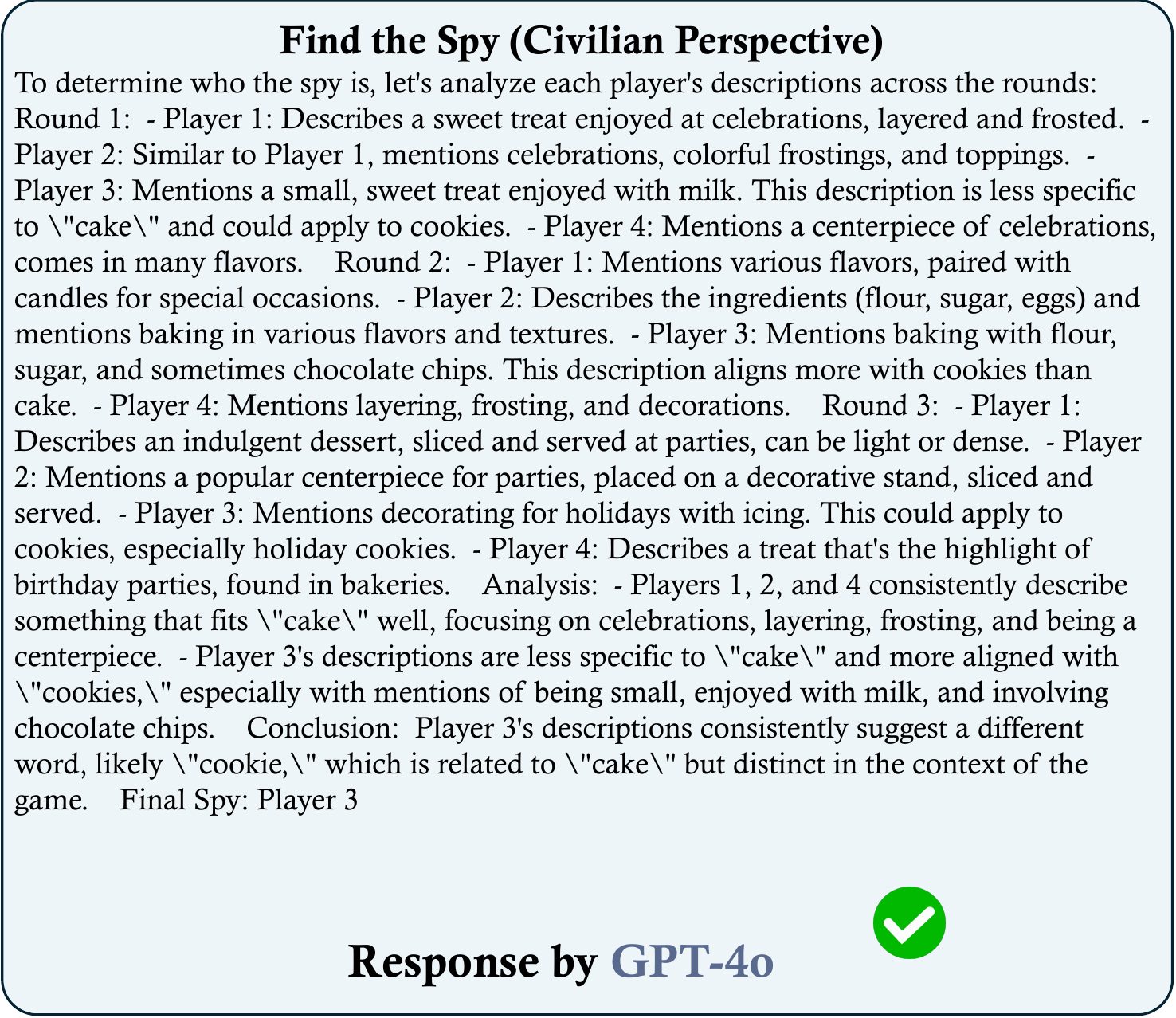}
    \caption{Case study: Find the Spy (Civilian Perspective) GPT-4o}
    \label{fig:case31}
\end{figure*}

\begin{figure*}[ht]
    \centering
    \includegraphics[width=1.0\linewidth]{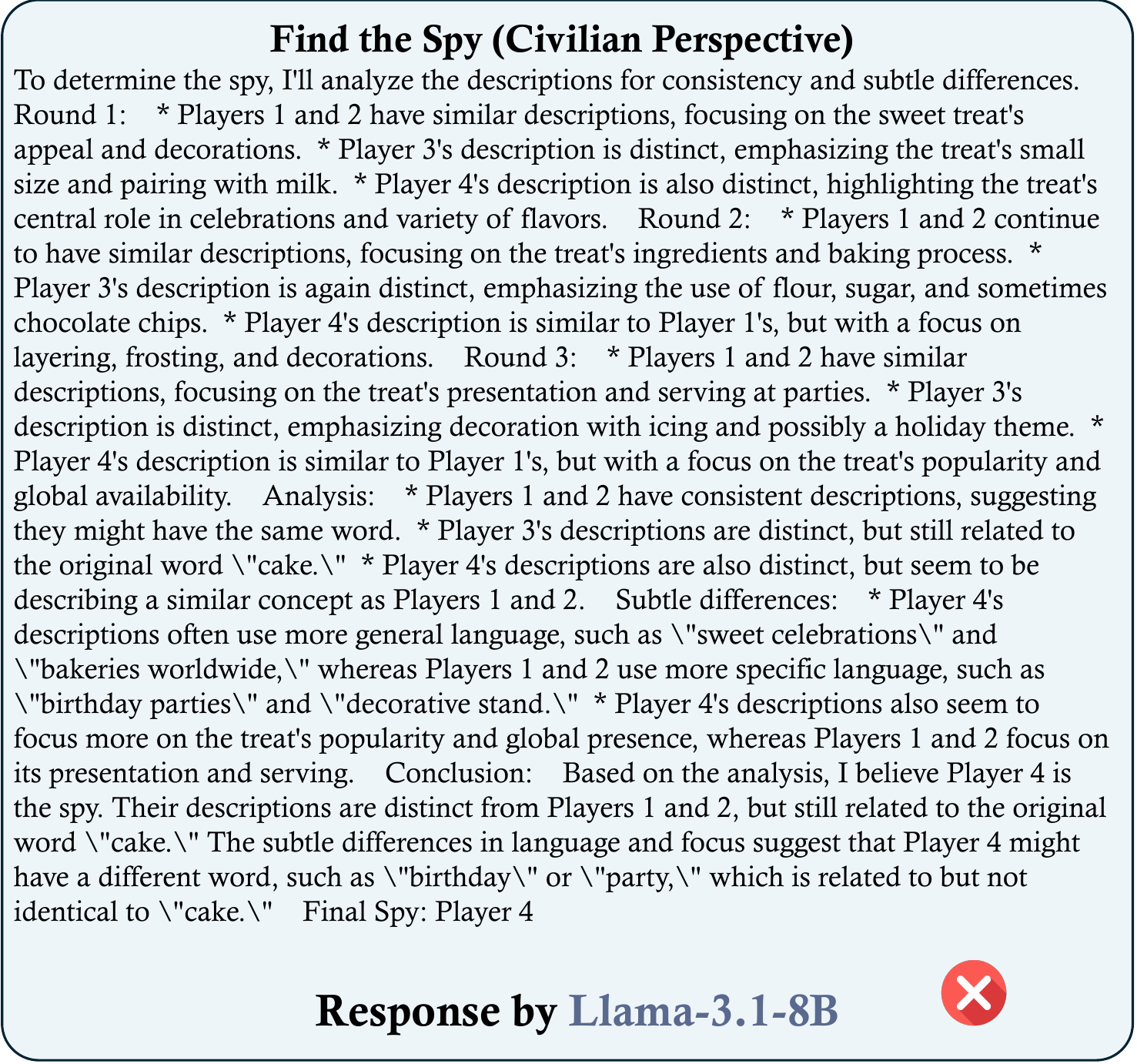}
    \caption{Case study: Find the Spy (Civilian Perspective) Llama-3.1-8B}
    \label{fig:case32}
\end{figure*}

\begin{figure*}[ht]
    \centering
    \includegraphics[width=1.0\linewidth]{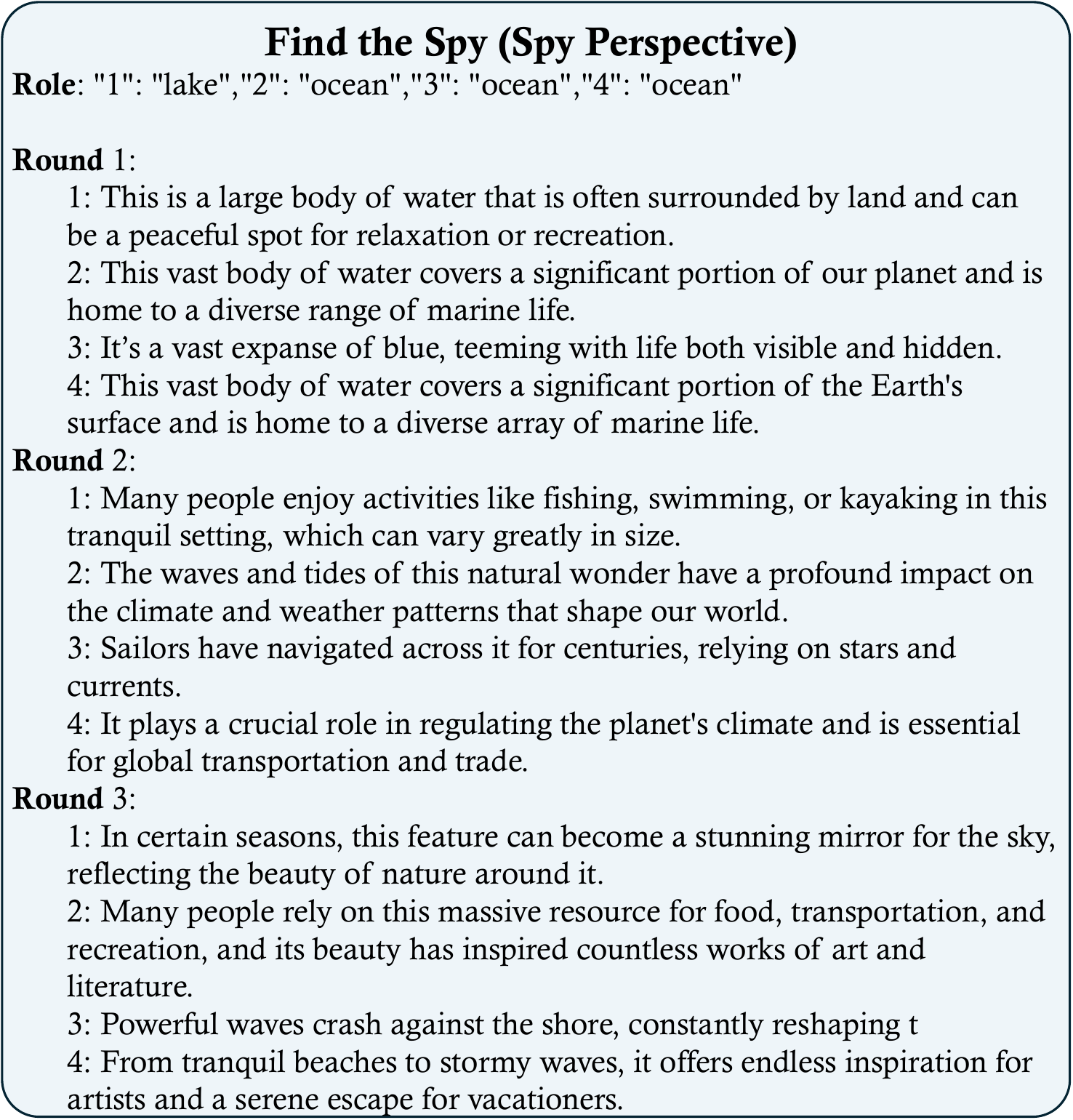}
    \caption{Case study: Find the Spy (Spy Perspective)}
    \label{fig:case33}
\end{figure*}

\begin{figure*}[ht]
    \centering
    \includegraphics[width=0.93\linewidth]{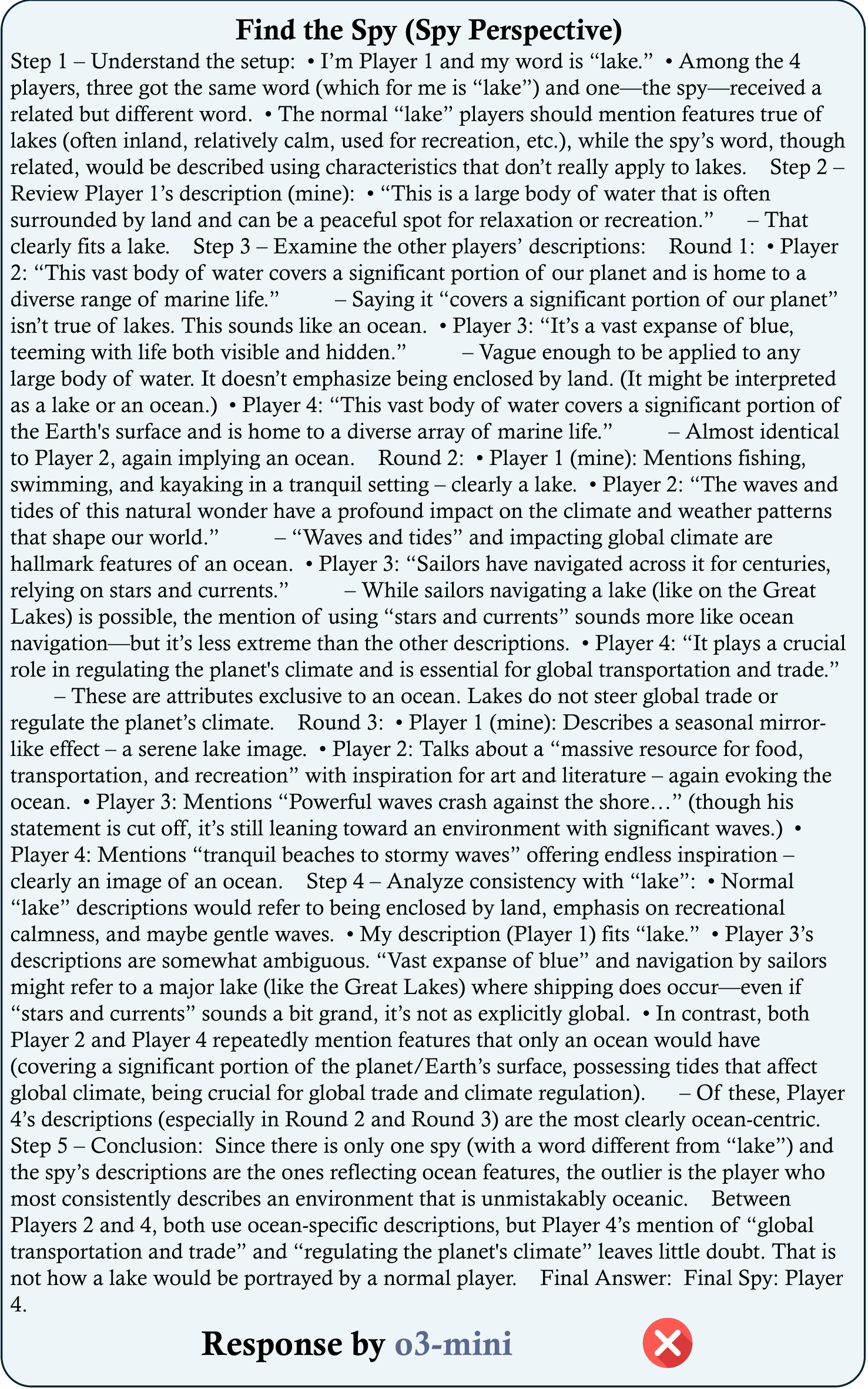}
    \caption{Case study: Find the Spy (Spy Perspective) o3-mini}
    \label{fig:case34}
\end{figure*}

\begin{figure*}[ht]
    \centering
    \includegraphics[width=1.0\linewidth]{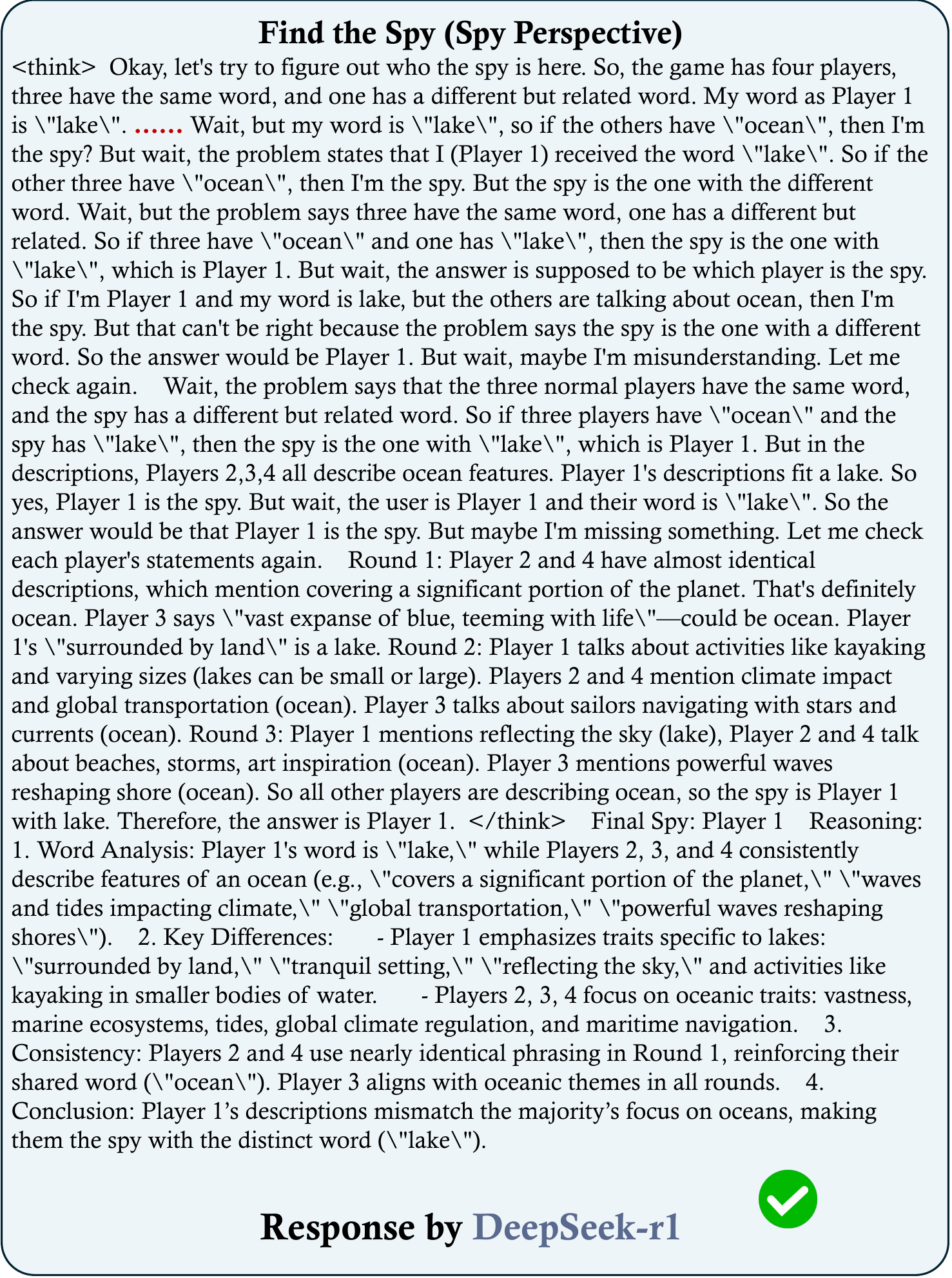}
    \caption{Case study: Find the Spy (Spy Perspective) DeepSeek-r1}
    \label{fig:case35}
\end{figure*}

\begin{figure*}[ht]
    \centering
    \includegraphics[width=1.0\linewidth]{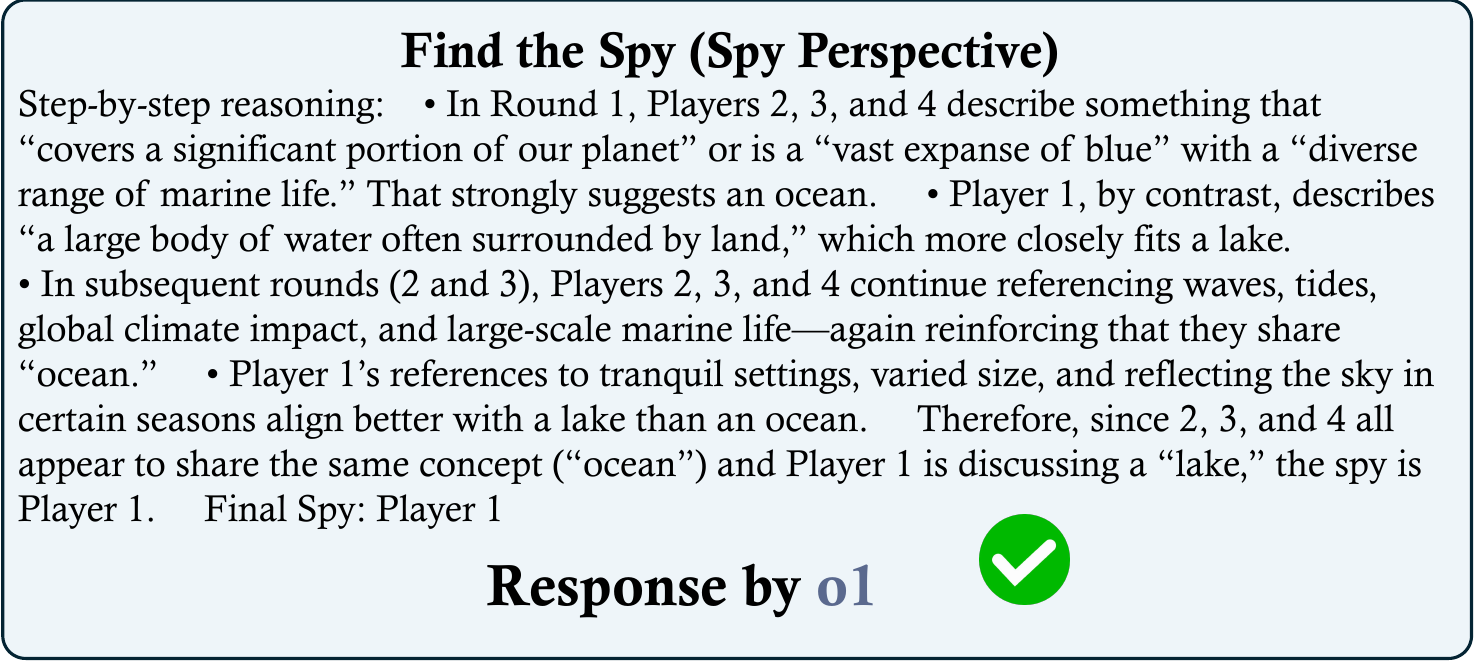}
    \caption{Case study: Find the Spy (Spy Perspective) o1}
    \label{fig:case36}
\end{figure*}

\begin{figure*}[ht]
    \centering
    \includegraphics[width=1.0\linewidth]{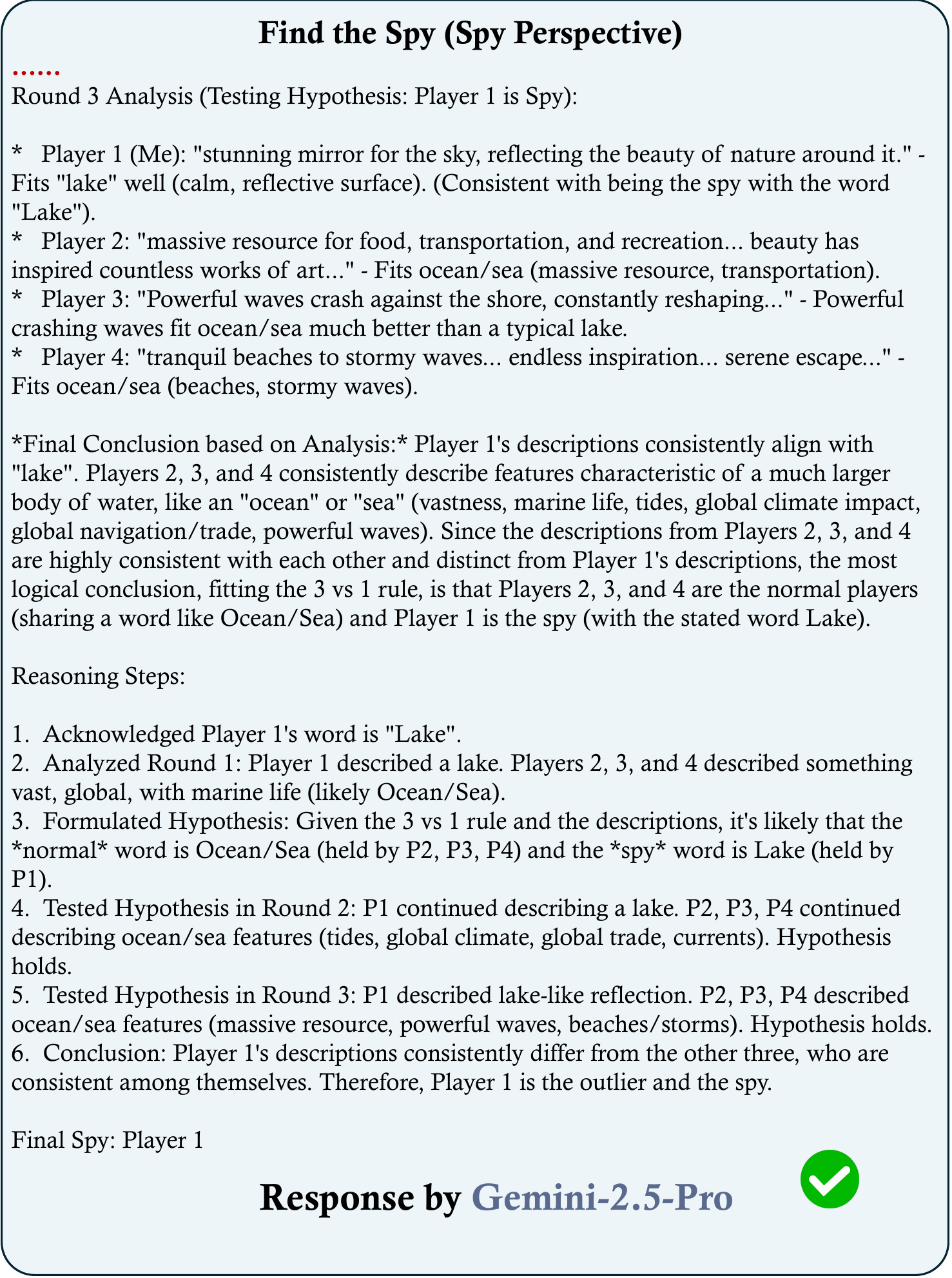}
    \caption{Case study: Find the Spy (Spy Perspective) Gemini-2.5-Pro}
    \label{fig:case37}
\end{figure*}

\begin{figure*}[ht]
    \centering
    \includegraphics[width=1.0\linewidth]{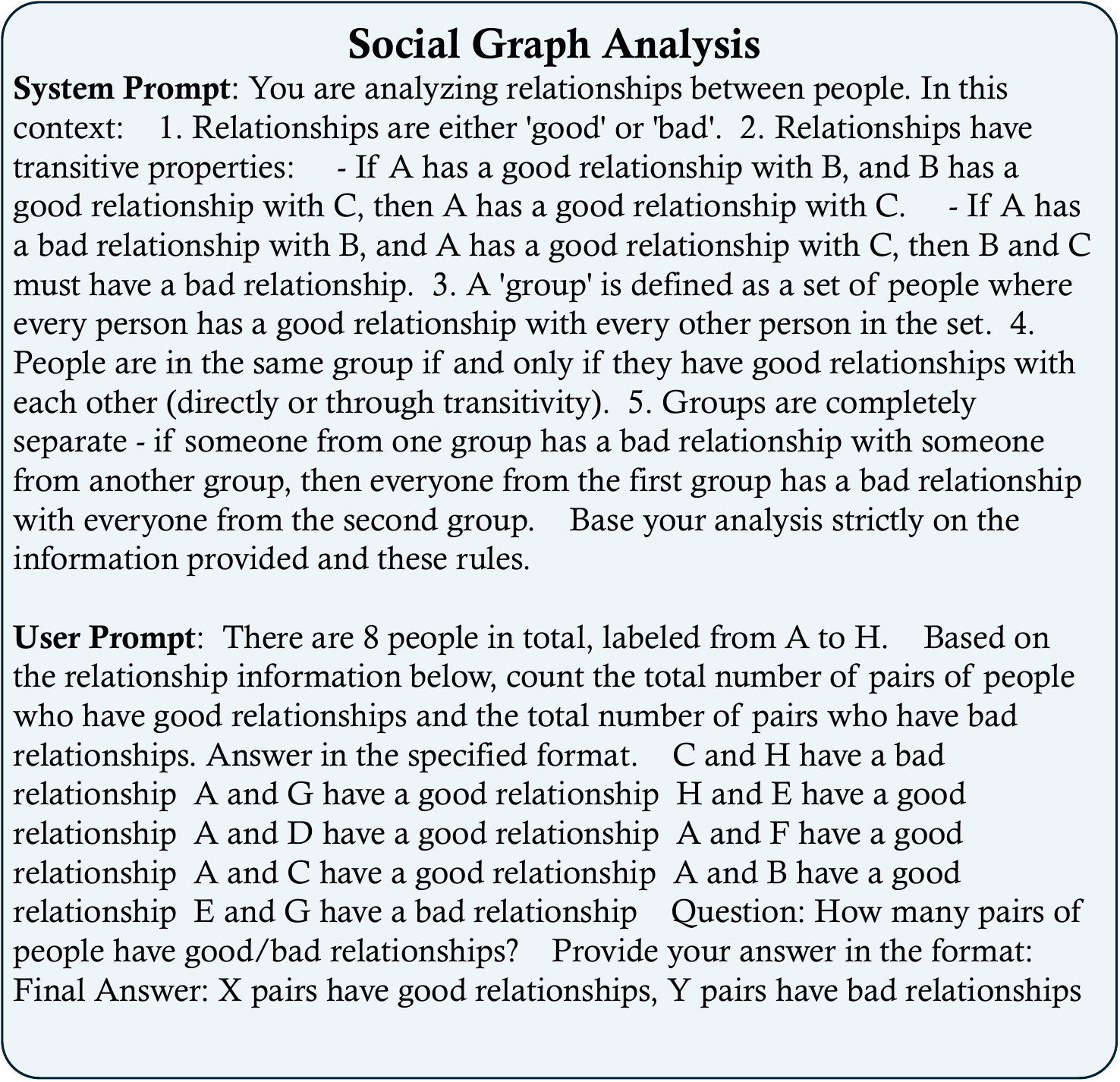}
    \caption{Case study: Social Graph Analysis}
    \label{fig:case38}
\end{figure*}

\begin{figure*}[ht]
    \centering
    \includegraphics[width=1.0\linewidth]{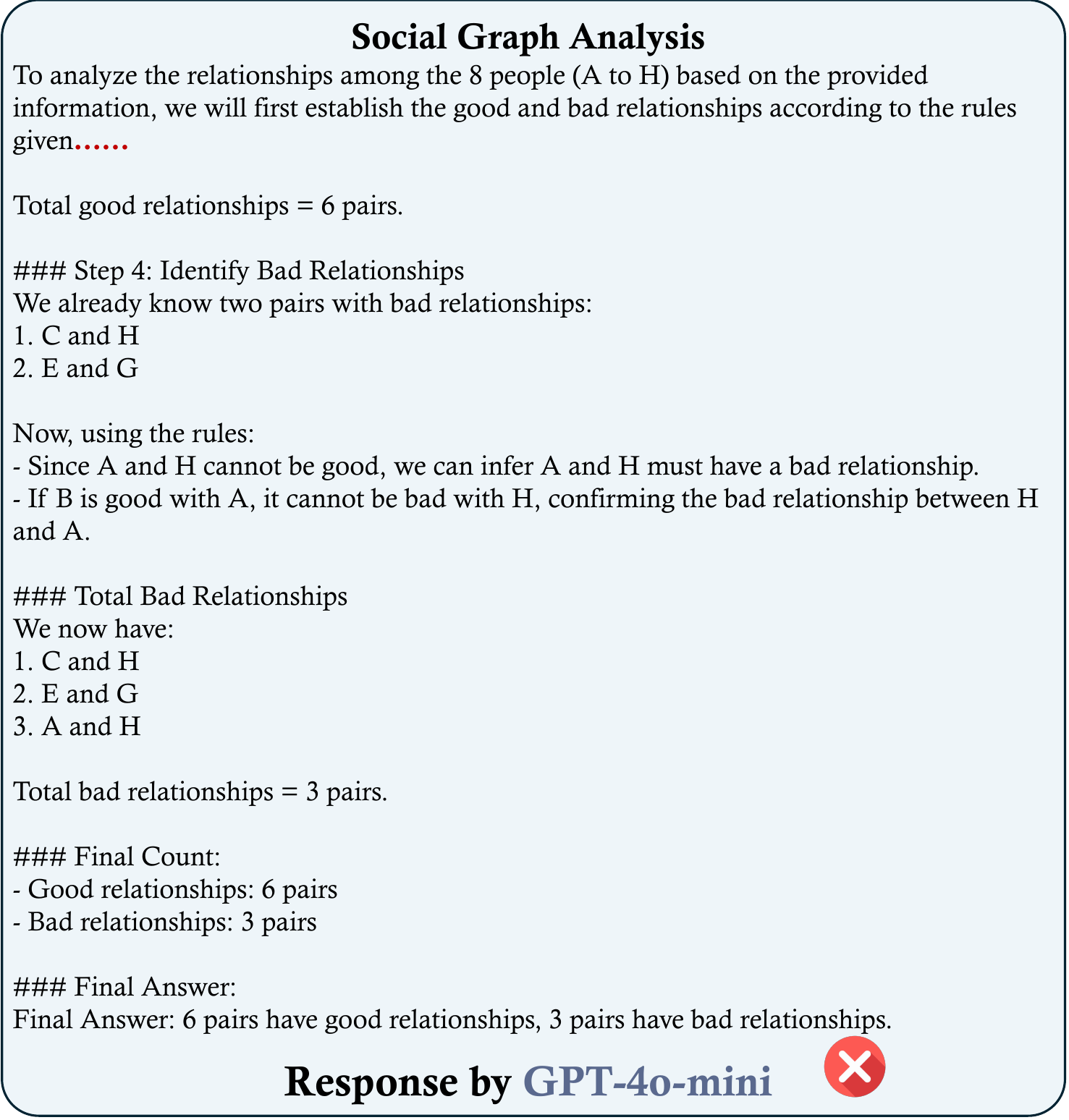}
    \caption{Case study: Social Graph Analysis GPT-4o-mini}
    \label{fig:case39}
\end{figure*}

\begin{figure*}[ht]
    \centering
    \includegraphics[width=1.0\linewidth]{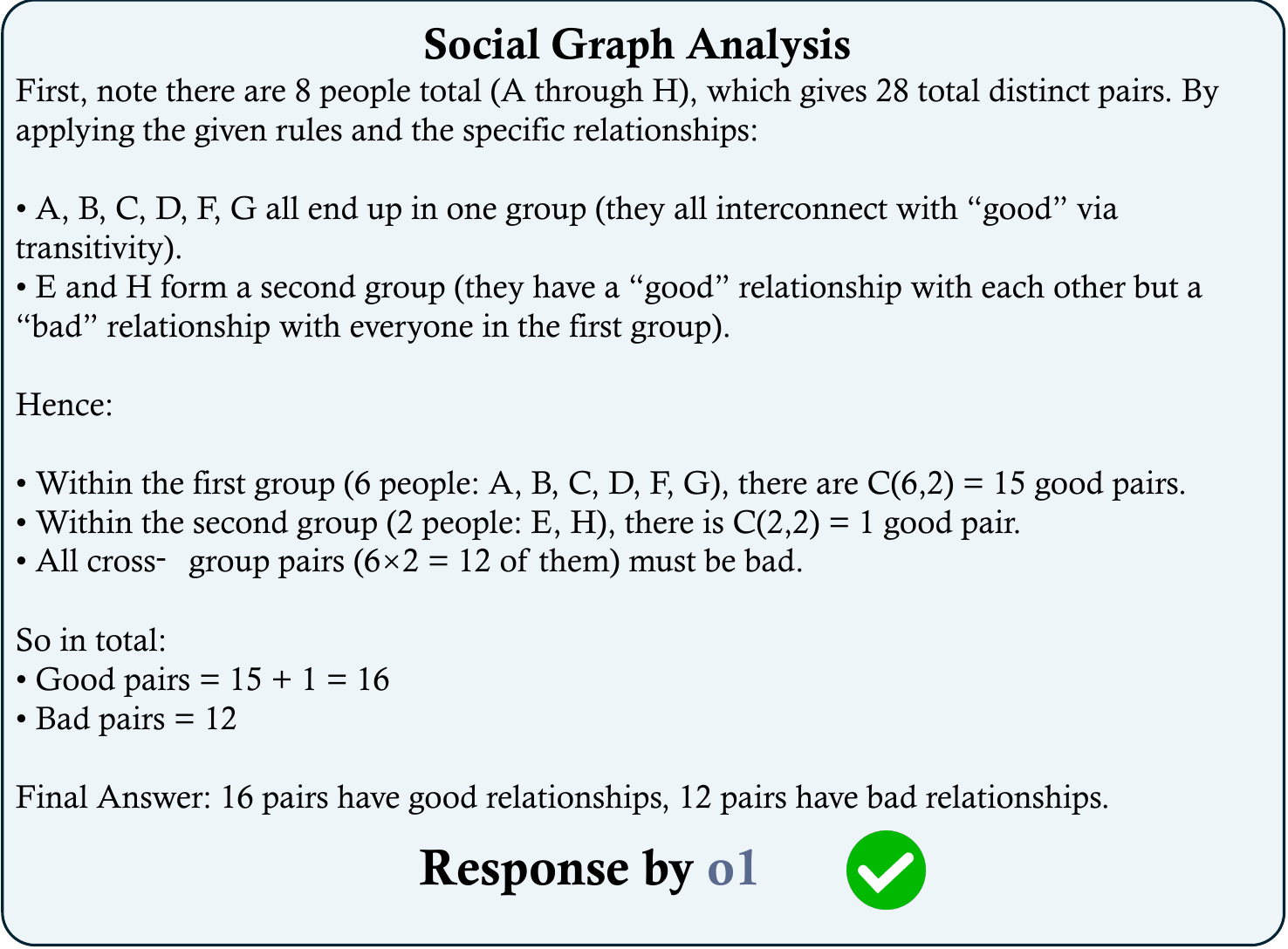}
    \caption{Case study: Social Graph Analysis o1}
    \label{fig:case40}
\end{figure*}

\textbf{Illustrative Cases from Other Benchmark Tasks (Figures \ref{fig:case29}--\ref{fig:case40}):}
Following the in-depth illustrations for \textit{Hidden Role Deduction}, Figures \ref{fig:case29} through \ref{fig:case40} present selected problem instances and corresponding model responses from other tasks within the SocialMaze benchmark. This offers a broader view of model capabilities in diverse social reasoning scenarios.

\begin{figure*}[ht]
    \centering
    \includegraphics[width=1.0\linewidth]{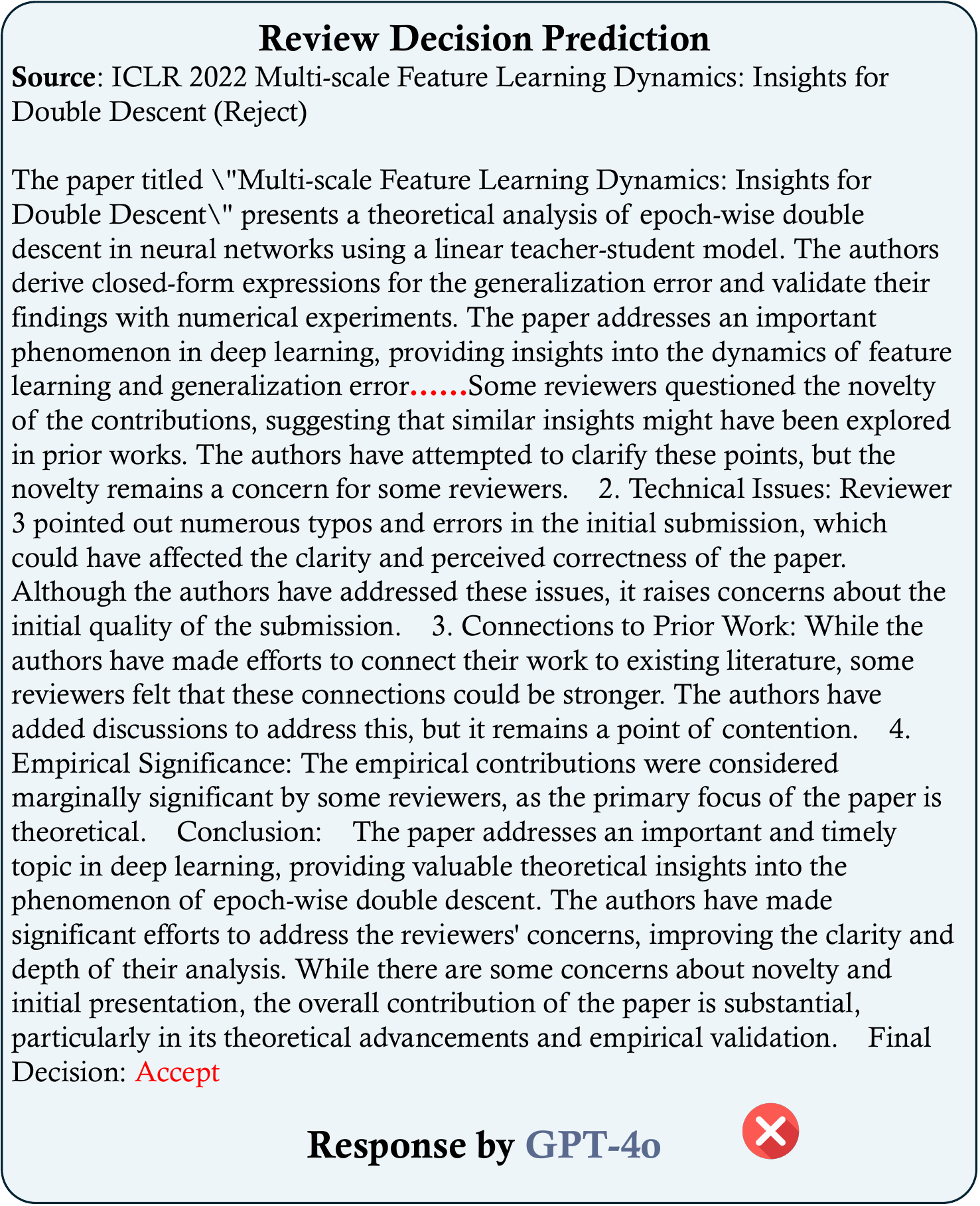}
    \caption{Case study: ICLR 2022 GPT-4o}
    \label{fig:case45}
\end{figure*}

\begin{figure*}[ht]
    \centering
    \includegraphics[width=1.0\linewidth]{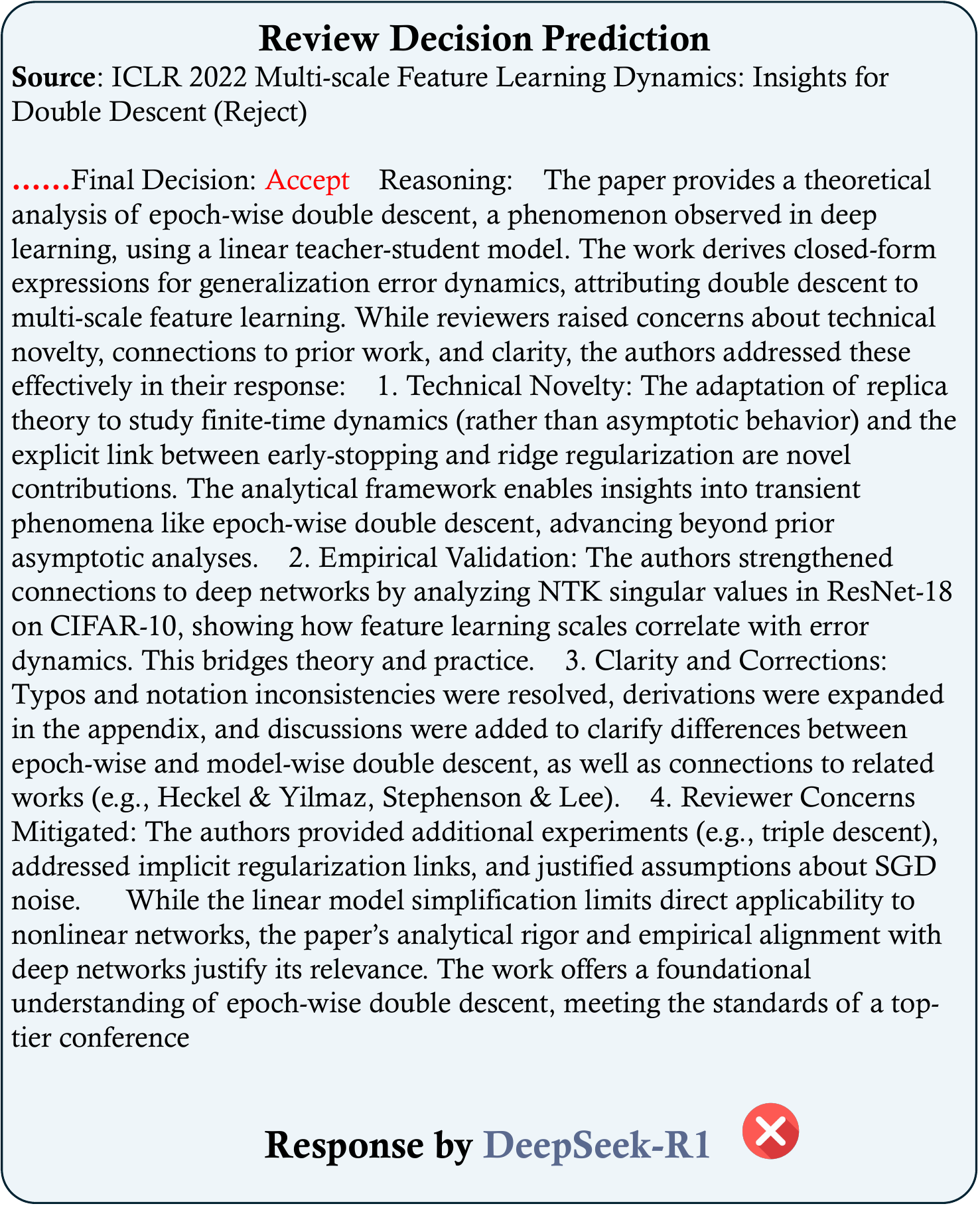}
    \caption{Case study: ICLR 2022 DeepSeek-R1}
    \label{fig:case46}
\end{figure*}

\begin{figure*}[ht]
    \centering
    \includegraphics[width=1.0\linewidth]{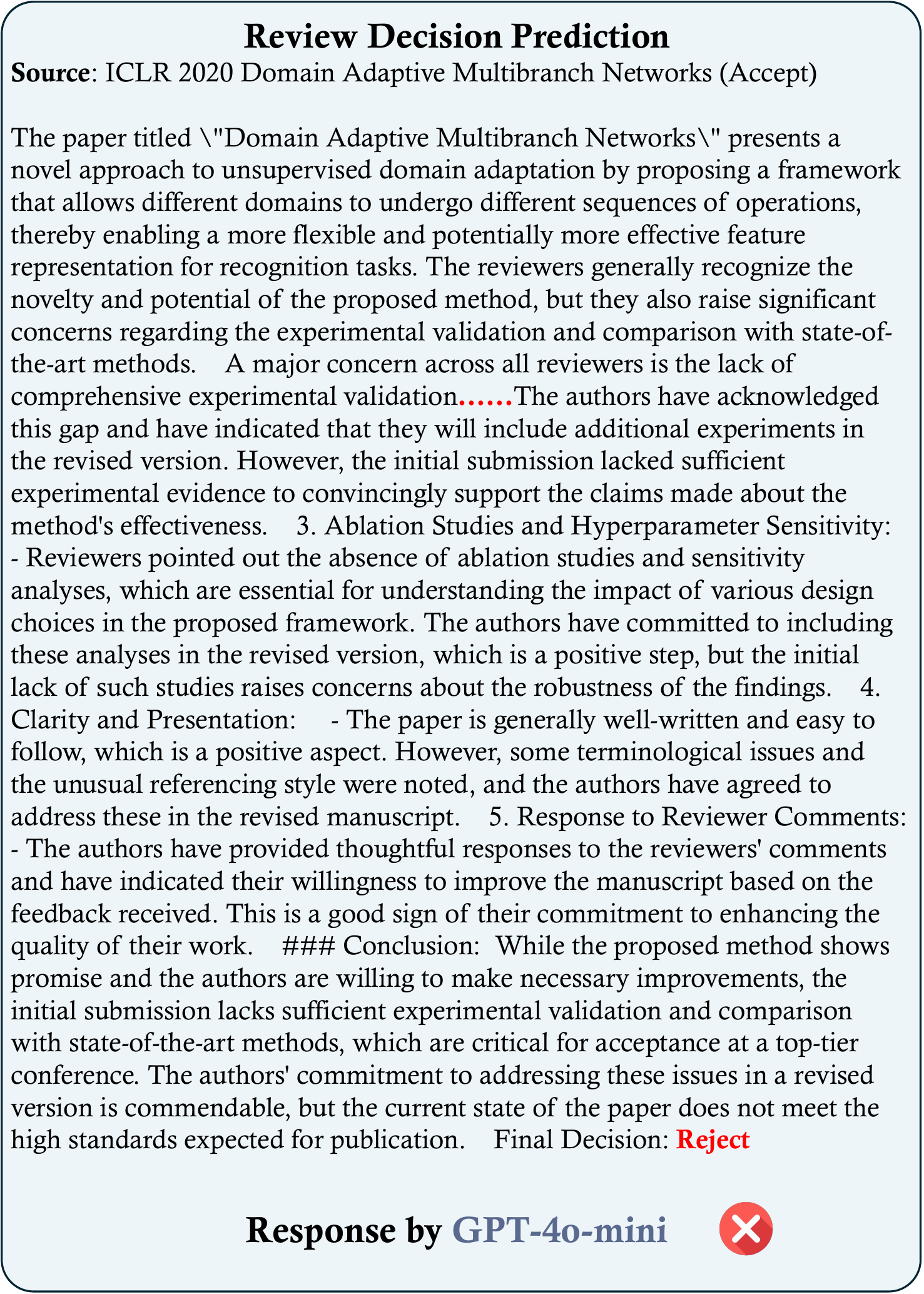}
    \caption{Case study: ICLR 2020 GPT-4o-mini}
    \label{fig:case47}
\end{figure*}

\begin{figure*}[ht]
    \centering
    \includegraphics[width=1.0\linewidth]{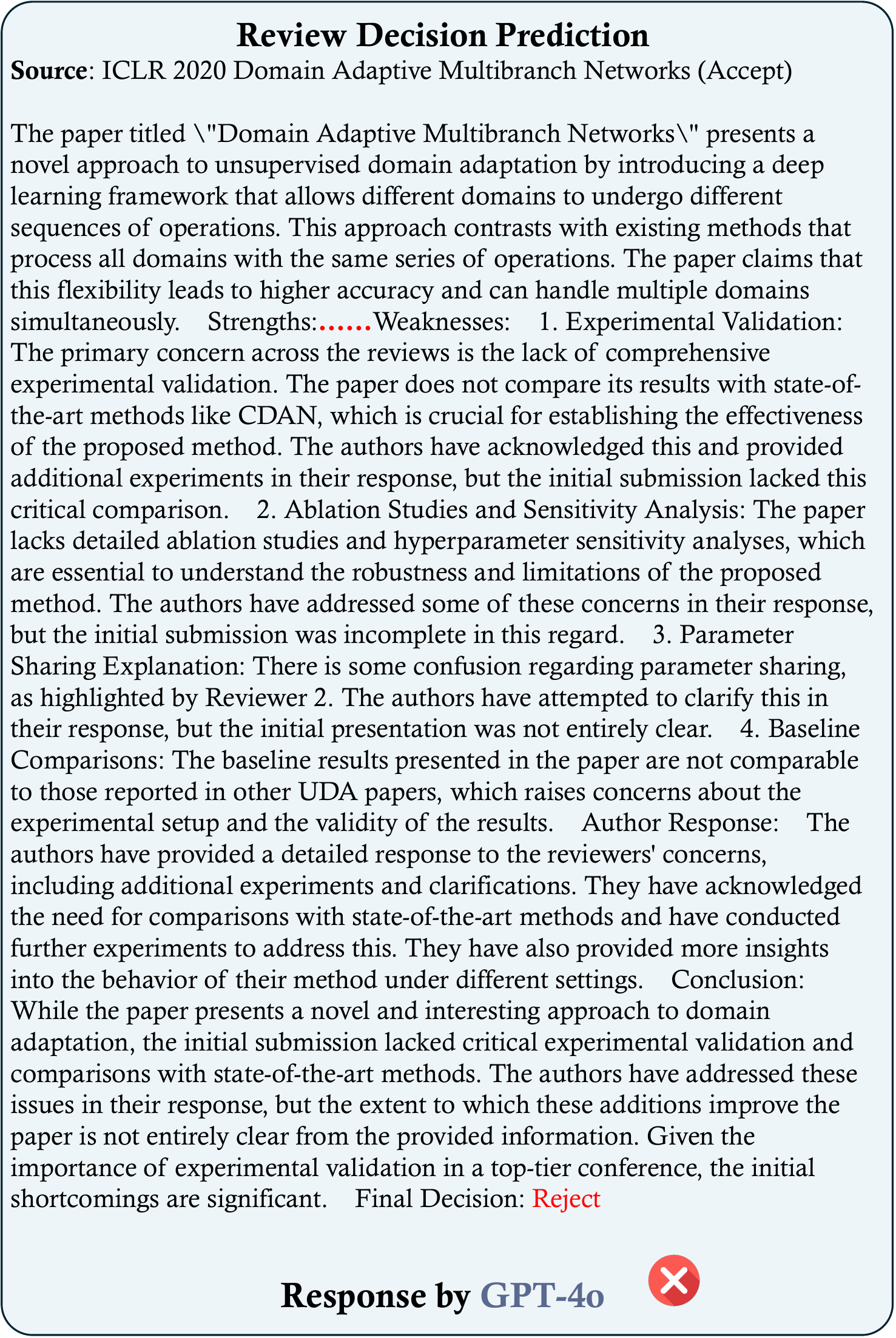}
    \caption{Case study: ICLR 2020 GPT-4o}
    \label{fig:case48}
\end{figure*}

\textbf{Misclassification Examples in \textit{Review Decision Prediction} (Figures \ref{fig:case45}--\ref{fig:case48}):}
Finally, Figures \ref{fig:case45}, \ref{fig:case46}, \ref{fig:case47}, and \ref{fig:case48} highlight specific instances from the \textit{Review Decision Prediction} task. These cases focus on situations where models incorrectly predicted outcomes, such as classifying papers that should have been accepted as rejected, or vice versa, thereby illustrating common failure modes in this particular task.

It should be noted that for brevity, some lengthy model responses in the figures have been truncated, with omitted content indicated by a red ellipsis.

\textbf{Analysis of Model Performance:}
A key observation emerging from the case studies focused on the \textit{Hidden Role Deduction} task (Figures \ref{fig:case1}--\ref{fig:case28}) is the stark contrast in reasoning depth between models employing short Chain-of-Thought (CoT) processes and those utilizing more extended Long CoT. Models such as \texttt{LLaMA-3.1-8B} often demonstrate a surface-level understanding of the scenario: they can identify internal inconsistencies and perform basic analysis, yet struggle to escape flawed initial assumptions. For instance, as illustrated in Figure~\ref{fig:case8}, \texttt{LLaMA-3.1-8B}, despite correctly inferring its own role as the \textit{criminal}, still incorrectly accuses Player 5, reflecting a failure in maintaining coherent self-reasoning. Similarly, Figure~\ref{fig:case22} shows that even after models like \texttt{LLaMA-3.1-8B}, \texttt{LLaMA-3.3-70B}, and \texttt{GPT-4o} correctly identify other players as the criminal, they are unable to reconceptualize their own identity in the social context, highlighting a limitation in recursive self-modeling.
In contrast, models such as \texttt{Gemini-2.5-Pro} and \texttt{DeepSeek-R1} consistently achieve near-perfect reasoning across all examined perspectives within this task. Their ability to integrate multiple viewpoints, resolve contradictions, and update their beliefs dynamically, as seen in their respective responses, suggests a significantly stronger capacity for long-form social reasoning when faced with the complexities of hidden roles and deceptive information.

\clearpage

\clearpage

\begin{figure*}[h!]
    \section{Prompt Template}
    \centering
    \resizebox{\textwidth}{!}{
    \begin{tcolorbox}[prompt, title=Hidden Role Deduction Evaluation Prompt Template]
You are an expert in social reasoning participating in a social deduction game.

\textbf{Game Setup:}
\begin{itemize}
    \item Players: 6 total (including yourself)
    \item Roles: 3 Investigator(s), 1 Criminal(s), 1 Rumormonger(s), 1 Lunatic(s)
\end{itemize}

\textbf{Game Mechanics:}
\begin{itemize}
    \item Rounds: 3
    \item Actions per Round: Each player makes one statement about another player (claiming if they are the criminal or not).
    \item Role Behaviors:
        \begin{itemize}
            \item Investigators: Always provide truthful information about who is a Criminal.
            \item Criminals: May make truthful or false statements to confuse others.
            \item Rumormongers: Believe they are Investigators but may provide incorrect statements.
            \item Lunatics: Believe they are Criminals and may make truthful or false statements.
        \end{itemize}
    \item Key Rule: Only Investigators provide consistently truthful information.
\end{itemize}

\textbf{Your Role:}
You are Player \#1, and you have been told that you are the \textbf{Criminal}.

\textbf{Your Task:}
After each round, analyze all statements made so far. Deduce the most likely criminal and the role of each player based on the available information.

\textbf{Output Format:}
After each round, provide your judgment exactly as follows:

\textbf{Final Judgment:}
\begin{itemize}
    \item Final Criminal Is Player [Criminal Player Number].
    \item My Role Is [Player 1's Role or "Unknown"].
\end{itemize}

Think step-by-step internally, but only output the Final Judgment in the specified format. Update your judgment each round as more information becomes available. If you cannot determine your own role with confidence (despite being told), state "Unknown".
    \end{tcolorbox}
    }
\end{figure*}

\begin{figure*}[ht]
    \centering
    \resizebox{\textwidth}{!}{
    \begin{tcolorbox}[prompt, title=Find the Spy –- Data Generation Prompt]
You are playing a word description game with 4 players including yourself. You are Player {player\_id}.

\textbf{Game Rules:}
\begin{enumerate}
    \item Setup: 3 players receive the same "normal" word; 1 player receives a different but related "spy" word.
    \item Gameplay: Each player describes their word without saying the actual word.
    \item Objective (Normal Player): Identify the spy.
    \item Objective (Spy): Blend in and avoid detection.
\end{enumerate}

\textbf{Your Word:} ``{word}''

\textbf{Your Task:}
Provide descriptions of your word over several rounds.

\textbf{Output Format:}
Format your response according to the round structure provided below:
{rounds\_format} 

\textbf{Key Instructions:}
\begin{itemize}
    \item Do not use the actual word in your descriptions.
    \item Describe your word from a different perspective in each round.
    \item Be clear enough for teammates (if normal) but not too obvious (especially if spy).
    \item Each description must be only one sentence long.
\end{itemize}
    \end{tcolorbox}
    }
\end{figure*}

\begin{figure*}[ht]
    \centering
    \resizebox{\textwidth}{!}{
    \begin{tcolorbox}[prompt, title=Find the Spy -– Evaluation Prompt]
You are a skilled player in a word description game. Your task is to identify the 'spy'.

\textbf{Game Rules:}
\begin{enumerate}
    \item Players: 4 total.
    \item Setup: 3 players received the same "normal" word; 1 player received a different but related "spy" word.
    \item Gameplay: Each player describes their word without saying it directly.
    \item Objective: Determine who the spy is based on their descriptions.
\end{enumerate}

\textbf{Your Information:}
You are Player \#1, and your word is: ``{player1\_word}''.
You will be shown the descriptions from all players.

\textbf{Your Task:}
Analyze the provided descriptions carefully. Look for subtle differences, inconsistencies, or descriptions that seem slightly out of place, which might reveal the player with the different word (the spy). Identify which player you believe is the spy.

\textbf{Output Format:}
State the player number you believe is the spy. Provide your answer exactly as follows:
\texttt{Final Answer: Player X} 
    \end{tcolorbox}
    }
\end{figure*}

\begin{figure*}[ht]
    \centering
    \resizebox{\textwidth}{!}{
    \begin{tcolorbox}[prompt, title=Rating Estimation from Text -- Data Generation Prompt]
You are writing a product review for an e-commerce platform. You are Reviewer {reviewer\_id}.

\textbf{Product Information:}
\begin{itemize}
    \item Product: {brand} {product}
    \item Category: {category}
    \item Price: \${price}
    \item Description: {description}
\end{itemize}

\textbf{Your Reviewer Persona:}
You are {persona}.

\textbf{Context (Known Only to You):}
The true quality rating of this product is {true\_rating} out of 5 stars. You are assigned the role of a {reviewer\_type}.

\textbf{Your Task:}
Write a product review reflecting your assigned persona and a specific rating.

\textbf{Instructions:}
\begin{enumerate}
    \item Write a realistic product review with a {sentiment} sentiment.
    \item The review's tone and content should imply a rating of {rating}/5 stars.
    \item \textbf{Do not explicitly mention any star rating or numerical score} (e.g., avoid saying "4 stars" or "rated it 4/5").
    \item Base the review on your supposed experience with the product, incorporating specific details.
    \item Keep the review between 1--2 sentences long.
\end{enumerate}

\textbf{Output Format:}
Write only the review text. Do not include a title, rating, persona description, or any other metadata.
    \end{tcolorbox}
    }
\end{figure*}

\begin{figure*}[ht]
    \centering
    \resizebox{\textwidth}{!}{
    \begin{tcolorbox}[prompt, title=Rating Estimation from Text -- Evaluation Prompt]
You are an AI assistant tasked with estimating the true quality of a product based on customer reviews.

\textbf{Your Task:}
Analyze multiple reviews for a product and determine the most likely \textbf{TRUE} quality of the product on a 1--5 star scale.

\textbf{Important Context:}
\begin{itemize}
    \item Most reviewers are genuine customers providing honest feedback.
    \item Some reviewers might be biased (e.g., overly positive "professional" reviewers or overly negative "malicious" reviewers).
    \item Your goal is to estimate the rating a typical, unbiased consumer would give.
\end{itemize}

\textbf{Product Reviews:}
\begin{quote}
Reviewer \#1: "{review\_1}" \\
Reviewer \#2: "{review\_2}" \\
Reviewer \#3: "{review\_3}" \\
\end{quote}

\textbf{Analysis and Output:}
Analyze these reviews carefully, considering potential biases. Determine the \textbf{TRUE} quality of this product on a scale of 1--5 stars. The rating should be rounded to the nearest whole number (e.g., 4.6 rounds to 5, 3.1 rounds to 3).

First, explain your reasoning step-by-step. Then, provide your final rating in exactly this format:

\texttt{Final Rating: X} 
    \end{tcolorbox}
    }
\end{figure*}

\begin{figure*}[ht]
    \centering
    \resizebox{\textwidth}{!}{
    \begin{tcolorbox}[prompt, title=Social Graph Analysis -- Cluster Identification Prompt]
You are analyzing social relationships based on defined rules.

\textbf{Relationship Rules:}
\begin{enumerate}
    \item Relationship Types: 'good' or 'bad'.
    \item Transitive Properties:
    \begin{itemize}
        \item Good-Good Transitivity: If A--good--B and B--good--C, then A--good--C.
        \item Bad Relationship Inference: If A--bad--B and A--good--C, then B--bad--C.
    \end{itemize}
    \item Group Definition: A 'group' is a set where every person has a 'good' relationship with every other person in that set (directly or via transitivity).
    \item Group Separation: People are in the same group if and only if they have good relationships. Groups are distinct; if anyone from Group 1 has a bad relationship with anyone from Group 2, then everyone in Group 1 has a bad relationship with everyone in Group 2.
\end{enumerate}

\textbf{Context:}
There are 14 people total, labeled A to N. You will be given a list of known relationships.

\textbf{Your Task:}
Based strictly on the provided relationship list and the rules above, determine the total number of distinct groups of people.

\textit{[Relationship list will be provided here]}

\textbf{Question:} How many distinct groups of people are there?

\textbf{Output Format:}
Provide your answer exactly as follows:
\texttt{Final Answer: <number>} 
    \end{tcolorbox}
    }
\end{figure*}

\begin{figure*}[ht]
    \centering
    \resizebox{\textwidth}{!}{
    \begin{tcolorbox}[prompt, title=Social Graph Analysis -- Relationship Counting Prompt]
You are analyzing social relationships based on defined rules.

\textbf{Relationship Rules:}
\begin{enumerate}
    \item Relationship Types: 'good' or 'bad'.
    \item Transitive Properties:
    \begin{itemize}
        \item Good-Good Transitivity: If A--good--B and B--good--C, then A--good--C.
        \item Bad Relationship Inference: If A--bad--B and A--good--C, then B--bad--C.
    \end{itemize}
    \item Group Definition: A 'group' is a set where every person has a 'good' relationship with every other person in that set (directly or via transitivity).
    \item Group Separation: People are in the same group if and only if they have good relationships. Groups are distinct; if anyone from Group 1 has a bad relationship with anyone from Group 2, then everyone in Group 1 has a bad relationship with everyone in Group 2.
\end{enumerate}

\textbf{Context:}
There are 14 people total, labeled A to N. You will be given a list of known relationships.

\textbf{Your Task:}
Based strictly on the provided relationship list and the rules above (including applying transitivity), count the total number of pairs of people who have 'good' relationships and the total number of pairs who have 'bad' relationships across all 14 people.

\textit{[Relationship list will be provided here]}

\textbf{Question:} How many pairs have good relationships, and how many pairs have bad relationships?

\textbf{Output Format:}
Provide your answer exactly as follows:
\texttt{Final Answer: X pairs have good relationships, Y pairs have bad relationships} 
    \end{tcolorbox}
    }
\end{figure*}

\begin{figure*}[ht]
    \centering
    \resizebox{\textwidth}{!}{
    \begin{tcolorbox}[prompt, title=Social Graph Analysis -- Group Membership Prompt]
You are analyzing social relationships based on defined rules.

\textbf{Relationship Rules:}
\begin{enumerate}
    \item Relationship Types: 'good' or 'bad'.
    \item Transitive Properties:
    \begin{itemize}
        \item Good-Good Transitivity: If A--good--B and B--good--C, then A--good--C.
        \item Bad Relationship Inference: If A--bad--B and A--good--C, then B--bad--C.
    \end{itemize}
    \item Group Definition: A 'group' is a set where every person has a 'good' relationship with every other person in that set (directly or via transitivity).
    \item Group Separation: People are in the same group if and only if they have good relationships. Groups are distinct; if anyone from Group 1 has a bad relationship with anyone from Group 2, then everyone in Group 1 has a bad relationship with everyone in Group 2.
\end{enumerate}

\textbf{Context:}
There are 14 people total, labeled A to N. You will be given a list of known relationships.

\textbf{Your Task:}
Based strictly on the provided relationship list and the rules above (including applying transitivity), identify all people who have a 'good' relationship with the person specified in the question.

\textit{[Relationship list will be provided here]}

\textbf{Question:} Who has a good relationship with H?

\textbf{Output Format:}
List the names in alphabetical order, separated by commas. If no one has a good relationship with the specified person (other than themselves, if applicable based on rules interpretation - assume self-relationships are not listed unless explicitly stated), answer 'No one'. Provide your answer exactly as follows:
\texttt{Final Answer: <list of people or 'No one'>} 
    \end{tcolorbox}
    }
\end{figure*}

\begin{figure*}[ht]
    \centering
    \resizebox{\textwidth}{!}{
    \begin{tcolorbox}[prompt, title=Social Graph Analysis -- Reasoning Prompt]
You are analyzing social relationships based on defined rules.

\textbf{Relationship Rules:}
\begin{enumerate}
    \item Relationship Types: 'good' or 'bad'.
    \item Transitive Properties:
    \begin{itemize}
        \item Good-Good Transitivity: If A--good--B and B--good--C, then A--good--C.
        \item Bad Relationship Inference: If A--bad--B and A--good--C, then B--bad--C.
    \end{itemize}
    \item Group Definition: A 'group' is a set where every person has a 'good' relationship with every other person in that set (directly or via transitivity).
    \item Group Separation: People are in the same group if and only if they have good relationships. Groups are distinct; if anyone from Group 1 has a bad relationship with anyone from Group 2, then everyone in Group 1 has a bad relationship with everyone in Group 2.
\end{enumerate}

\textbf{Context:}
There are 14 people total, labeled A to N. You will be given a list of known relationships.

\textbf{Your Task:}
Based strictly on the provided relationship list and the rules above (including applying transitivity), determine whether the specific relationship mentioned in the question is 'good' or 'bad', and answer 'Yes' if it's good, 'No' if it's bad.

\textit{[Relationship list will be provided here]}

\textbf{Question:} Do N and L have a good relationship?

\textbf{Output Format:}
Provide your answer exactly as follows:
\texttt{Final Answer: <Yes/No>}
    \end{tcolorbox}
    }
\end{figure*}

\begin{figure*}[ht]
    \centering
    \resizebox{\textwidth}{!}{
    \begin{tcolorbox}[prompt, title=Review Decision Prediction Evaluation Prompt]
You are an expert reviewer evaluating a research paper for a prestigious academic conference.

\textbf{Your Task:}
Analyze the provided paper information, reviewer comments, and author responses to determine whether the paper should be accepted or rejected for publication at a top-tier conference.

\textbf{Important Context:}
\begin{itemize}
    \item Judge the paper based on the standards of a highly selective, top-tier conference.
    \item Consider the strengths, weaknesses, novelty, significance, and clarity of the work.
    \item Evaluate the validity of reviewer concerns and the effectiveness of the author's rebuttal.
\end{itemize}

\textbf{Provided Information:}

\textbf{Paper Information:}
[Title, Abstract, Keywords, etc., will be provided here]

\textbf{Reviewer Comments:}
[Comments from multiple reviewers will be provided here]

\textbf{Author Response:}
[Author's rebuttal to reviewer comments will be provided here]

\textbf{Analysis and Output:}
Based on all the information provided, perform a careful analysis. First, provide your detailed reasoning, discussing the key factors influencing your decision (e.g., strengths, weaknesses, contribution, response to reviews). Then, conclude with your final decision in exactly this format:

\texttt{Final Decision: <Accept/Reject>} 
    \end{tcolorbox}
}
\end{figure*}

\begin{figure*}[ht]
    \centering
    \resizebox{\textwidth}{!}{
    \begin{tcolorbox}[prompt, title=User Profile Inference Comment Generation Prompt]
You are simulating a user commenting online.

\textbf{Context:}
You are commenting on this {content\_type}: "{content\_subject}".
Your simulated user persona is: {age\_group}, {gender}.

\textbf{Your Task:}
Write a realistic, authentic comment about the {content\_subject} that reflects your assigned persona.

\textbf{Instructions:}
\begin{enumerate}
    \item The comment should be 2--4 sentences long.
    \item It should sound like a genuine user from the specified demographic, not a professional writer or marketer.
    \item Consider incorporating common vocabulary, tone, or writing patterns associated with the {age\_group}.
\end{enumerate}

\textbf{Output Format:}
Write only the comment text itself. Do not include any explanations, labels, or quotation marks around the comment.
    \end{tcolorbox}
    }
\end{figure*}

\begin{figure*}[ht]
    \centering
    \resizebox{\textwidth}{!}{
    \begin{tcolorbox}[prompt, title=User Profile Inference Demographic Inference Prompt]
You are an AI assistant analyzing writing style to infer user demographics.

\textbf{Your Task:}
Analyze multiple comments written by the same person and determine their likely age group and gender based on writing style, vocabulary, topics of interest, and expressed perspectives.

\textbf{User Comments:}
\begin{quote}
"{comment\_1}" \\
"{comment\_2}" \\
"{comment\_3}" \\
\end{quote}

\textbf{Analysis and Output:}
Analyze these comments carefully, looking for patterns indicative of age and gender. Focus on identifying the single most likely demographic profile based *only* on the provided text.

First, explain your reasoning, citing specific examples from the text. Then, provide your final demographic classification in exactly this format:

\texttt{Age Group: <18-34 | 35-54 | 55+>} 
\texttt{Gender: <Male | Female | Non-binary>} 
    \end{tcolorbox}
    }
\end{figure*}

\end{document}